%% file: penn.tex
\documentclass{article}

\usepackage[final]{neurips_2022}

\usepackage[utf8]{inputenc} %
\usepackage[T1]{fontenc}    %
\usepackage{hyperref}       %
\usepackage{url}            %
\usepackage{booktabs}       %
\usepackage{amsfonts}       %
\usepackage{nicefrac}       %
\usepackage{microtype}      %
\usepackage{xcolor}         %

\usepackage[pdftex]{graphicx}

\usepackage{makecell}
\usepackage{caption}
\usepackage{graphicx}
\usepackage{stackengine}
\usepackage{textcomp}
\usepackage{multirow}
\usepackage{wrapfig}
\usepackage{hhline}
\usepackage{enumitem}

\input{math_commands.tex}

\newcommand{\supp}[1]{^{(#1)}}

\newcommand{\vR}{\mR}

\newcommand{\pdiff}[2]{\frac{\partial #1}{\partial #2}}
\newcommand{\inpdiff}[2]{\partial #1 / \partial #2}
\newcommand{\diff}[2]{\frac{\mathrm{d} #1}{\mathrm{d} #2}}

\newcommand{\Eqsref}[1]{Equations~\ref{#1}}
\newcommand{\Tabref}[1]{Table~\ref{#1}}
\newcommand{\Appref}[1]{Appendix~\ref{#1}}

\newcommand{\AD}{A\mbox{-}D}

\title{
  Physics-Embedded Neural Networks:
  Graph Neural PDE Solvers with Mixed Boundary Conditions
}

\author{%
  Masanobu Horie\\
  RICOS Co. Ltd.\\
  University of Tsukuba\\
  \texttt{horie@ricos.co.jp} \\
  \And
  Naoto Mitsume\\
  University of Tsukuba\\
  \texttt{mitsume@kz.tsukuba.ac.jp}\\
}

\begin{document}

\maketitle

\begin{abstract}
  Graph neural network (GNN) is a promising approach to learning and predicting physical phenomena described in boundary value problems, such as partial differential equations (PDEs) with boundary conditions. However, existing models inadequately treat boundary conditions essential for the reliable prediction of such problems. In addition, because of the locally connected nature of GNNs, it is difficult to accurately predict the state after a long time, where interaction between vertices tends to be global. We present our approach termed physics-embedded neural networks that considers boundary conditions and predicts the state after a long time using an implicit method. It is built based on an $\mathrm{E}(n)$-equivariant GNN, resulting in high generalization performance on various shapes. We demonstrate that our model learns flow phenomena in complex shapes and outperforms a well-optimized classical solver and a state-of-the-art machine learning model in speed-accuracy trade-off. Therefore, our model can be a useful standard for realizing reliable, fast, and accurate GNN-based PDE solvers. The code is available at \url{https://github.com/yellowshippo/penn-neurips2022}.
\end{abstract}

\section{Introduction}
\label{sec:introduction}
Partial differential equations (PDEs) are of interest to many scientists because of their application in various fields such as mathematics, physics, and engineering.
Numerical analysis is used to solve PDEs because most PDE problems in real life cannot be solved analytically.
For example, predicting fluid behavior in complex shapes is an essential topic because it is helpful for product design, disaster reduction, weather forecasting, and many others; however, it is a difficult problem and takes time to solve using classical solvers.
Machine learning is a promising approach to predicting such phenomena because it can utilize data similar to the state to be predicted, while classical solvers cannot.

However, the main challenge in dealing with complex phenomena such as fluids is to guarantee generalization performance because possible states in complex systems can be huge and may not be covered using a purely data-driven approach.
Therefore, we must apply appropriate inductive biases to machine learning models.
Many approaches successfully introduced various inductive biases such as local connectedness using graph neural networks (GNNs) and symmetry under coordinate transformations using equivariance.

While these methods have made great progress in solving PDEs using machine learning, there is still room for improvement.
First, there is need for an efficient and provable way to respect boundary conditions like Dirichlet and Neumann, i.e., mixed boundary conditions.
Rigorous fulfillment of Dirichlet boundary conditions is indispensable because they are hard constraints and different Dirichlet conditions correspond to different problems users would like to solve.
Second, there is need to reinforce the treatment of global interaction to predict the state after a long time, where interactions tend to be global.
GNNs have excellent generalization properties because of their locally-connected nature; however, they may miss global interaction due to their localness.

\begin{figure}[t]
  \centering
  \includegraphics[trim={2.3cm 9.5cm 2.3cm 2cm},clip,width=0.9\linewidth]
  {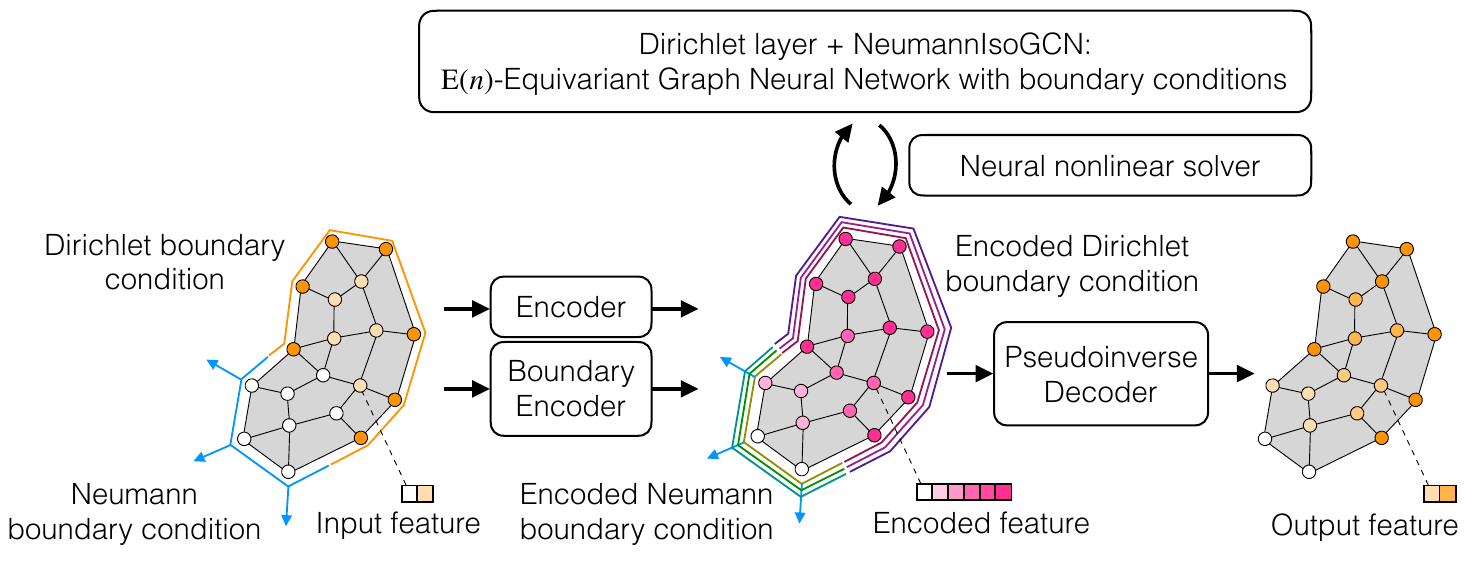}
  \caption{Overview of the proposed method.
  On decoding input features, we apply boundary encoders to boundary conditions.
  Thereafter, we apply a nonlinear solver consisting of an
  $\mathrm{E}(n)$-equivariant graph neural network in the encoded space.
  Here, we apply encoded boundary conditions for each iteration of the nonlinear solver.
  After the solver stops, we apply the pseudoinverse decoder to satisfy Dirichlet boundary conditions.
  }
  \label{fig:overview}
\end{figure}

We propose physics-embedded neural networks (PENNs), a machine learning framework to address these issues by embedding physics in the models.
We build our model based on IsoGCN~\citep{horie2021isometric}, a lightweight
$\mathrm{E}(n)$-equivariant GNN to reflect physical symmetry and realize fast prediction.
Furthermore, we construct a method to consider mixed boundary conditions.
Finally, we reconsider a way to stack GNNs based on a nonlinear solver, which naturally introduces the global pooling to GNNs as the global interaction with high interpretability.
In experiments, we demonstrate that our treatment of Neumann boundary conditions improves the predictive performance of the model, and our method can fulfill Dirichlet boundary conditions with no error.
Our method also achieves state-of-the-art performance compared to a classical, well-optimized numerical solver and a baseline machine learning model in speed-accuracy trade-off.
\Figref{fig:overview} shows the overview of the proposed model.
Our main contributions are summarized as follows:
\begin{itemize}
  \item
    We construct models to satisfy mixed boundary conditions: the \textit{boundary encoder}, \textit{Dirichlet layer}, \textit{pseudoinverse decoder}, and \textit{NeumannIsoGCN} (NIsoGCN).
    The considered models show provable fulfillment of boundary conditions, while existing models cannot.
  \item
    We propose \textit{neural nonlinear solvers}, which realize global connections to stably predict the state after a long time.
  \item
    We demonstrate that the proposed model shows state-of-the-art performance in speed-accuracy trade-off,
    and all the proposed components are compatible with $\mathrm{E}(n)$-equivariance.
\end{itemize}

\section{Background and related work}
\label{sec:background_related_works}
In this section, we review the foundation of PDEs to clarify the problems we solve and introduce related works where machine learning models are used to solve PDEs.

\subsection{Partial differential equations (PDEs) with boundary conditions}
\label{sec:pde}
A general form of the $d$-dimensional temporal PDEs that we consider can be expressed as follows:
\begin{align}
  \pdiff{\vu}{t} (t, \vx) &= \mathcal{D}(\vu)(t, \vx)
  & (t, \vx) \in (0, T) \times \Omega,
  \label{eq:pde}
  \\
  \vu(t = 0, \vx) &= \hat{\vu}_0(\vx) & \vx \in \Omega,
  \label{eq:initial}
  \\
  \vu(t, \vx) &= \hat{\vu}(t, \vx) & (t, \vx) \in (0, T) \times \partial\Omega_\mathrm{Dirichlet},
  \label{eq:Dirichlet}
  \\
  \hat{\vf}(\nabla \vu(t, \vx), \vn(\vx)) &= \bm{0}
  & (t, \vx) \in (0, T) \times \partial\Omega_\mathrm{Neumann},
  \label{eq:Neumann}
\end{align}
where $\Omega$ is the domain, $\partial \Omega$ is the boundary of $\Omega$, and
$\partial \Omega_\mathrm{Dirichlet}$ and
$\partial \Omega_\mathrm{Neumann}$ are boundaries with Dirichlet and Neumann
(mixed) boundary conditions.
$\hat{\cdot}$ is a known function, and $\mathcal{D}$ is a known nonlinear differential operator,
which can be nonlinear and contain spatial differential operators (see \eqref{eq:D_NS} for an example of $\mathcal{D}$).
$\vn(\vx)$ is the normal vector at $\vx \in \partial\Omega$.
\Eqref{eq:Dirichlet} is called the Dirichlet boundary condition, where
the value on $\partial\Omega_\mathrm{Dirichlet}$ is set as a constraint.
\Eqref{eq:Neumann} corresponds to the Neumann boundary condition, where
the value of the derivative $\vu$ in the direction of $\vn$ is set on
$\partial\Omega_\mathrm{Neumann}$
rather than the value of $\vu$.
When $\vu: (0, T) \times \Omega \to \mathbb{R}^f$ satisfies \Eqsref{eq:pde}~--~\ref{eq:Neumann},
it is called the solution of the (initial-) boundary value problem.

\subsubsection{Discretization}
\label{sec:discretization}
PDEs are defined in a continuous space to make differentials meaningful.
Discretization can be applied in space and time so that computers can solve PDEs easily.
In numerical analysis of complex-shaped domains, we commonly use meshes (discretized data of shapes), which can be regarded as graphs.
We denote the position of the $i$th vertex as $\vx_i$ and the value of a function $f$ at the $\vx_i$ as $f_i$.\footnote{
  Strictly speaking, components of the PDE e.g.\ $\mathcal{D}$ and $\Omega$ can be different before and after discretization.
  However, we use the same notation regardless of discretization to keep the notation simple.}

The simplest method to discretize time is the explicit Euler method formulated as:
\begin{align}
  \vu(t + \Delta t, \vx_i) = \vu(t, \vx_i) + \mathcal{D}(\vu)(t, \vx_i) \Delta t,
  \label{eq:explicit}
\end{align}
which updates $\vu(t, \vx_i)$ with a small increment
$\mathcal{D}(\vu)(t, \vx_i) \Delta t$.
Another way to have time discretization is the implicit Euler method formulated as:
\begin{align}
  \vu(t + \Delta t, \vx_i) = \vu(t, \vx_i) + \mathcal{D}(\vu)(t + \Delta t, \vx_i) \Delta t,
  \label{eq:implicit}
\end{align}
which solves \Eqref{eq:implicit} rather than simply updating variables to
ensure the original PDE is satisfied numerically.
The equation can be viewed as a nonlinear optimization problem by formulating it as:
\begin{align}
  \vR(\vv) &:= \vv - \vu(t, \cdot) - \mathcal{D}(\vv) \Delta t,
  \label{eq:residual}
  \\
  \mathrm{Solve}_\vv \ \vR(\vv)(\vx_i) &= \bm{0}, \ \forall i,
  \label{eq:optimization}
\end{align}
where
$\vR(\vv)$
is the residual vector of the discretized PDE.
The solution of \Eqref{eq:optimization} corresponds to
$\vu(t + \Delta t, \vx)$.
By letting
$\nabla\phi = \vR$ for an appropriate
$\phi$,
solving \Eqref{eq:optimization} corresponds to optimizing $\phi$ in an
$(f \times n)$-dimensional space, where
$n$ is the number of vertices in the considered mesh.
A simple way to solve such an optimization problem is to apply gradient descent formulated as:
\begin{align}
  \vv\supp{0} := \vu(t, \cdot),
  \hspace{10pt}
  \vv\supp{i+1} := \vv\supp{i} - \alpha\supp{i} \vR(\vv\supp{i}),
  \label{eq:gradient_descent}
\end{align}
where
$\alpha\supp{i} \in \mathbb{R}$ is determined using line search.
However, due to the high computational cost of the search,
$\alpha$ can be fixed to a small value,
which corresponds to the explicit Euler method with the time step size $\alpha \Delta t$.
\citet{barzilai1988two} suggested another simple yet effective way
to determine the step size using a two-point approximation to the secant equation underlying quasi-Newton methods.

\subsection{Neural PDE solvers}
We review machine learning models used to solve PDEs called neural PDE solvers, typically formulated as
$\vu(t_{n+1}, \vx_i) \approx \mathcal{F}_\mathrm{NN}(\vu)(t_n, \vx_i)$
for $(t_n, \vx_i) \in \{t_0, t_1, \dots\} \times \Omega$, where
$\mathcal{F}_\mathrm{NN}$ is a machine learning model.

\subsubsection{Physics-informed neural networks (PINNs)}
\citet{raissi2019physics} made a pioneering work combining PDE information and neural networks, called PINNs, by adding loss to monitor how much the output satisfies the equations.
PINNs can be used to solve forward and inverse problems and extract physical states from measurements~\citep{pang2019fpinns,mao2020physics,cai2021flow}.
However, PINNs' outputs should be functions of space because PINNs rely on automatic differentiation to obtain loss regarding PDEs.
This design constraint significantly limits the model's generalization ability because the solution of a PDE could be entirely different when the shape of the domain or boundary condition changes.
Besides, the loss reflecting PDEs helps models learn physics at training time; however, prediction by PINN models can be out of physics because of lacking PDE information inside the model.
Therefore, these methods are not applicable in building models that are generalizable over shape and boundary condition variations.
As seen in \Secref{sec:method}, our model contains PDE information inside and does not take absolute positions of vertices, thus resulting in high generalizability
(See \Figref{fig:fluid_results}).

\subsubsection{Graph neural network based PDE solvers}
As discussed in \Secref{sec:discretization}, one can regard a mesh as a graph.
GNNs can take any graphs as inputs~\citep{gori2005new,scarselli2008graph,kipf2017semi,gilmer2017neural},
having the possibility to generalize over various graphs, i.e., meshes.
Therefore, GNNs are strong candidates for learning mesh-structured numerical analysis data, as seen in
\citet{alet19a, chang2020learning, pfaff2021learning}.
\citet{brandstetter2022message} advanced these works
for efficient and stable prediction.
Their method could also consider boundary conditions by feeding them to the models as inputs.
Here, one could expect the model to learn to satisfy boundary conditions approximately, while there is no guarantee to fulfill hard constraints such as Dirichlet conditions.
In contrast, our model ensures the satisfaction of boundary conditions.
Besides, most GNNs use local connections with a fixed number of message passings, which lacks consideration of global interaction.
We suggest an effective way to incorporate a global connection with GNN through the neural nonlinear solver.

\subsubsection{Equivariant models}
In addition to GNNs, another essential concept to help machine learning models generalize is equivariance.
Equivariance is characterized by using group action as
$f(g\cdot x) = g \cdot f(x)$ for $f: X \to Y$ and $g \in G$ acting on $X$ and $Y$.
In particular, $\mathrm{E}(n)$-equivariance is essential to predict the solutions of physical PDEs because it describes rigid body motion,
i.e., translation, rotation, and reflection.
\citet{ling2016reynolds} and \citet{wang2021incorporating} introduced equivariance to a simple neural network and CNN to predict flow phenomena.
Both works showed that equivariance improved predictive and generalization performance compared to models without equivariance.
\citet{horie2021isometric} proposed $\mathrm{E}(n)$-equivariant GNNs based on GCNs~\citep{kipf2017semi}, called IsoGCNs.
A form of their model is formulated as:
\begin{align}
  \left[\nabla \psi\right]_i \approx \left[\mathrm{IsoGCN}_{0 \to 1}(\psi)\right]_i
  :=
  {\left[\sum_{l \in \mathcal{N}_i}
  \frac{\vx_l - \vx_i}{\Vert\vx_l - \vx_i\Vert}
  \otimes
  \frac{\vx_l - \vx_i}{\Vert\vx_l - \vx_i\Vert}
  \right]}^{-1}
  \sum_{j \in \mathcal{N}_i}
  \frac{\psi_j - \psi_i}{\Vert\vx_j - \vx_i\Vert}
  \frac{\vx_j - \vx_i}{\Vert\vx_j - \vx_i\Vert} \mW,
\end{align}
where $\mathcal{N}_i$ is the neighborhood of the $i$th vertex,
$\otimes$ is the tensor product operator, and
$\mW$ is a trainable matrix acting on feature index.
Here, we denote $\mathrm{IsoGCN}_{0 \to 1}$ an IsoGCN layer that
converts the input scalar (rank-0 tensor) field $\psi$ to
the output vector (rank-1 tensor) field.
This layer corresponds to the gradient operator, which helps learn PDEs
because spatial derivatives such as gradient play an essential role in PDEs.
They applied the model to the heat equation problem, showing
high predictive performance and fast prediction,
while boundary condition treatment was out of their scope.

\section{Proposed method}
\label{sec:method}
We present our model architecture.
We adopt an encode-process-decode architecture, proposed by \citet{battaglia2018relational},
which has been applied successfully in various previous works, e.g., \citet{horie2021isometric,brandstetter2022message}.
Our key concept is to encode input features, including information on boundary conditions, apply a GNN-based nonlinear solver loop reflecting boundary conditions in the encoded space, then decode carefully to satisfy boundary conditions in the output space.

\subsection{Dirichlet boundary model}
As demonstrated theoretically and experimentally in literature~\citep{hornik1991approximation,cybenko1992approximation,nakkiran2021deep},
the expressive power of neural networks comes from encoding in a higher-dimensional space, where the corresponding boundary conditions are not trivial.
However, if there are no boundary condition treatments in layers inside the processor,
which resides in the encoded space,
the trajectory of the solution can be far from the one with boundary conditions.
Therefore, boundary condition treatments in an encoded space are essential for obtaining reliable neural PDE solvers that fulfill boundary conditions.

To ensure the same encoded space between variables and boundary conditions, we use the same encoder for variables and the corresponding Dirichlet boundary conditions,
which we term the \textit{boundary encoder}, as follows:
\begin{align}
  \vh_i = \vf_\mathrm{encode}(\vu_i)
  \ \mathrm{in} \ \Omega,
  \ \ \ \hat{\vh}_i = \vf_\mathrm{encode}(\hat{\vu}_i)
  \ \mathrm{on} \ \partial\Omega_\mathrm{Dirichlet}
\end{align}

One can easily apply Dirichlet boundary conditions in the aforementioned encoded space using
the \textit{Dirichlet layer} defined as:
\begin{align}
  \mathrm{DirichletLayer}(\vh_i) = \left\{
    \begin{array}{cc}
      \vh_i, & \vx_i \notin \partial\Omega_\mathrm{Dirichlet}
      \\
      \hat{\vh}_i, & \vx_i \in \partial\Omega_\mathrm{Dirichlet}
    \end{array}
    \right.
\end{align}
This process is necessary to return to the state respecting the boundary conditions after some operations in the processor, which might disrespect the conditions.

After the processor layers, we decode the hidden features using functions satisfying:
\begin{align}
  \vf_\mathrm{decode} \circ \vf_\mathrm{encode}(\hat{\vu}_i) = \hat{\vu}_i
  \ \mathrm{on} \ \partial\Omega_\mathrm{Dirichlet}
  \label{eq:pseudoinverse}
\end{align}
This condition ensures that the encoded boundary conditions correspond to the ones in the original physical space.
Demanding that \Eqref{eq:pseudoinverse} holds for arbitrary
$\hat{\vu}$;
we obtain
$\vf_\mathrm{decode}\circ\vf_\mathrm{encode} = \mathrm{Id}_\vu$, resulting in
$\vf_\mathrm{decode} = \vf_\mathrm{encode}^+$,
which we call the \textit{pseudoinverse decoder}.
It is pseudoinverse because
$\vf_\mathrm{encode}$, in particular encoding in a higher-dimensional space, may not be invertible.
Therefore, we construct
$\vf_\mathrm{encode}^+$ using pseudoinverse matrices.
For more details,
see \Appref{app:pseudoinverse}.

\subsection{Neumann boundary model}
\citet{matsunaga2020improved} proposed a wall boundary model to deal with Neumann boundary conditions
for the least squares moving particle semi-implicit (LSMPS) method~\citep{tamai2014least}, a framework to solve PDEs using particles.
The LSMPS method is the origin of the IsoGCN's gradient operator, so one can imagine that the wall boundary model may introduce a sophisticated treatment of Neumann boundary conditions into IsoGCN.
We modified the wall boundary model to adapt to the situation where the vertices are on the Neumann boundary, which differs from the situation of particle simulations
(see \Appref{app:NIsoGCN} for more details).
Our formulation of IsoGCN with Neumann boundary conditions, which is termed \textit{NeumannIsoGCN} (NIsoGCN), is expressed as:
\begin{align}
  \mathrm{NIsoGCN}_{0 \to 1}(\psi)
  &:=
  {\mM_i}^{-1}
  {\left[
  \sum_{j \in \mathcal{N}_i}
  \frac{\psi_j - \psi_i}{\Vert\vx_j - \vx_i\Vert}
  \frac{\vx_j - \vx_i}{\Vert\vx_j - \vx_i\Vert}
  + w_i \vn_i \hat{g}_i
  \right]} \mW
  \label{eq:NIsoGCN}
  \\
  \mM_i &:= \sum_{l \in \mathcal{N}_i}
  \frac{\vx_l - \vx_i}{\Vert\vx_l - \vx_i\Vert}
  \otimes
  \frac{\vx_l - \vx_i}{\Vert\vx_l - \vx_i\Vert}
  + w_i \vn_i \otimes \vn_i,
  \label{eq:moment_NIsoGCN}
\end{align}
where $\hat{g}_i$ is the value of the Neumann boundary condition at $\vx_i$,
$\mW$ is a trainable matrix,
and $w_i > 0$ is an untrainable parameter to control the strength of the Neumann constraint.
As $w_i \to \infty$, the model strictly satisfies the given Neumann condition in the direction
$\vn_i$,
while the directional derivatives in the direction of
$(\vx_j - \vx_i)$
tend to be relatively neglected.
Thus, we keep the value of
$w_i$
moderate to consider derivatives in both
$\vn$ and $\vx$ directions.
In particular, we set
$w_i = 10.0$,
assuming that around ten vertices may virtually exist "outside" the boundary on a flat surface in a 3D space.

NIsoGCN is a straightforward generalization of the original IsoGCN by letting
$\vn_i = \bm{0}$ when $\vx_i \notin \partial\Omega_\mathrm{Neumann}$.
This model can also be generalized to vectors or higher rank tensors, similarly to the original IsoGCN's construction
(see \Appref{app:NIsoGCN}).
Therefore, NIsoGCN can express any spatial differential operator, constituting
$\mathcal{D}$
in PDEs.

\subsection{Neural nonlinear solver}
As reviewed in \Secref{sec:pde}, one can regard solving PDEs as optimization.
Here, we adopt the Barzilai--Borwein method~\citep{barzilai1988two}
to solve \Eqref{eq:optimization} in the encoded space.
In our case, the step size $\alpha\supp{i}$ of gradient descent is approximated as:
\begin{align}
  \alpha\supp{i} \approx \alpha\supp{i}_\mathrm{BB} :=\frac{
    \left<\vh\supp{i} - \vh\supp{i - 1}, \vR(\vh\supp{i}) - \vR(\vh\supp{i-1})\right>_\Omega}{
    \left<\vR(\vh\supp{i}) - \vR(\vh\supp{i - 1}), \vR(\vh\supp{i}) - \vR(\vh\supp{i-1})\right>_\Omega},
  \label{eq:BB}
\end{align}
wherer
$\vR(\vh)$ is the residual vector in the encoded space and
$\left<\vf, \vg\right>_\Omega := \sum_{\vx_i \in \Omega}\vf(\vx_i) \cdot \vg(\vx_i)$
denotes the inner product over the mesh.
Because the inner product is taken all over the mesh (graph), computing
$\alpha_\mathrm{BB}\supp{i}$ corresponds to global pooling.
With that view, one can find similarities between \Eqref{eq:gradient_descent}
and deep sets~\citep{zaheer2017deep},
which is a successful method to learn point cloud data
and has a strong background regarding permutation equivariance.
For more details, see \Appref{app:BB}.

Our aim is to use \Eqref{eq:BB}, approximating the nonlinear differential operator
$\mathcal{D}$ in \Eqref{eq:residual} with NIsoGCN.
By doing this, we expect the processor to consider both local and global information,
which may have an advantage over simply stacking GNNs corresponding to the explicit method as discussed in \Secref{sec:discretization}.
Combinations of solvers and neural networks are already suggested in, e.g., NeuralODE~\citep{chen2018neural}.
The novelty of our study is the extension of existing methods for solving PDEs with spatial structure and the incorporation of global pooling into the solver, enabling us to capture global interaction, which we refer to as
the \textit{neural nonlinear solver}.
Finally, the update from the state at the
$i$th iteration $\vh^{(i)}$ to the
$(i+1)$th in the neural nonlinear solver is expressed as:
\begin{align}
  \vh\supp{i+1}
  = \mathrm{DirichletLayer}\left(
    \vh\supp{i} - \alpha_\mathrm{BB}\supp{i}
    \left[
      \vh\supp{i} - \vh\supp{0} - \mathcal{D}_\mathrm{NIsoGCN}(\vh\supp{i})\Delta t
    \right]
  \right),
  \label{eq:neural_nonlinear_solver}
\end{align}
where
$\vh\supp{0}$ is the encoded $\vu(t, \cdot)$ reflecting \Eqref{eq:gradient_descent} and
$\mathcal{D}_\mathrm{NIsoGCN}$ is an $\mathrm{E}(n)$-equivariant GNN reflecting
the structure of $\mathcal{D}$ using differential operators provided by NIsoGCN.
Here, \Eqref{eq:neural_nonlinear_solver} enforces hidden features to satisfy the encoded PDE, including boundary conditions,
motivating us to call our model \textit{physics-embedded neural networks}
because it embeds physics (PDEs) in the model rather than in the loss.

\bgroup
\def\arraystretch{1.3}
\begin{table}[t]
  \caption{MSE loss ($\pm$ the standard error of the mean)
  on test dataset of gradient prediction.
  $\hat{g}_\mathrm{Neumann}$ is the loss computed only on the boundary
  where the Neuman condition is set.
  }
  \centering
  \label{tab:gradient}
  \scalebox{1.0}{
    \begin{tabular}{lrr}
      \\[-8pt]
      \toprule
      Method
      & \makecell[l]{$\nabla \phi (\times 10^{-3})$}
      & \makecell[l]{$\hat{g}_\mathrm{Neumann} (\times 10^{-3})$}
      \\
      \hline
      Original IsoGCN &
      $192.72 \pm 1.69$ &
      $1390.95 \pm 7.93$
      \\
      \textbf{NIsoGCN} (Ours) &
      $6.70 \pm 0.15$ &
      $3.52 \pm 0.02$
      \\
      \bottomrule
    \end{tabular}
  }
\end{table}
\egroup

\begin{figure}[t]
  \centering

  \stackunder[3pt]
  {\includegraphics[trim={15cm -10cm 35cm 0cm},clip,width=0.15\textwidth]
    {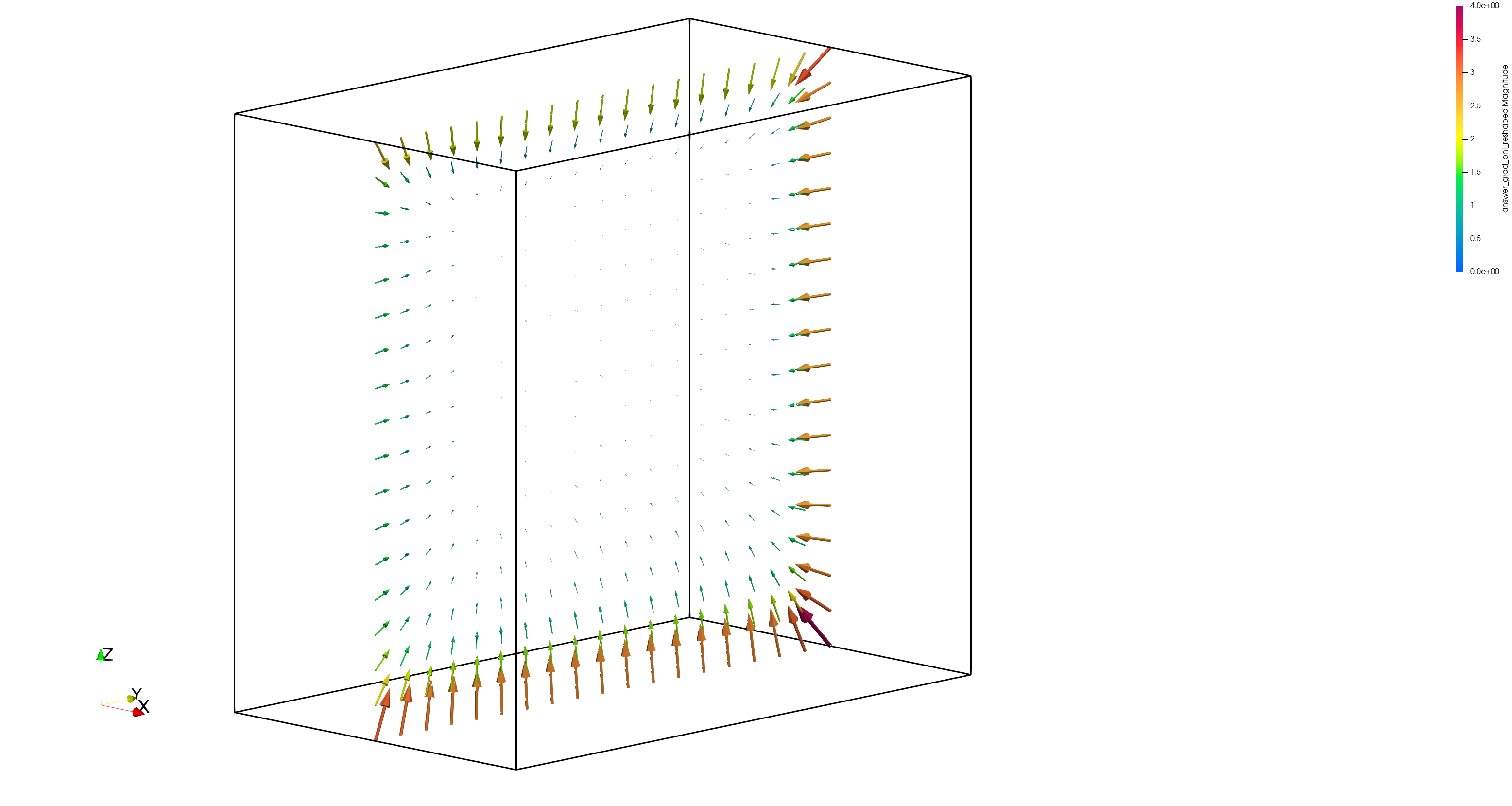}}
  {\hphantom{Ground truth}}
  \hspace{30pt}
  \stackunder[3pt]
  {\includegraphics[trim={15cm -10cm 35cm 0cm},clip,width=0.15\textwidth]
    {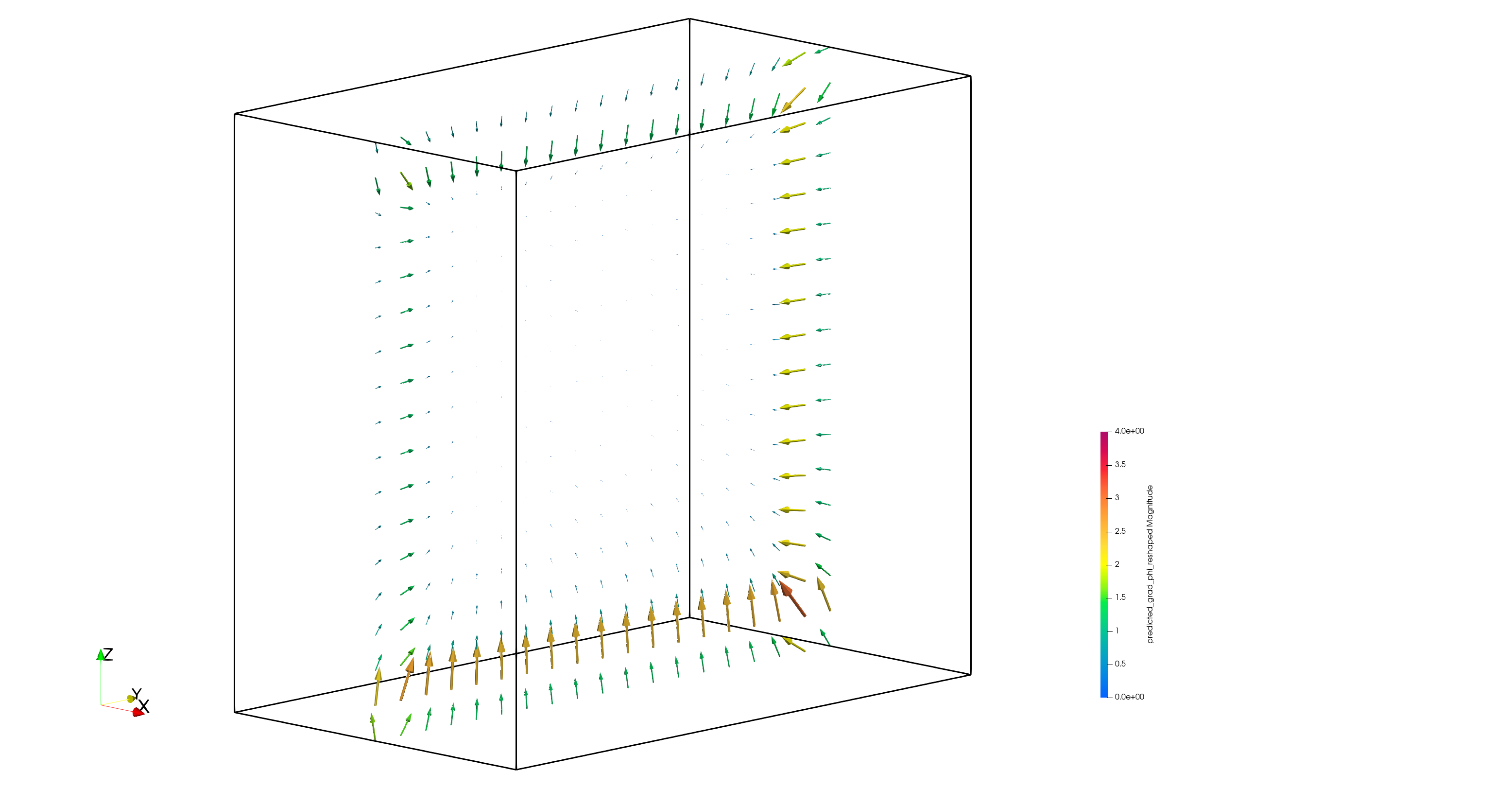}}
  {\hphantom{Original IsoGCN}}
  \hspace{30pt}
  \stackunder[3pt]
  {\includegraphics[trim={15cm -10cm 35cm 0cm},clip,width=0.15\textwidth]
    {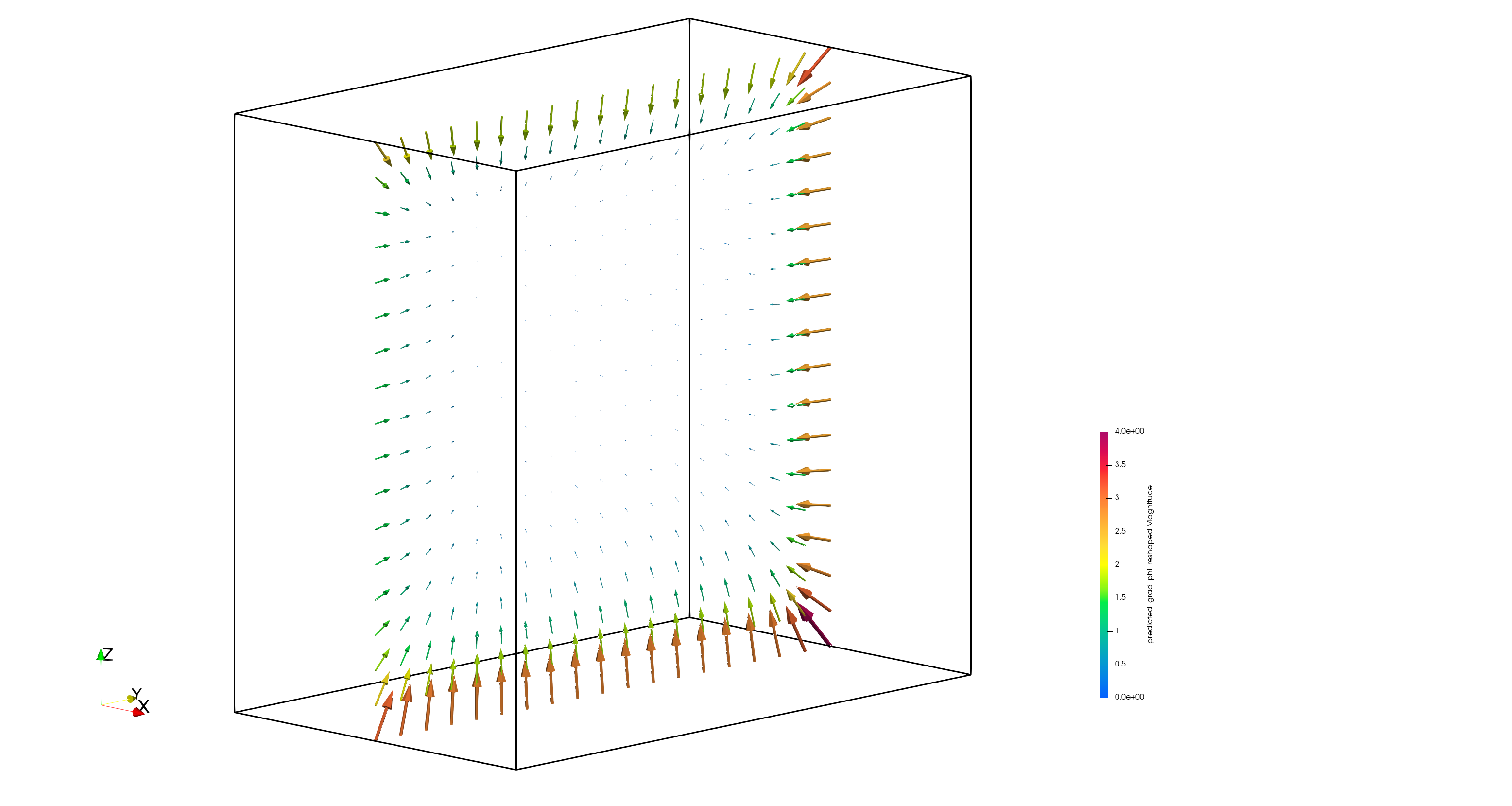}}
  {\hphantom{NIsoGCN}}
  \hspace{30pt}
  \stackunder[3pt]
  {\includegraphics[trim={0cm -3cm 0cm 0cm},width=0.07\textwidth]
    {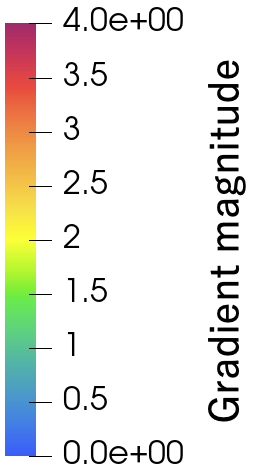}}
  {}
  \\[-10pt]
  \stackunder[-10pt]
  {\hphantom{\includegraphics[trim={15cm -10cm 35cm 0cm},clip,width=0.15\textwidth]
    {figs/grad/learned_grad/answer.png}}}
  {Ground truth}
  \hspace{30pt}
  \stackunder[-10pt]
  {\includegraphics[trim={15cm -10cm 35cm 0cm},clip,width=0.15\textwidth]
    {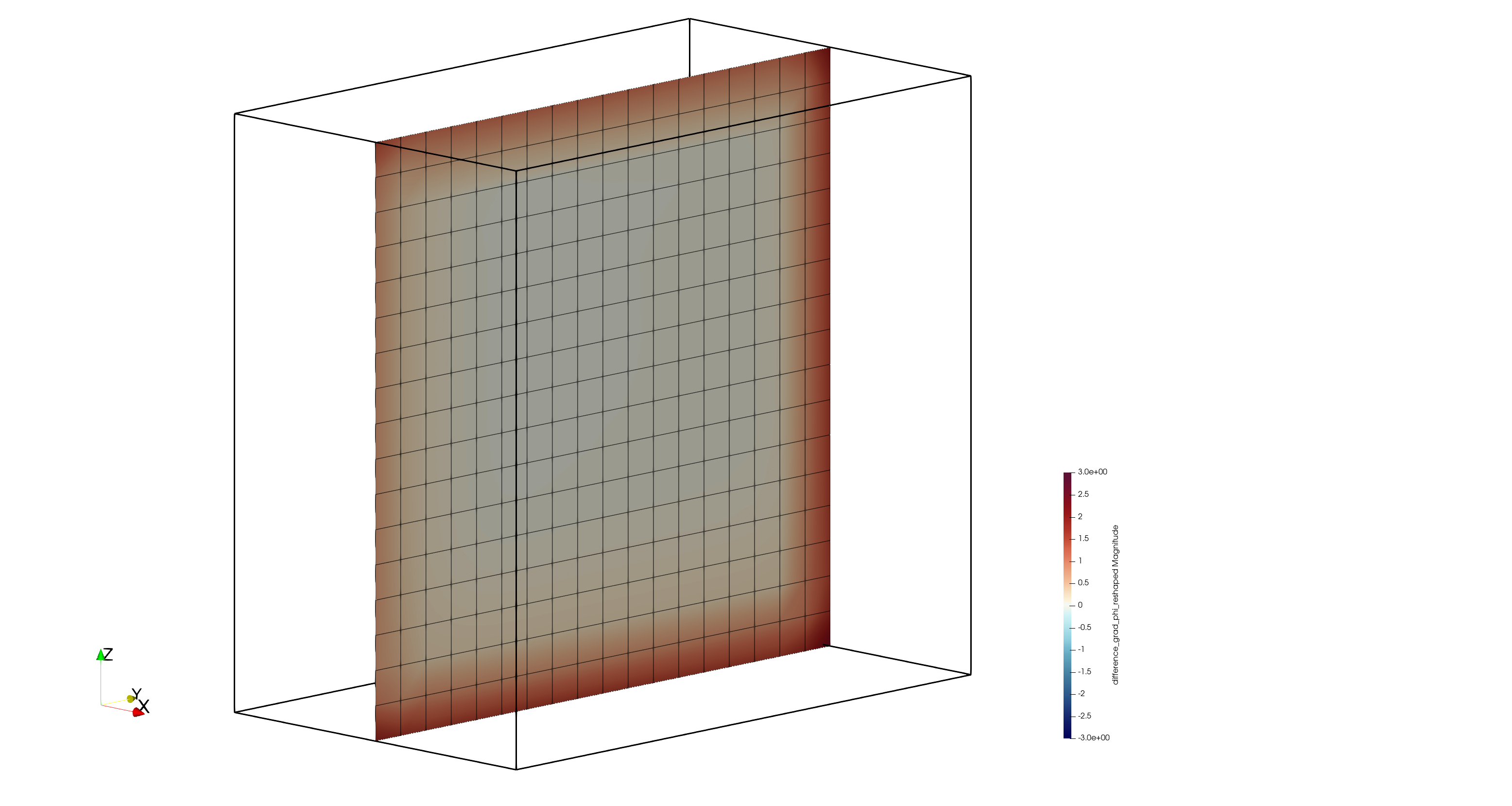}}
  {Original IsoGCN}
  \hspace{30pt}
  \stackunder[-10pt]
  {\includegraphics[trim={15cm -10cm 35cm 0cm},clip,width=0.15\textwidth]
    {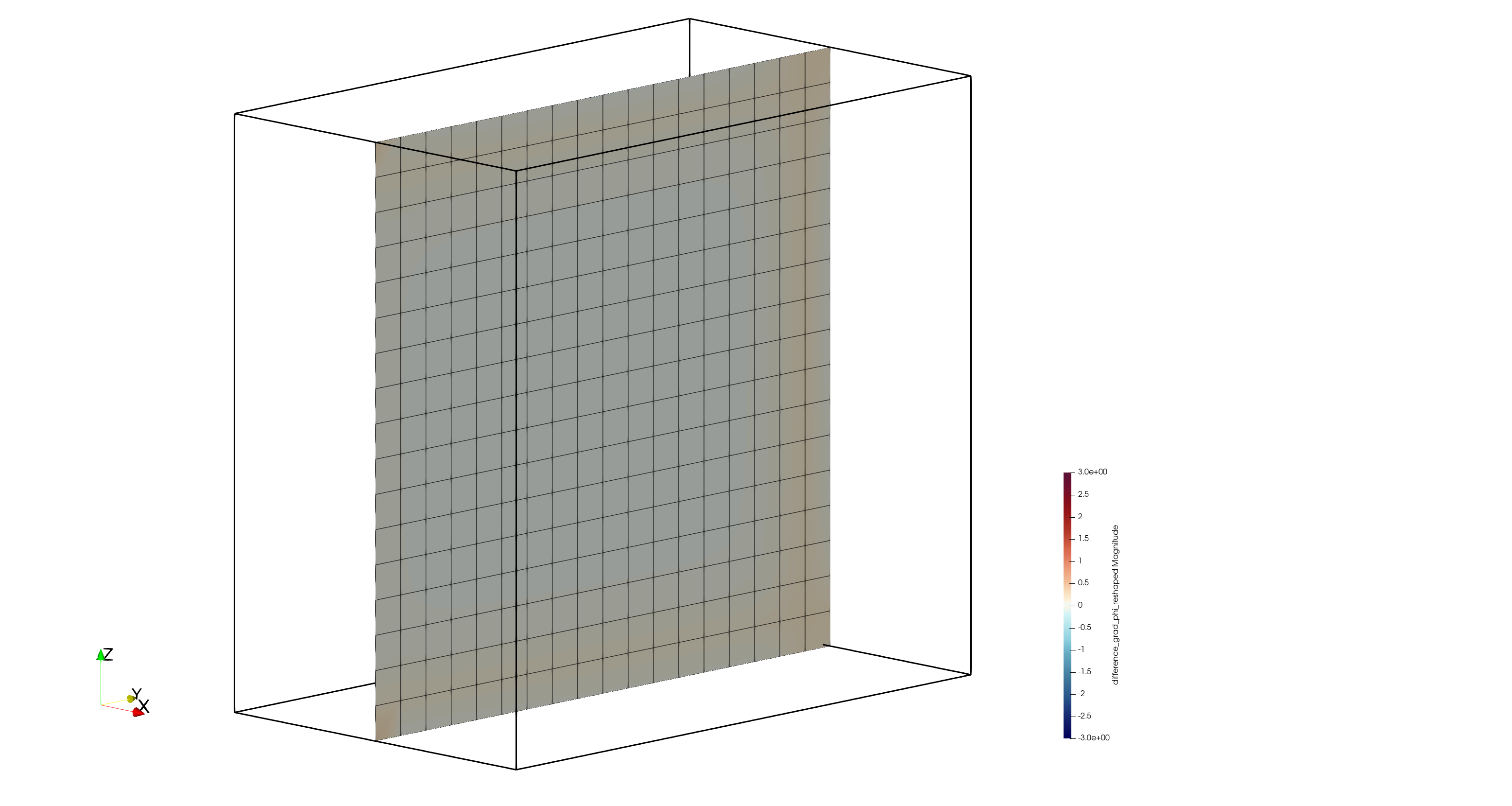}}
  {NIsoGCN}
  \hspace{30pt}
  \stackunder[-10pt]
  {\includegraphics[trim={0cm 0cm 0cm 0cm},width=0.07\textwidth]
    {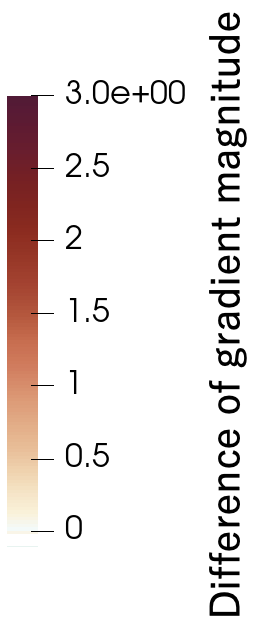}}
  {}

  \caption{
    Gradient field (top) and the magnitude of error between the predicted
    gradient and the ground truth (bottom) of a test data sample, sliced
    on the center of the mesh.
  }
  \label{fig:grad_results}
\end{figure}
\section{Experiments}
\label{sec:experiments}
Using numerical experiments, we demonstrate the proposed model's validity, expressibility, and computational efficiency.
We use two types of datasets: 1) the gradient dataset to verify the correctness of NIsoGCN and 2) the incompressible flow dataset to demonstrate the speed and accuracy of the model.
We also present ablation study results to corroborate the effectiveness of the proposed method.
The implementation of our model is based on the original IsoGCN's code.\footnote{
  \url{https://github.com/yellowshippo/isogcn-iclr2021}, Apache License 2.0.\label{footnote:isogcn}}
Our implementation is available online.\footnote{
  \url{https://github.com/yellowshippo/penn-neurips2022}, Apache License 2.0.\label{footnote:penn}}
All the details of the experiments
and another simple experiment
can be found in \Appref{app:gradient}, \ref{app:fluid}, and~\ref{app:ad}.

\subsection{Gradient dataset}
As done in \citet{horie2021isometric}, we conducted experiments to predict the gradient field from a given scalar field to verify the expressive power of NIsoGCN.
We generated cuboid-shaped meshes randomly with 10 to 20 cells in the X, Y, and Z directions.
We then generated random scalar fields over these meshes using polynomials of degree 10 and computed their gradient fields analytically.
Our training, validation, and test datasets consisted of 100 samples.
\Tabref{tab:gradient} and \Figref{fig:grad_results} show that the proposed NIsoGCN improves gradient prediction, especially near the boundary, showing that our model successfully considers Neumann boundary conditions.

\bgroup
\def\arraystretch{1.3}
\begin{table}[bt]
  \caption{MSE loss ($\pm$ the standard error of the mean)
  on test dataset of incompressible flow.
  If "Trans." is "Yes," it means evaluation is done on randomly
  rotated and transformed test dataset.
  $\hat{\cdot}_\mathrm{Dirichlet}$ is the loss computed only on the boundary
  where the Dirichlet condition is set for each $\vu$ and $p$.
  MP-PDE's results are based on the time window size equaling 40
  as it showed the best performance in the tested MP-PDEs.
  For complete results, see \Tabref{tab:fluid_results_detailed}.
  }
  \label{tab:fluid_results}
  \centering
  \scalebox{1.0}{
    \begin{tabular}{llrrrr}
      \\[-8pt]
      \toprule
      Method
      &
      Trans.
      & \makecell{$\vu$\\$(\times 10^{-4})$}
      & \makecell{$p$\\$(\times 10^{-3})$}
      & \makecell{$\hat{\vu}_\mathrm{Dirichlet}$\\$(\times 10^{-4})$}
      & \makecell{$\hat{p}_\mathrm{Dirichlet}$\\$(\times 10^{-3})$}
      \\
      \hline
      \multirow{2}{*}{\makecell[l]{MP-PDE\\TW = 20}} &
No &
$\boldsymbol{1.30} \pm 0.01$ &
$1.32 \pm 0.01$ &
$0.45 \pm 0.01$ &
$0.28 \pm 0.02$
\\
& Yes &
$1953.62 \pm 7.62$ &
$281.86 \pm 0.78$ &
$924.73 \pm 6.14$ &
$202.97 \pm 3.81$
      \\[3pt]
      \multirow{2}{*}{\textbf{PENN} (Ours)} &
No &
$4.36 \pm 0.03$ &
$\boldsymbol{1.17} \pm 0.01$ &
$\boldsymbol{0.00} \pm 0.00$ &
$\boldsymbol{0.00} \pm 0.00$
\\
& Yes &
$\boldsymbol{4.36} \pm 0.03$ &
$\boldsymbol{1.17} \pm 0.01$ &
$\boldsymbol{0.00} \pm 0.00$ &
$\boldsymbol{0.00} \pm 0.00$
\\
    \bottomrule
    \end{tabular}
  }
\end{table}
\egroup

\begin{figure}[t]
  \centering

  \stackunder[3pt]
  {\includegraphics[trim={27cm 19cm 27cm 19cm},clip,width=0.21\textwidth]
    {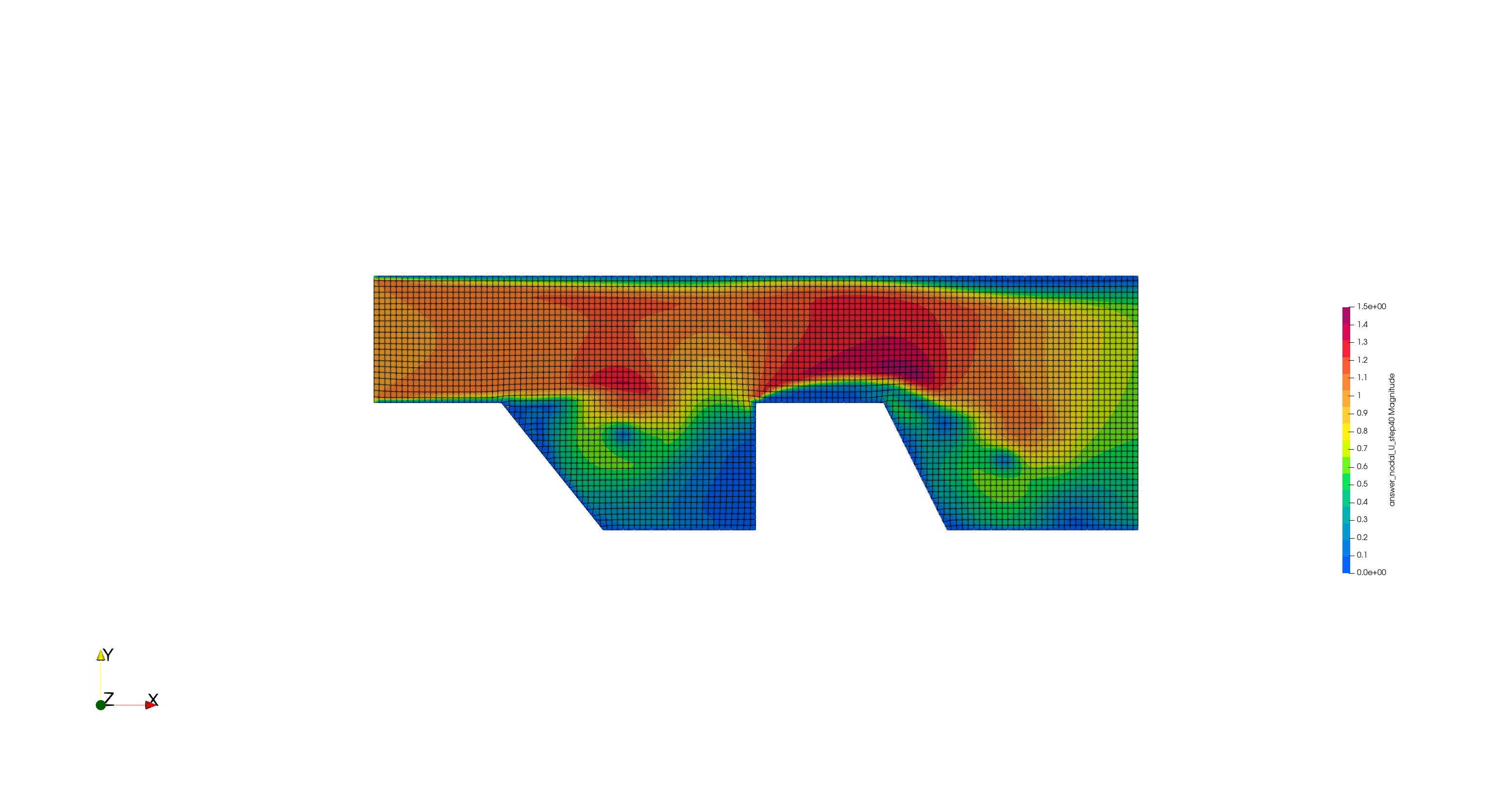}}
  {}
  \hspace{5pt}
  \stackunder[3pt]
  {\includegraphics[trim={27cm 19cm 27cm 19cm},clip,width=0.21\textwidth]
    {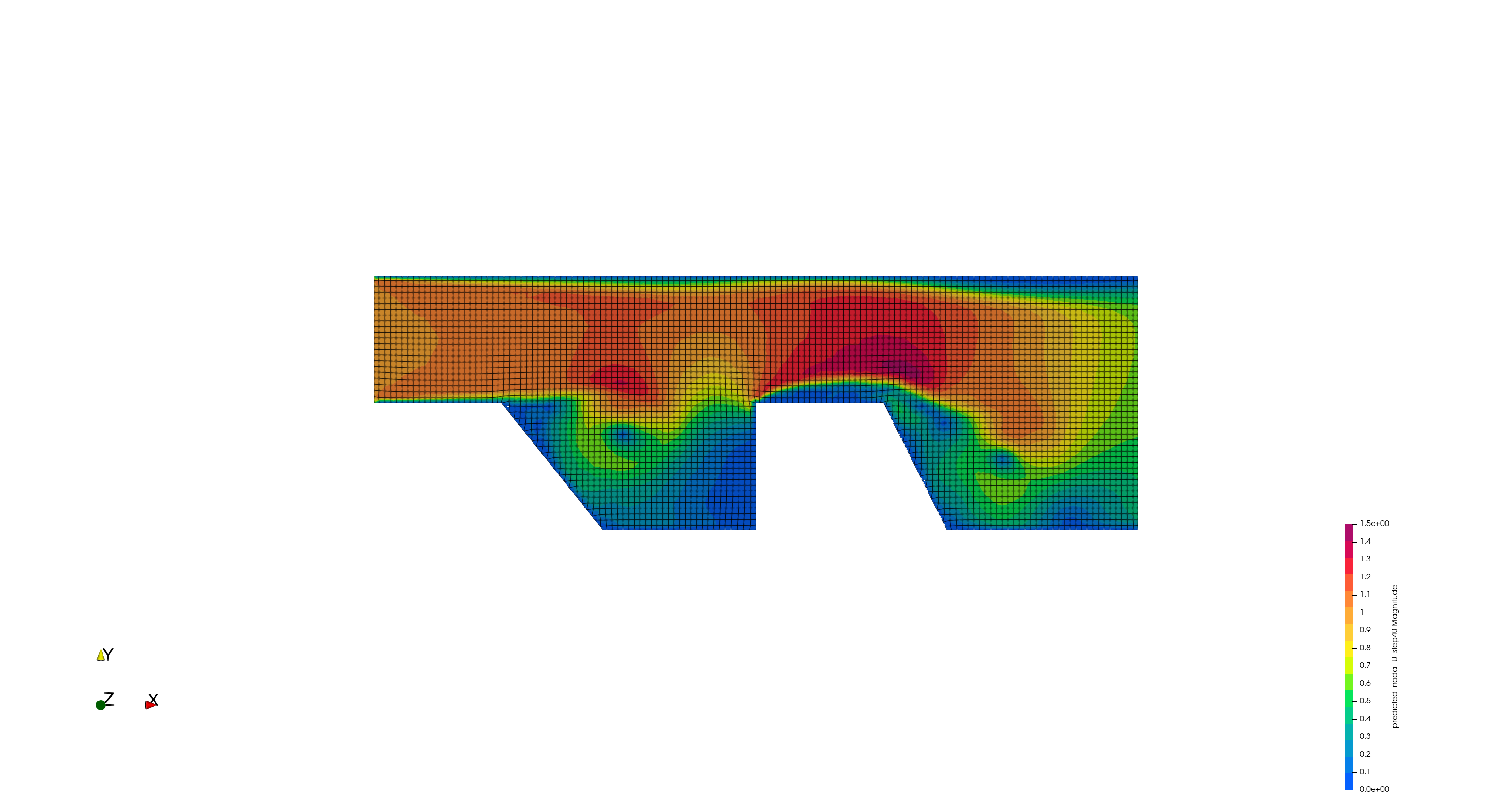}}
  {}
  \hspace{5pt}
  \stackunder[3pt]
  {\includegraphics[trim={27cm 19cm 27cm 19cm},clip,width=0.21\textwidth]
    {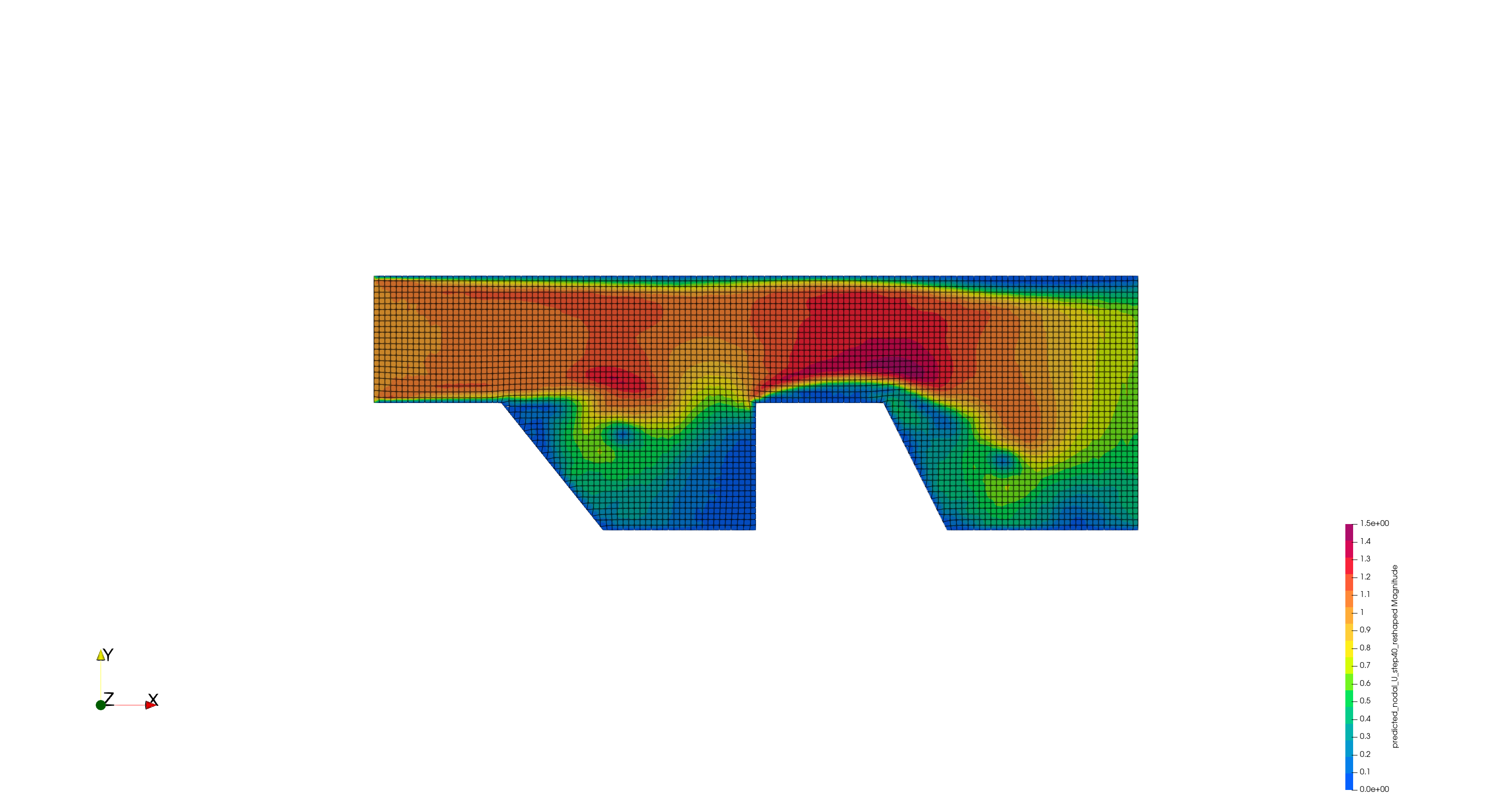}}
  {}
  {\includegraphics[trim={0cm 100cm 0cm 100cm},clip,width=0.1\textwidth]
    {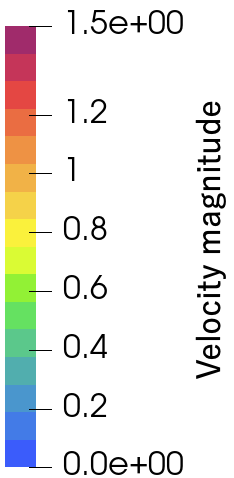}}
  \\[-10pt]

  \stackunder[3pt]
  {\includegraphics[trim={0 0 0 0},clip,width=0.05\textwidth]
    {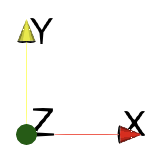}}
  {}
  {\includegraphics[trim={30cm 0cm 30cm 0cm},clip,width=0.18\textwidth]
    {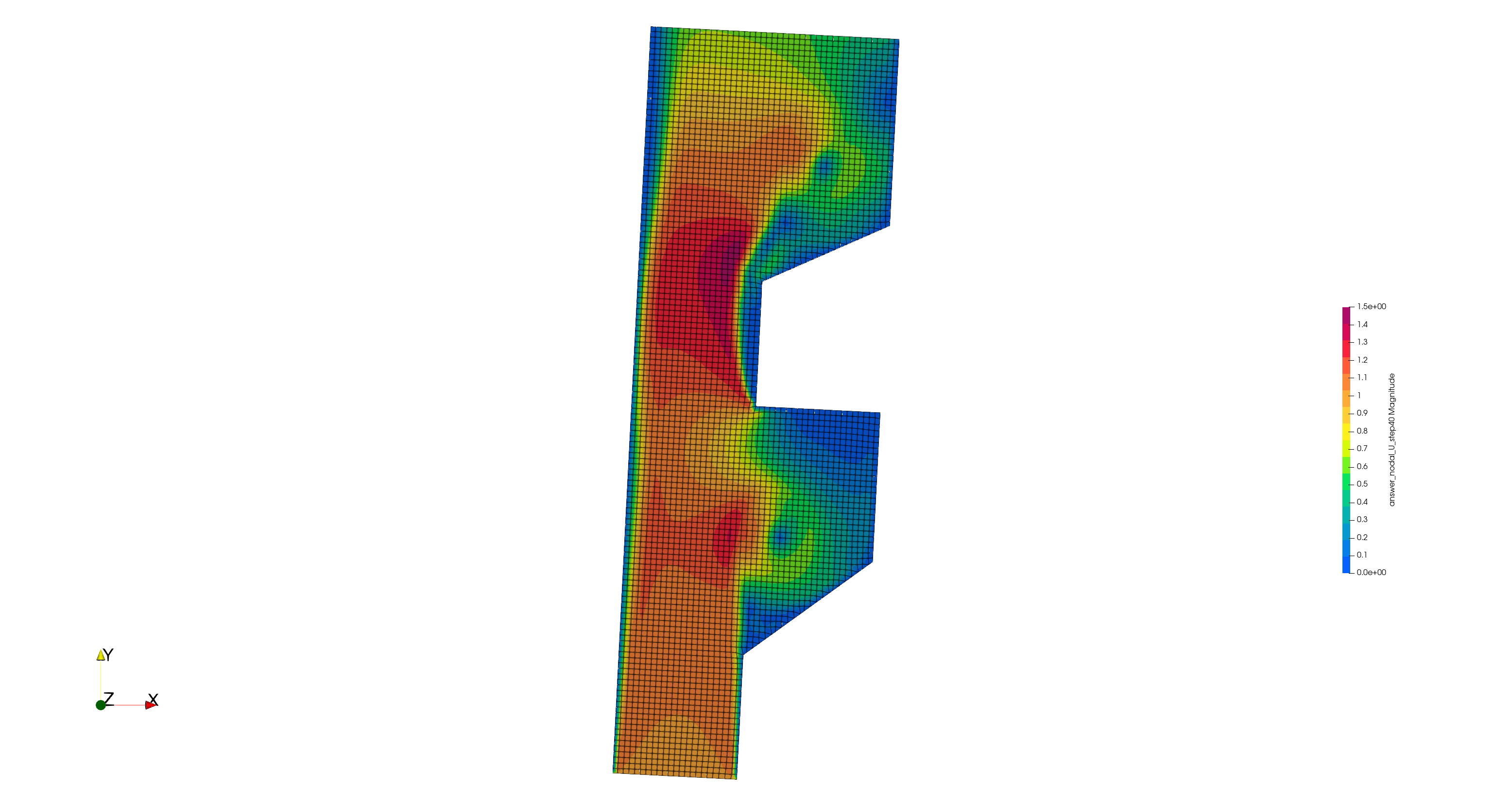}}
  {}
  \hspace{15pt}
  \stackunder[3pt]
  {\includegraphics[trim={30cm 0cm 30cm 0cm},clip,width=0.18\textwidth]
    {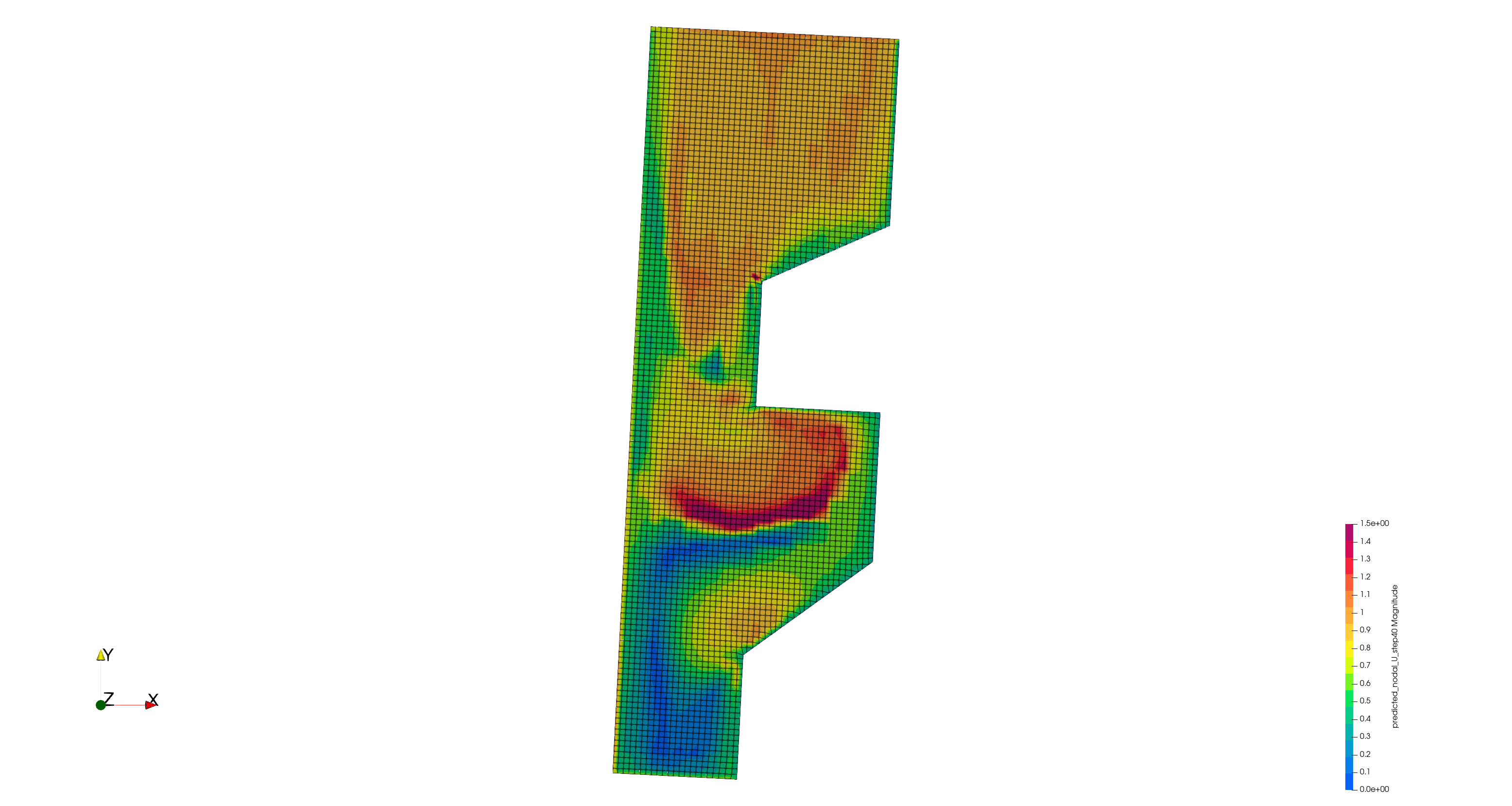}}
  {}
  \hspace{15pt}
  \stackunder[3pt]
  {\includegraphics[trim={30cm 0cm 30cm 0cm},clip,width=0.18\textwidth]
    {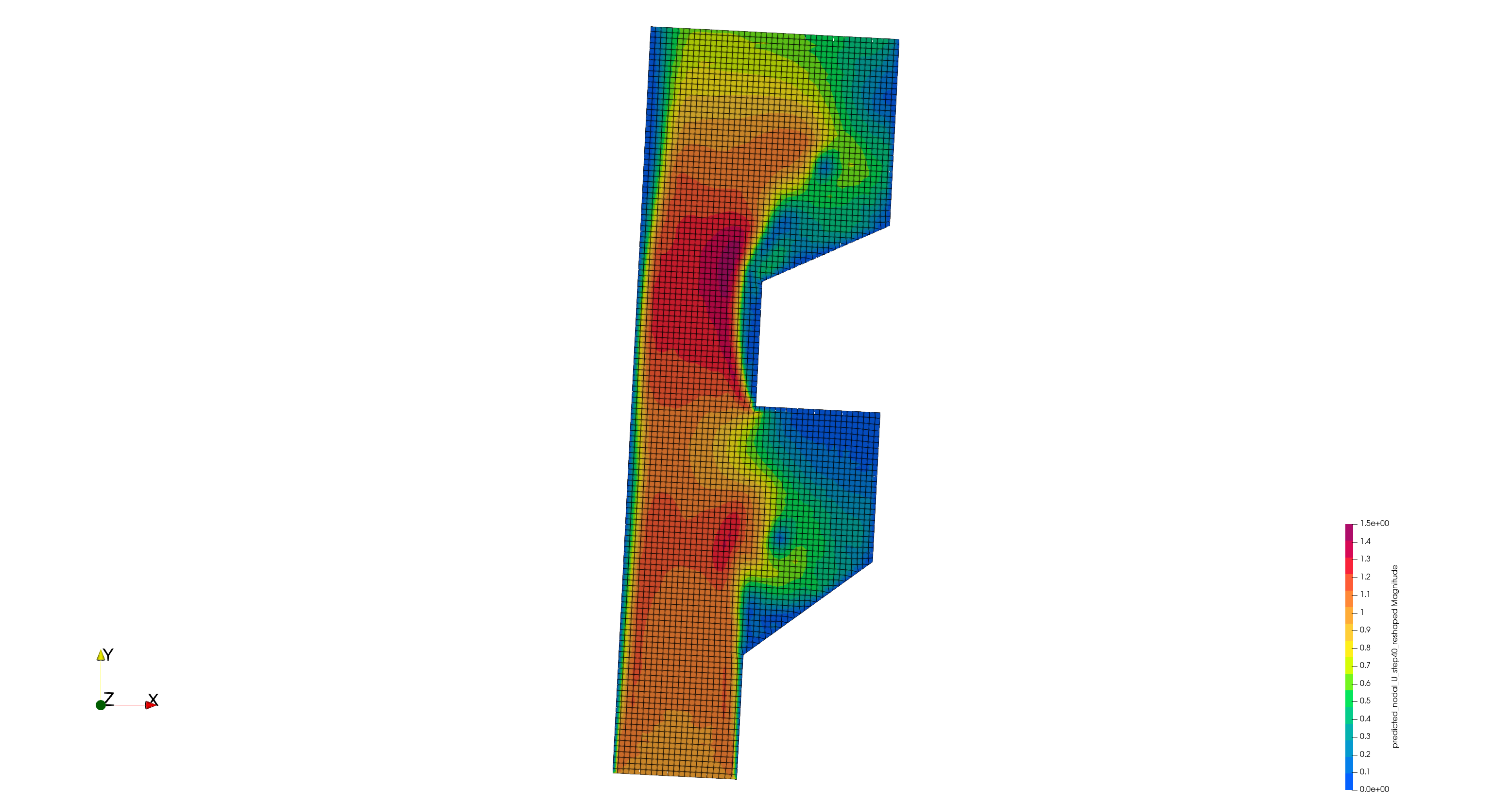}}
  {}
  {\includegraphics[trim={0cm -2cm 0cm 0cm},clip,width=0.1\textwidth]
    {figs/fluid/colorbar_u_16.png}}
  \\[10pt]

  \stackunder[-10pt]
  {\includegraphics[trim={27cm 19cm 27cm 19cm},clip,width=0.21\textwidth]
    {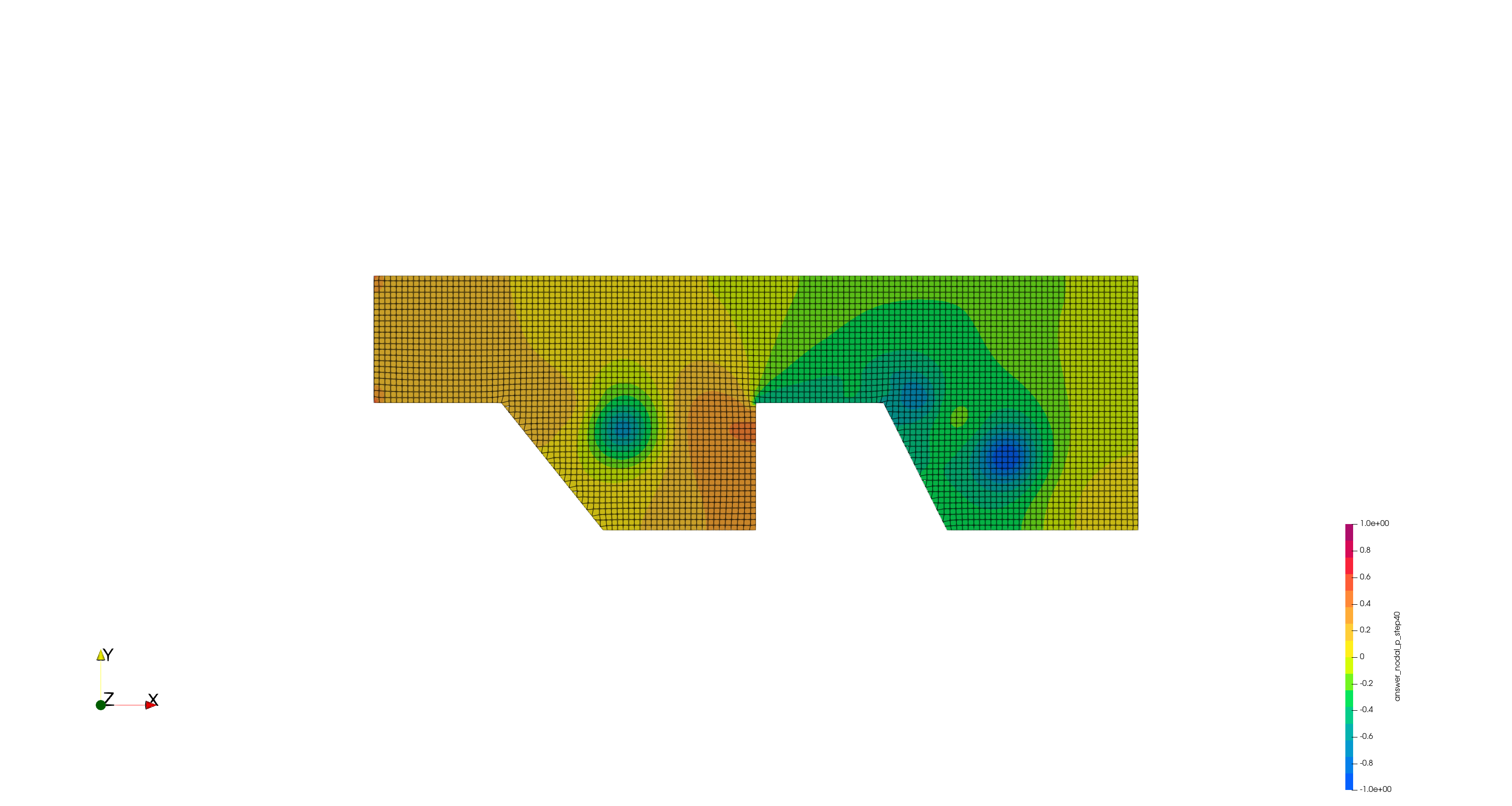}}
    {}
  \hspace{5pt}
  \stackunder[-10pt]
  {\includegraphics[trim={27cm 19cm 27cm 19cm},clip,width=0.21\textwidth]
    {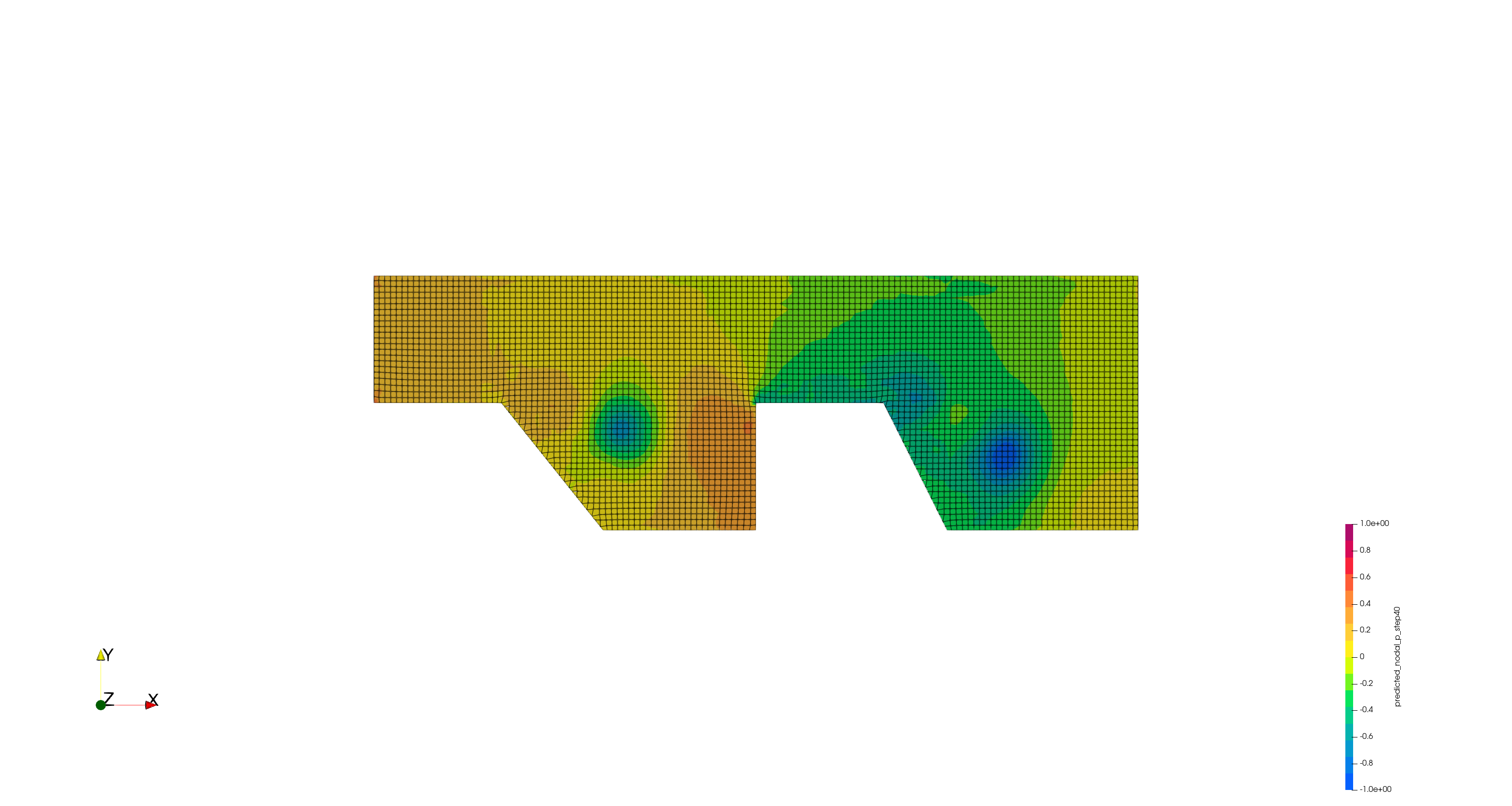}}
    {}
  \hspace{5pt}
  \stackunder[-10pt]
  {\includegraphics[trim={27cm 19cm 27cm 19cm},clip,width=0.21\textwidth]
    {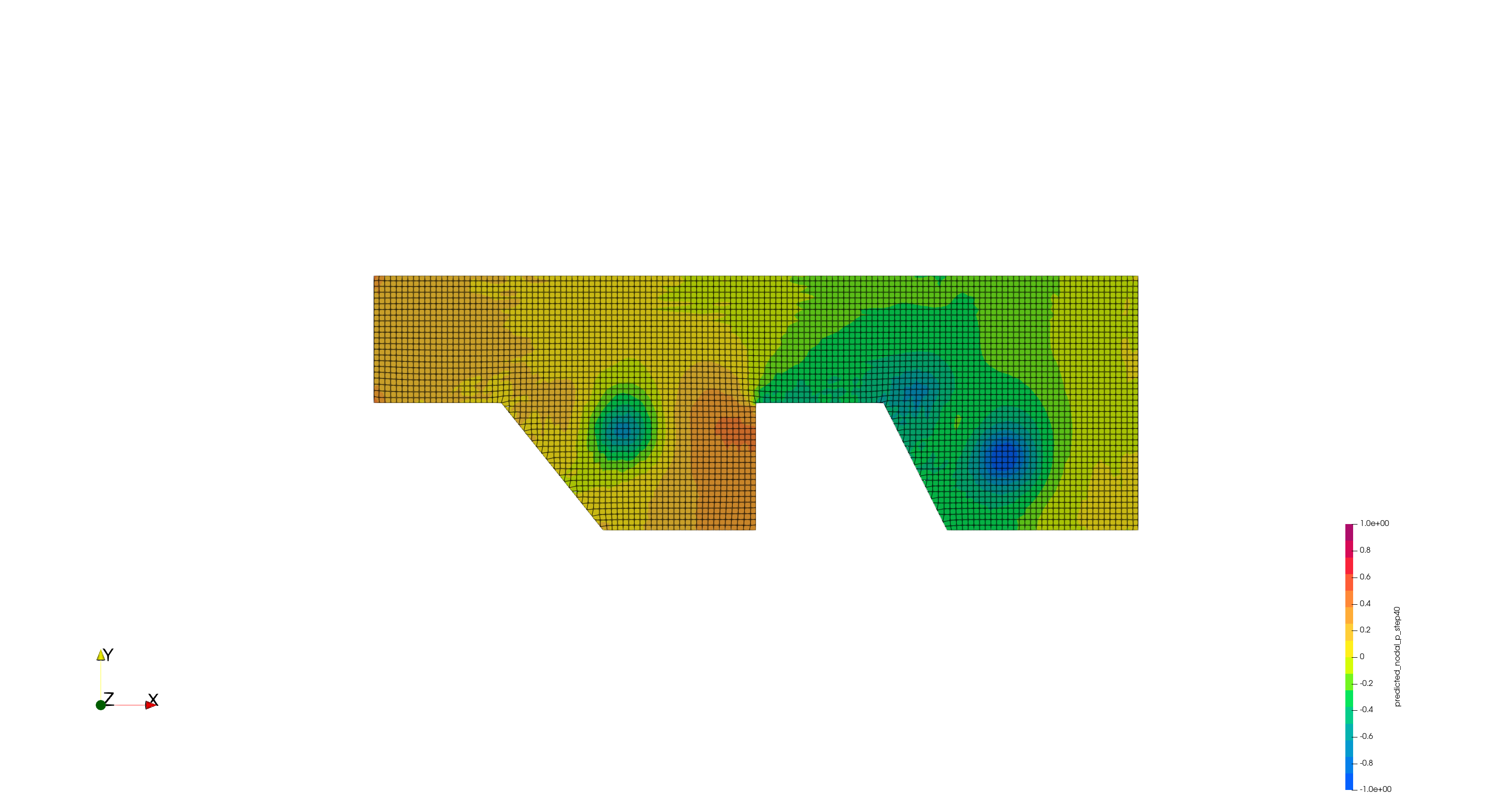}}
    {}
  {\includegraphics[trim={0cm 100cm 0cm 100cm},width=0.1\textwidth]
    {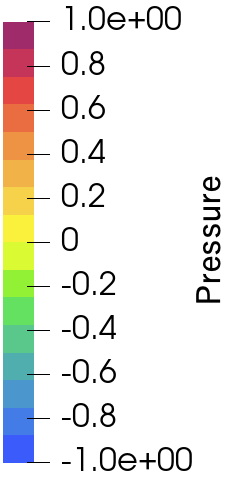}}
  \\[-10pt]

  \stackunder[3pt]
  {\includegraphics[trim={0 0 0 0},clip,width=0.05\textwidth]
    {figs/fluid/coord.png}}
  {}
  \stackunder[0pt]
  {\includegraphics[trim={30cm 0cm 30cm 0cm},clip,width=0.18\textwidth]
    {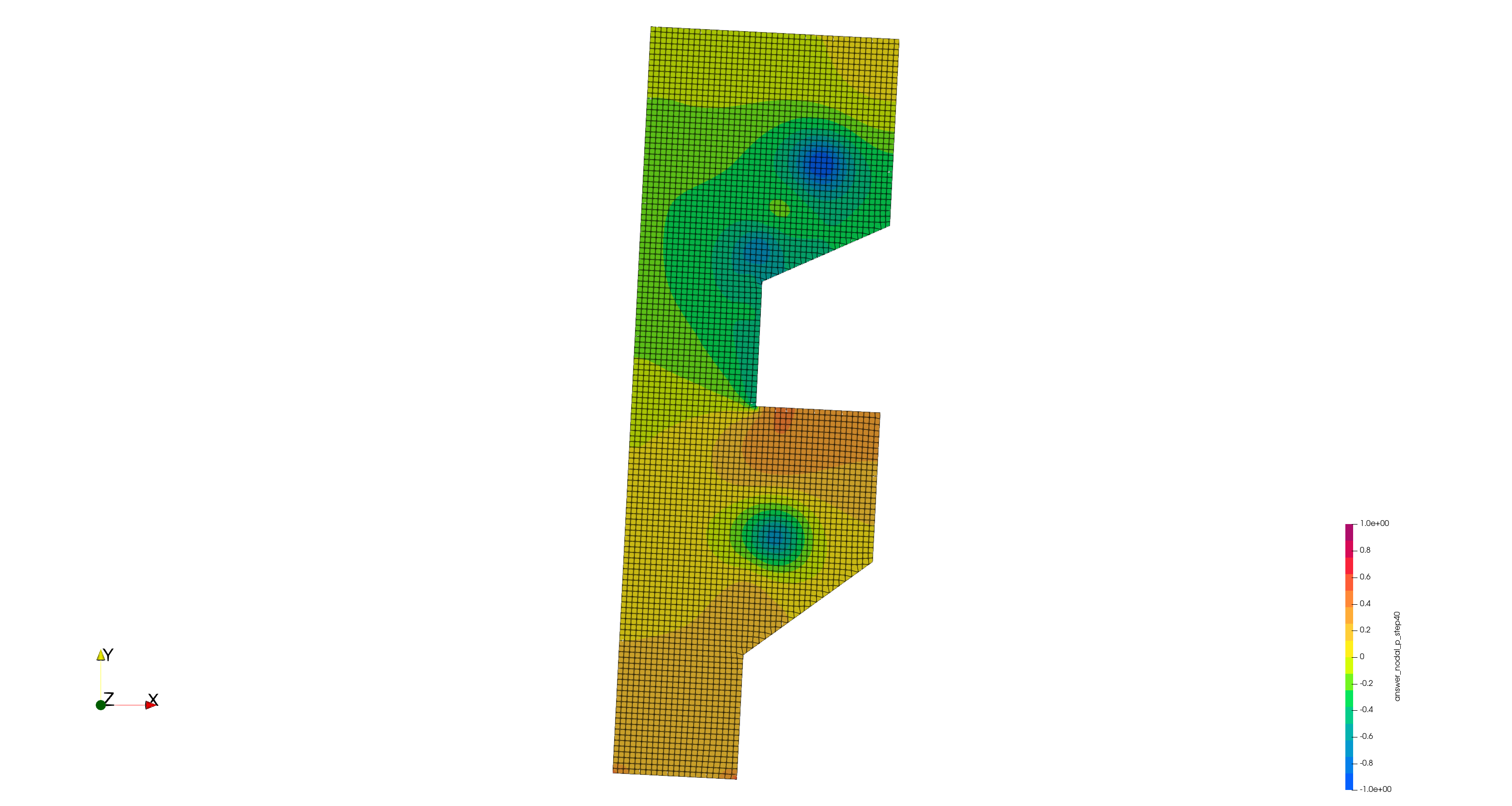}}
  {Ground truth}
  \hspace{15pt}
  \stackunder[0pt]
  {\includegraphics[trim={30cm 0cm 30cm 0cm},clip,width=0.18\textwidth]
    {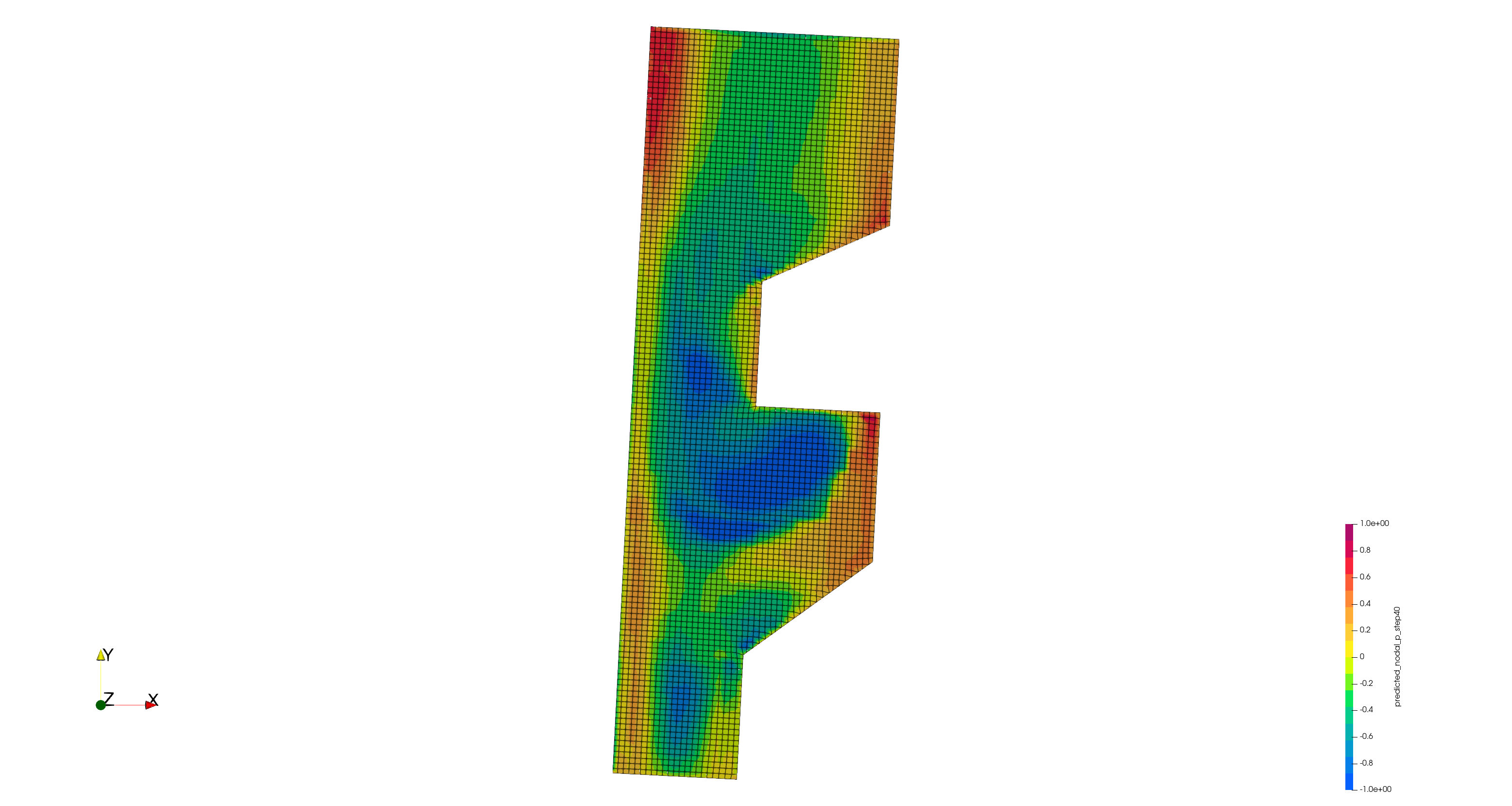}}
  {MP-PDE (TW=20)}
  \hspace{15pt}
  \stackunder[0pt]
  {\includegraphics[trim={30cm 0cm 30cm 0cm},clip,width=0.18\textwidth]
    {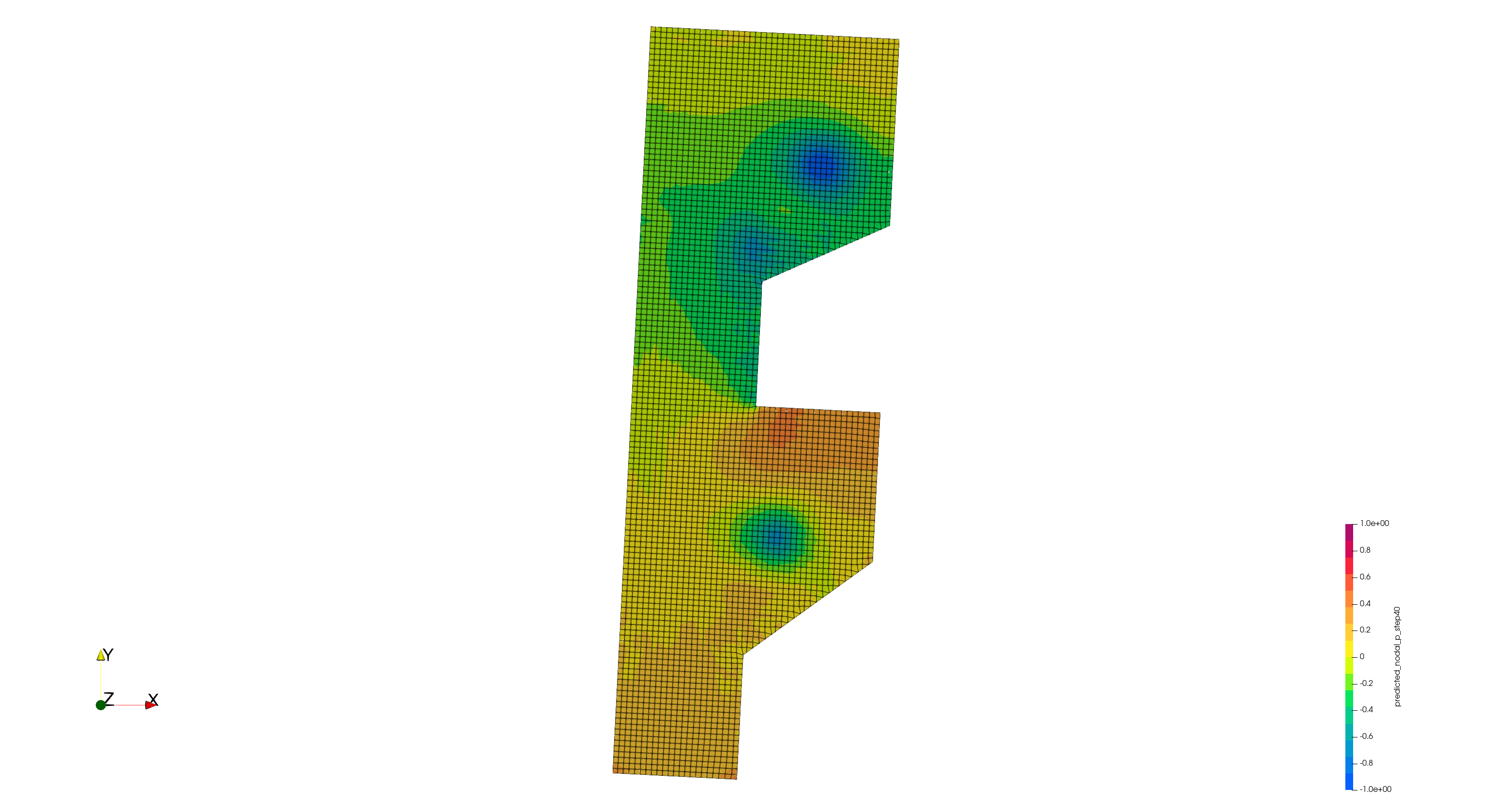}}
  {\makecell{PENN (Ours)}}
  {\includegraphics[trim={0cm -2cm 0cm 0cm},clip,width=0.1\textwidth]
    {figs/fluid/colorbar_p_16.png}}

  \caption{
    Comparison of the velocity field (top two rows) and the pressure field
    (bottom two rows) without (first and third rows) and with (second and fourth rows)
    random rotation and translation.
    PENN prediction is consistent under rotation and translation
    due to the $\mathrm{E}(n)$-equivariance nature of the model,
    while MP-PDE's predictive performance degrades under transformations.
  }
  \label{fig:fluid_results}
\end{figure}

\subsection{Incompressible flow dataset}
We tested the expressive power our model by learning incompressible flow in complex shapes.
The corresponding nonlinear differential operator is denoted as:
\begin{align}
  \mathcal{D}_\mathrm{NS}(\vu) := - (\vu \cdot \nabla) \vu + \frac{1}{\mathrm{Re}} \nabla \cdot \nabla \vu - \nabla p,
  \label{eq:D_NS}
\end{align}
with the incompressible condition
$\nabla \cdot \vu = 0$,
where, in the present case, $\vu$ is the flow velocity field, $p$ is the pressure field, and
$\mathrm{Re}$ is the Reynolds number.

\subsubsection{Data}
To generate the dataset, we first generated pseudo-2D shapes, with one cell in the Z direction, by changing design parameters, starting from three template shapes.
Thereafter, we performed numerical analysis using a classical solver,
OpenFOAM,\footnote{
  \url{https://www.openfoam.com/}} with $\Delta t = 10^{-3}$, and the initial conditions were the solutions of potential flow, which can be computed quickly and stably using the classical solver.
The Reynolds number $\mathrm{Re}$ was around $10^3$.
The linear solvers used were generalized geometric-algebraic multi-grid for
$p$
and the smooth solver with the Gauss--Siedel smoother for
$\vu$.
Template shapes, design parameters, and boundary conditions used can be found in \Appref{app:fluid_dataset}.

To confirm the expressive power of the proposed model, we used coarse input meshes for machine learning models.
We generated these coarse meshes by setting cell sizes roughly four times larger than the original numerical analysis.
We obtained ground truth variables using interpolation.
The task was to predict flow velocity and pressure fields at $t = 4.0$ using information available before numerical analysis, e.g., initial conditions and the geometries of the meshes.
Training, validation, and test datasets consisted of 203, 25, and 25 samples, respectively.
We generated the dataset by randomly rotating and translating test samples to monitor the generalization ability of machine learning models.

\subsubsection{Machine learning models}
We constructed the PENN model corresponding to the incompressible Navier--Stokes equation.
In particular, we adopted the fractional step method, where the pressure field was also obtained as a PDE solution along with the velocity field.
We encoded each feature in a 4, 8, or 16-dimensional space.
After features were encoded, we applied a neural nonlinear solver containing NeumanIsoGCNs and Dirichlet layers, reflecting the fractional step method (See Equations \ref{eq:intermediate_velocity} and \ref{eq:fractional_step}).
Inside the nonlinear solver's loop, we had a subloop that solved the Poisson equation for pressure, which also reflected the considered PDE (See \Eqref{eq:pressure_poisson}).
We looped the solver for pressure five times and four or eight times for velocity.
After these loops stopped, we decoded the hidden features to obtain predictions for velocity and pressure, using the corresponding pseudoinverse decoders.

For the state-of-the-art baseline model, we selected MP-PDE~\citep{brandstetter2022message} as it also provides a way to deal with boundary conditions.
We used the authors' code\footnote{
  \url{https://github.com/brandstetter-johannes/MP-Neural-PDE-Solvers}}
with minimum modification to adapt to the task.
We tested various time window sizes such as 2, 4, 10, and 20, where one step corresponds to time step size
$\Delta t = 0.1$.
With changes in time window size, we changed the number of hops considered in one operation of the GNN of the baseline to have almost the same number of hops visible from the model when predicting the state at
$t = 4.0$.
The numbers of hidden features, 32, 64, and 128, were tested.
All models were trained for up to 24 hours using one GPU (NVIDIA A100 for NVLink 40GiB HBM2).

\subsubsection{Results}
\Tabref{tab:fluid_results} and \Figref{fig:fluid_results} show the comparison between MP-PDE and PENN.
The predictive performances of both models are at almost the same level when evaluated on the original test dataset.
The results show the great expressive power of the MP-PDE model because we kept most settings at default as much as possible and applied no task-specific tuning.
However, when evaluating them on the transformed dataset, the predictive performance of MP-PDE significantly degrades.
Nevertheless, PENN shows the same loss value up to the numerical error, confirming our proposed components are compatible with $\mathrm{E}(n)$-equivariance.
In addition, PENN exhibits no error on the Dirichlet boundaries, showing that our treatment of Dirichlet boundary conditions is rigorous.

\begin{wrapfigure}[20]{r}{0.45\textwidth}
  \centering
  \includegraphics[trim={0cm 0cm 0cm 1cm},clip,width=0.5\textwidth]
  {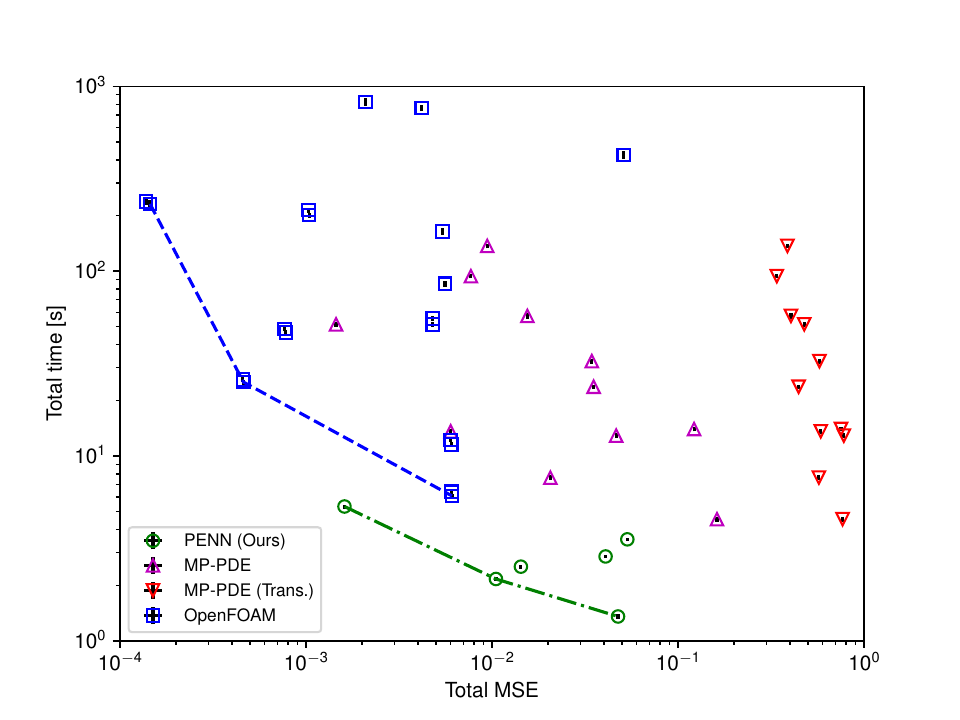}
  \caption{Comparison of computation time and total MSE loss ($\vu$ and $p$)
  on the test dataset (with and without transformation)
  between OpenFOAM, MP-PDE, and PENN.
  The error bar represents the standard error of the mean.
  All computation was done using one core of Intel Xeon CPU E5-2695 v2@2.40GHz.
  Data used to plot this figure are shown in Tables \ref{tab:fluid_compromise_penn}, \ref{tab:fluid_compromise_mppde}, and \ref{tab:fluid_compromise_openfoam}.
  }
  \label{fig:fluid_compromise}
\end{wrapfigure}

\Figref{fig:fluid_compromise} shows the speed-accuracy trade-off for OpenFOAM, MP-PDE, and PENN.
We varied mesh cell size, the time step size, linear sover settings for OpenFOAM
to have different computation speeds and accuracy.
The proposed model achieved the best performance in speed-accuracy trade-off between all the tested methods under fair comparison conditions.

\Tabref{tab:fluid_ablation} presents the results of the ablation study.
Comparison between models with and without the proposed components  shows that the proposed components, i.e., the boundary encoder, Dirichlet layer, pseudoinverse decoder, and neural nonlinear solver, significantly improve the models.
The neural nonlinear solver in the encoded space turned out to have the biggest impact on the performance, while the Dirichlet layer ensured reliable models that strictly respect Dirichlet boundary conditions.

\bgroup
\def\arraystretch{1.3}
\begin{table}[bt]
  \caption{Ablation study on
  the incompressible flow dataset.
  The value represents MSE loss ($\pm$ standard error of the mean) on the
  test dataset.
  "Divergent" means the implicit solver does not converge
  and the loss gets extreme value ($\sim 10^{14}$).
  }
  \centering
  \label{tab:fluid_ablation}
  \scalebox{1.0}{
    \begin{tabular}{lrrrr}
      \\[-8pt]
      \toprule
      Method
      & \makecell{$\vu$\\$(\times 10^{-4})$}
      & \makecell{$p$\\$(\times 10^{-3})$}
      & \makecell{$\hat{\vu}_\mathrm{Dirichlet}$\\$(\times 10^{-4})$}
      & \makecell{$\hat{p}_\mathrm{Dirichlet}$\\$(\times 10^{-3})$}
      \\
      \hline
      Without encoded boundary &
      Divergent &
      Divergent &
      Divergent &
      Divergent
      \\[3pt]
      \makecell[l]{
        Without boundary condition
        \\
        in the neural nonlinear solver
      } &
$65.10 \pm 0.38$ &
$21.70 \pm 0.09$ &
$0.00 \pm 0.00$ &
$0.00 \pm 0.00$
      \\[3pt]
      Without neural nonlinear solver &
$31.03 \pm 0.19$ &
$9.81 \pm 0.04$ &
$\boldsymbol{0.00} \pm 0.00$ &
$\boldsymbol{0.00} \pm 0.00$
      \\[3pt]
      Without boundary condition input &
$20.08 \pm 0.21$ &
$3.61 \pm 0.02$ &
$59.60 \pm 0.89$ &
$1.43 \pm 0.05$
      \\[3pt]
      Without Dirichlet layer &
$8.22 \pm 0.07$ &
$1.41 \pm 0.01$ &
$18.20 \pm 0.28$ &
$0.38 \pm 0.01$
      \\[3pt]
      Without pseudoinverse decoder
      &
$8.91 \pm 0.06$ &
$2.36 \pm 0.02$ &
$1.97 \pm 0.06$ &
$\boldsymbol{0.00} \pm 0.00$
      \\[3pt]
      \makecell[l]{
        Without pseudoinverse decoder
        \\
        with Dirichlet layer after decoding
      } &
$6.65 \pm 0.05$ &
$1.71 \pm 0.01$ &
$\boldsymbol{0.00} \pm 0.00$ &
$\boldsymbol{0.00} \pm 0.00$
      \\[5pt]
      \textbf{PENN} &
      $\boldsymbol{4.36} \pm 0.03$ &
      $\boldsymbol{1.17} \pm 0.01$ &
      $\boldsymbol{0.00} \pm 0.00$ &
      $\boldsymbol{0.00} \pm 0.00$
      \\
      \bottomrule
    \end{tabular}
  }
\end{table}
\egroup

\section{Conclusion}
\label{sec:conclusion}
We have presented an $\mathrm{E}(n)$-equivariant, GNN-based neural PDE solver, PENN, which can fulfill boundary conditions required for reliable predictions.
The model has superiority in embedding the information of PDEs (physics) in the model and speed-accuracy trade-off.
Therefore, our model can be a useful standard for realizing reliable, fast, and accurate GNN-based PDE solvers.
Although the property of our model is preferable, it also limits the applicable domain of the model
because we need to be familiar with the concrete form of the PDE of interest to construct the effective PENN model.
For instance, the proposed model cannot exploit its potential to solve inverse problems where explicit forms of the governing PDE are not available for such tasks.
Therefore, combining PINNs and PENNs could be the next direction of the research community.

\section{Potential negative societal impacts}
\label{sec:negative}
We have built a foundation to learn PDEs in a steerable manner rather than focusing on a specific application.
Because of that, we envisage minimal risk of direct abusing our present work.
However, as mentioned in the introduction, solving PDEs has many impacts on various domains, from both positive and negative aspects.
Thus, our work and possible successive ones may be abused, aiming to harm lives and the environment.
Therefore, the research community, including us, must be careful in using them and control the research direction to prevent abusing these technologies.

\begin{ack}
  This work was supported by PRESTO, Japan Science and Technology Agency, Grant Number JPMJPR21O9, Japan and by the New Energy and Industrial Technology Development Organization, Grant Number JPNP14012, Japan.
\end{ack}

\small{
\bibliographystyle{penn}
\bibliography{penn}
}

\medskip

\section*{Checklist}

\begin{enumerate}

\item For all authors...
\begin{enumerate}
  \item Do the main claims made in the abstract and introduction accurately reflect the paper's contributions and scope?
    \answerYes{See the abstract and \Secref{sec:introduction}.}
  \item Did you describe the limitations of your work?
    \answerYes{See \Secref{sec:conclusion}.}
  \item Did you discuss any potential negative societal impacts of your work?
    \answerYes{See \Secref{sec:negative}.}
  \item Have you read the ethics review guidelines and ensured that your paper conforms to them?
    \answerYes{Our paper, in particular \Secref{sec:negative}, is written in accordance with the guideline.}
\end{enumerate}

\item If you are including theoretical results...
\begin{enumerate}
  \item Did you state the full set of assumptions of all theoretical results?
    \answerNA{}
        \item Did you include complete proofs of all theoretical results?
    \answerNA{}
\end{enumerate}

\item If you ran experiments...
\begin{enumerate}
  \item Did you include the code, data, and instructions needed to reproduce the main experimental results (either in the supplemental material or as a URL)?
    \answerYes{See Footnote~\ref{footnote:penn}.}
  \item Did you specify all the training details (e.g., data splits, hyperparameters, how they were chosen)?
    \answerYes{See \Appref{app:gradient}, \ref{app:fluid}, \ref{app:ad}, and Footnote~\ref{footnote:penn}.}
  \item Did you report error bars (e.g., with respect to the random seed after running experiments multiple times)?
    \answerYes{See \Secref{sec:experiments}, \Tabref{tab:gradient}, \ref{tab:fluid_results}, \ref{tab:fluid_ablation}, and \Figref{fig:fluid_compromise}.}
  \item Did you include the total amount of compute and the type of resources used (e.g., type of GPUs, internal cluster, or cloud provider)?
    \answerYes{See \Secref{sec:experiments}.}
\end{enumerate}

\item If you are using existing assets (e.g., code, data, models) or curating/releasing new assets...
\begin{enumerate}
  \item If your work uses existing assets, did you cite the creators?
    \answerYes{See Footnote~\ref{footnote:isogcn}}
  \item Did you mention the license of the assets?
    \answerYes{See Footnote~\ref{footnote:isogcn}}
  \item Did you include any new assets either in the supplemental material or as a URL?
    \answerYes{See \Secref{sec:experiments}, \Tabref{tab:gradient}, \ref{tab:fluid_results}, \ref{tab:fluid_ablation}, and \Figref{fig:fluid_compromise}.}
  \item Did you discuss whether and how consent was obtained from people whose data you're using/curating?
    \answerNA{The dataset we used in the paper is generated by ourselves.}
  \item Did you discuss whether the data you are using/curating contains personally identifiable information or offensive content?
    \answerNA{The dataset is based on numerical analysis, so there is no concern about personally identifiable information nor offensive content.}
\end{enumerate}

\item If you used crowdsourcing or conducted research with human subjects...
\begin{enumerate}
  \item Did you include the full text of instructions given to participants and screenshots, if applicable?
    \answerNA{}
  \item Did you describe any potential participant risks, with links to Institutional Review Board (IRB) approvals, if applicable?
    \answerNA{}
  \item Did you include the estimated hourly wage paid to participants and the total amount spent on participant compensation?
    \answerNA{}
\end{enumerate}

\end{enumerate}

\clearpage
\appendix

\section{Details of the proposed method}
\label{app:method}

\subsection{Construction of pseudoinverse decoder}
\label{app:pseudoinverse}
We can construct the pseudoinverse decoders for a wide range of neural network architectures.
For instance, the pseudoinverse decoder for an multilayer perceptron (MLP) with one hidden layer
$f(x) = \sigma_2\left(W_2 \sigma_1(W_1 x + b_1) + b_2\right)$
can be constructed as:
\begin{align}
  f^+(h) = W^+_1 \sigma^{-1}_1\left(W^+_2 \sigma^{-1}_2(h) - b_2\right) - b_1,
\end{align}
where
$W^+$
is the pseudoinverse matrix of
$W$
and
$\sigma$
is an invertible activation function whose
$\mathrm{Dom}(\sigma) = \mathrm{Im}(\sigma) = \mathbb{R}$.
We chose LeakyReLU
\begin{align}
  \mathrm{LeakyReLU}(x) =
  \left\{
    \begin{array}{cc}
      x & (x \geq 0)
      \\
      a x & (x < 0),
    \end{array}
  \right.
\end{align}
where set
$a = 0.5$
because an extreme value of
$a$ (e.g., 0.01) could lead to an extreme value of gradient for the inverse function.
In addition, one may choose activation functions whose
$\mathrm{Im}(\sigma) \neq \mathbb{R}$,
such as $\tanh$.
However, in that case, we must ensure that the input value to the pseudoinverse decoder is in
$\mathrm{Im}(\sigma)$
(in case of $\tanh$, it is $(-1, 1)$);
otherwise, the computation would be invalid.

Besides, similar to the Dirichlet encoder and pseudoinverse decoder, we could define the specific encoder and decoder for the Neumann boundary condition.
However, this is not included in the contributions of our work because it does not improve the performance of our model, which may be because the Neumann boundary condition is a soft constraint in contrast to the Dirichlet one and expressive power seems more important than that inductive bias.

\subsection{Derivation of NIsoGCN}
\label{app:NIsoGCN}
\citet{matsunaga2020improved} derived a gradient model that can treat the Neumann boundary condition with an arbitrary convergence rate with regard to spatial resolution.
Here, we derive our gradient model, i.e., NIsoGCN, in a different way to simplify the discussion because we only need the first-order approximation for fast computation.

Before deriving NIsoGCN, we review introductory linear algebra using simple normation.
Using a unit basis
$\{\ve_i \in \mathbb{R}^d : \Vert\ve_i\Vert = 1\}_{i=1}^d$,
one can decompose a vector $\vv \in \mathbb{R}^d$ using:
\begin{align}
\vv = \sum_i (\vv \cdot \ve_i) \ve_i.
\end{align}
Now, consider replacing the basis
$\{\ve_i \in \mathbb{R}^d\}_{i=1}^d$
with a set of vectors
$\mB = \{\vb_i \in \mathbb{R}^d\}_{i=1}^{d'}$,
called a \textit{frame},
that spans the space but is not necessarily independent (thus, $d' \geq d$).
Using the frame, one can assume $\vv$ is decomposed as:
\begin{align}
  \vv = \sum_i(\vv \cdot \vb_i) \mA \vb_i,
\end{align}
where
$\mA$
is a matrix that corrects the "overcount" that may occur using the frame
(for instance, consider expanding
$(1, 0)^\top$
with
the frame $\{(1, 0)^\top, (-1, 0)^\top, (0, 1)^\top\}$).
A set $\{\mA \vb_i\}_{i=0}^{d'}$
is called a \textit{dual frame} for
$\mB$.
We can find the concrete form of
$\mA$
considering:
\begin{align}
  \vv &= \mA \sum_i(\vv \cdot \vb_i) \vb_i
  \\
  &= \mA \sum_i (\vb_i \otimes \vb_i) \vv.
  \label{eq:reconstruction}
\end{align}
Requiring that \Eqref{eq:reconstruction} holds for any $\vv \in \mathbb{R}^d$,
one can conclude
$\mA = \sum_i(\vb_i \otimes \vb_i)^{-1}$.
Finally, we obtain
\begin{align}
  \vv &= \left[\vb_i \otimes \vb_i\right]^{-1} \sum_i(\vv \cdot \vb_i) \vb_i
\end{align}
For more details on frames, see, e.g., \citet{han2007frames}.

Then, we can derive NIsoGCN at the $i$th vertex on the Neumann boundary,
by letting
\begin{align}
  \mB &= \left\{
    \frac{\vx_{j_1} - \vx_i}{\Vert\vx_{j_1} - \vx_i\Vert},
    \frac{\vx_{j_2} - \vx_i}{\Vert\vx_{j_2} - \vx_i\Vert},
    \dots,
    \frac{\vx_{j_m} - \vx_i}{\Vert\vx_{j_m} - \vx_i\Vert},
    \sqrt{w_i} \vn_i
  \right\},
  \label{eq:frame_NIsoGCN}
\end{align}
where
$\{j_1, j_2, \dots, j_m\}$
are indices of neighboring vertices to the $i$th vertex.
In addition, we assume the approximated gradient of a scalar field $\psi$ at the $i$th vertex,
$\left< \nabla \psi \right>_i$,
satisfies the following conditions:
\begin{align}
  \left< \nabla \psi \right>_i \cdot
    \frac{\vx_{j_k} - \vx_i}{\Vert\vx_{j_k} - \vx_i\Vert}
    &= \frac{\psi_{j_k} - \psi_i}{\Vert\vx_{j_k} - \vx_i\Vert},
    &(k = 1, \dots, m),
    \label{eq:slope}
  \\
  \left< \nabla \psi \right>_i \cdot \vn &= \hat{g}_i. &
    \label{eq:directional_Neumann}
\end{align}
\Eqref{eq:slope} is a natural assumption because we expect the directional derivative in the direction of
$(\vx_{j_k} - \vx_i) / \Vert\vx_{j_k} - \vx_i\Vert$
should correspond to the slope of
$\psi$ in the same direction.
\Eqref{eq:directional_Neumann} is the Neumann boundary condition, which we want to satisfy.
Finally, by substituting \Eqsref{eq:frame_NIsoGCN},~\ref{eq:slope},
and~\ref{eq:directional_Neumann}, we obtain NIsoGCN, i.e.,
\Eqref{eq:NIsoGCN}.

To apply NIsoGCN to $\vt$, the rank $k$ tensors ($k \geq 1$), one can recursively define the operation as:
\begin{align}
  \mathrm{NIsoGCN}_{k \to k+1}(\vt)
  := \left(
    \begin{array}{c}
      \mathrm{NIsoGCN}_{k-1 \to k}(\vt_{1})
      \\
      \mathrm{NIsoGCN}_{k-1 \to k}(\vt_{2})
      \\
      \mathrm{NIsoGCN}_{k-1 \to k}(\vt_{3})
      \\
    \end{array}
  \right),
\end{align}
where
$\vt_{i}$ is the $i$th component of $\vt$, resulting in the rank $(k-1)$ tensor.
In case of the rank 1 tensor $\vv$, it can be formulated as:
\begin{align}
  \mathrm{NIsoGCN}_{1 \to 2}(\vv)
  := \left(
    \begin{array}{c}
      \mathrm{NIsoGCN}_{0 \to 1}(v_{1})
      \\
      \mathrm{NIsoGCN}_{0 \to 1}(v_{2})
      \\
      \mathrm{NIsoGCN}_{0 \to 1}(v_{3})
      \\
    \end{array}
  \right)
  \approx
  \left(
    \begin{array}{ccc}
      \inpdiff{v_1}{x} &
      \inpdiff{v_1}{y} &
      \inpdiff{v_1}{z}
      \\
      \inpdiff{v_2}{x} &
      \inpdiff{v_2}{y} &
      \inpdiff{v_2}{z}
      \\
      \inpdiff{v_3}{x} &
      \inpdiff{v_3}{y} &
      \inpdiff{v_3}{z}
    \end{array}
  \right)
  = \nabla \vv
  .
\end{align}
Please note that each component $v_i$ has multiple features in the encoded space, e.g., 16 or 64,
resulting in
$\mathrm{NIsoGCN}_{1 \to 2}(\vv)$
represents multiple rank 2 tensors for each vertex (see Figure 1 of \citet{horie2021isometric}).

As discussed in \citet{horie2021isometric}, IsoGCNs (NIsoGCNs) correspond to
spatial differential operators as:
\begin{itemize}
  \item
    $\mathrm{NIsoGCN}_{0 \to 1}(\psi)$: Gradient $\nabla \psi$ (rank 0 tensor to rank 1 tensor)
  \item
    $\mathrm{NIsoGCN}_{1 \to 0}(\vv)$: Divergence $\nabla \cdot \vv$ (rank 1 tensor to rank 0 tensor)
  \item
    $\mathrm{NIsoGCN}_{0 \to 1 \to 0}(\psi) := \mathrm{NIsoGCN}_{1 \to 0} \circ \mathrm{NIsoGCN}_{0 \to 1} (\psi)$: Laplacian $\nabla \cdot \nabla \psi$ (rank 0 tensor to rank 1 tensor to rank 0 tensor)
  \item
    $\mathrm{NIsoGCN}_{1 \to 2}(\vv)$: Jacobian $\nabla \vv$ (rank 1 tensor to rank 2 tensor)
  \item
    $\mathrm{NIsoGCN}_{0 \to 1 \to 2}(\psi) := \mathrm{NIsoGCN}_{l \to 2} \circ \mathrm{NIsoGCN}_{0 \to 1} (\psi)$: Hessian $\nabla \nabla \psi$ (rank 0 tensor to rank 1 tensor to rank 2 tensor)
\end{itemize}
Because NIsoGCN contains a learnable weight matrix (see \Eqref{eq:NIsoGCN}), the component learns to predict the derivative of the corresponding tensor rank in an encoded space. This feature of NIsoGCNs enables us to construct machine learning models corresponding to PDE in the encoded space.

\subsection{Derivation of the step size in the Barzilai--Borwein method}
\label{app:BB}
We derive \Eqref{eq:BB} by applying the Barzilai--Borwein method to our case.
We start with \Eqref{eq:optimization}, which corresponds to a nonlinear problem:
\begin{align}
  \vR(\vv) &:= \vv - \vu(t, \cdot) - \mathcal{D}(\vv) \Delta t,
  \\
  \mathrm{Solve}_\vv \ \vR(\vv)(\vx_i) &= \bm{0}, \ \forall i,
\end{align}
We consider solving it by applying the linear iterative method using the Taylor expansion, assuming the update
$\Delta \vv\supp{i} := \vv\supp{i+1} - \vv\supp{i}$
is small enough.
The iterative method is expressed as:
\begin{align}
\vv\supp{0} &= \vu(t, \cdot),
\\
\vv\supp{i+1} &= \vv\supp{i} + \Delta \vv\supp{i},
\\
\mR(\vv\supp{i} + \Delta \vv\supp{i})
&\approx \mR(\vv\supp{i}) + \nabla_\vv \mR(\vv\supp{i})\Delta\vv\supp{i} = \bm{0},
  \label{eq:linear_update}
\end{align}
where
$\nabla_\vv \mR(\vv\supp{i})$
denotes the Jacobian matrix with the shape of $n \times n$ ($n$ roughly corresponds to the number of vertices of the mesh).
To optain update, we may solve \Eqref{eq:linear_update} as:
\begin{align}
  \Delta \vv\supp{i} = \left[ \nabla_\vv \vR(\vv\supp{i}) \right]^{-1} \vR(\vv\supp{i}),
\end{align}
corresponding to the Newton--Raphson method.
However, it may take enormous computation resources because
$\nabla_\vv \mR(\vv\supp{i})$
is usually a huge matrix.
Instead, we can approximate:
\begin{align}
  \left[ \nabla_\vv \vR(\vv\supp{i}) \right]^{-1} \approx \alpha\supp{i},
  \label{eq:approx_alpha}
\end{align}
which corresponds to gradient descent:
\begin{align}
  \Delta \vv\supp{i} \approx \alpha\supp{i} \vR(\vv\supp{i}).
\end{align}
Substituting \Eqref{eq:approx_alpha} into \Eqref{eq:linear_update}, we obtain:
\begin{align}
\mR(\vv\supp{i+1})
&\approx
  \mR(\vv\supp{i}) + \frac{1}{\alpha\supp{i}}\Delta\vv\supp{i}.
  \label{eq:approx_alpha2}
\end{align}
We want to find a good $\alpha\supp{i}$ satisfying \Eqref{eq:approx_alpha2} the best.
Thus, we obtain $\alpha_\mathrm{BB}\supp{i}$ through:
\begin{align}
  \alpha\supp{i}_\mathrm{BB}
&=
\argmin_\alpha \mathcal{L}(\alpha),
\\
  \mathbb{R} \ni \mathcal{L}(\alpha)
:&=
\frac{1}{2} \left\Vert
\Delta\vv\supp{i}
- \alpha\Delta\mR\supp{i}
\right\Vert^2, \ \mathrm{where} \ \Delta \mR\supp{i} = \mR(\vv\supp{i+1}) - \mR(\vv\supp{i}).
\end{align}
Because of the convexity of the problem, it is enough to find alpha satisfying:
\begin{align}
  \left.\diff{\mathcal{L}}{\alpha}\right|_{\alpha\supp{i}_\mathrm{BB}}
  &= \left<\Delta \vv\supp{i} - \alpha_\mathrm{BB}\supp{i} \Delta \mR\supp{i}, -\Delta \mR\supp{i} \right> = 0,
\end{align}
where $<\cdot, \cdot>$ denotes the inner product in the corresponding space.
Using the linearity of the inner product, we obtain:
\begin{align}
  \left<\Delta \vv\supp{i} - \alpha_\mathrm{BB}\supp{i} \Delta \mR\supp{i}, -\Delta \mR\supp{i} \right> &= 0,
  \\
  - \left<\Delta \vv\supp{i}, \Delta \mR\supp{i}\right>
  + \alpha_\mathrm{BB}\supp{i} \left< \Delta \mR\supp{i}, \Delta \mR\supp{i} \right> &= 0,
  \\
  \alpha_\mathrm{BB}\supp{i} &= \frac{\left<\Delta \vv\supp{i}, \Delta \mR\supp{i}\right>}{\left< \Delta \mR\supp{i}, \Delta \mR\supp{i} \right>}.
  \label{eq:BB_derivation}
\end{align}
\Eqref{eq:BB_derivation} is equivalent to \Eqref{eq:BB}.

As seen from the derivation, $\alpha\supp{i}_\mathrm{BB}$ is determined to satisfy \Eqref{eq:approx_alpha2} as much as possible for all vertices and all feature components.
That means $\alpha\supp{i}_\mathrm{BB}$ has global information because it considers all vertices, making the global interaction possible.
In addition, $\alpha\supp{i}_\mathrm{BB}$ is equivariant because it is scalar, which does not depend on coordinate.
Therefore, $\alpha\supp{i}_\mathrm{BB}$ is suitable for realizing efficient PDE solvers with
$\mathrm{E}(n)$-equivariance.

\section{Experiment details: gradient dataset}
\label{app:gradient}

\begin{figure}[t]
  \centering
  \includegraphics[trim={0cm 4cm 1cm 2cm},clip,width=0.99\linewidth]
  {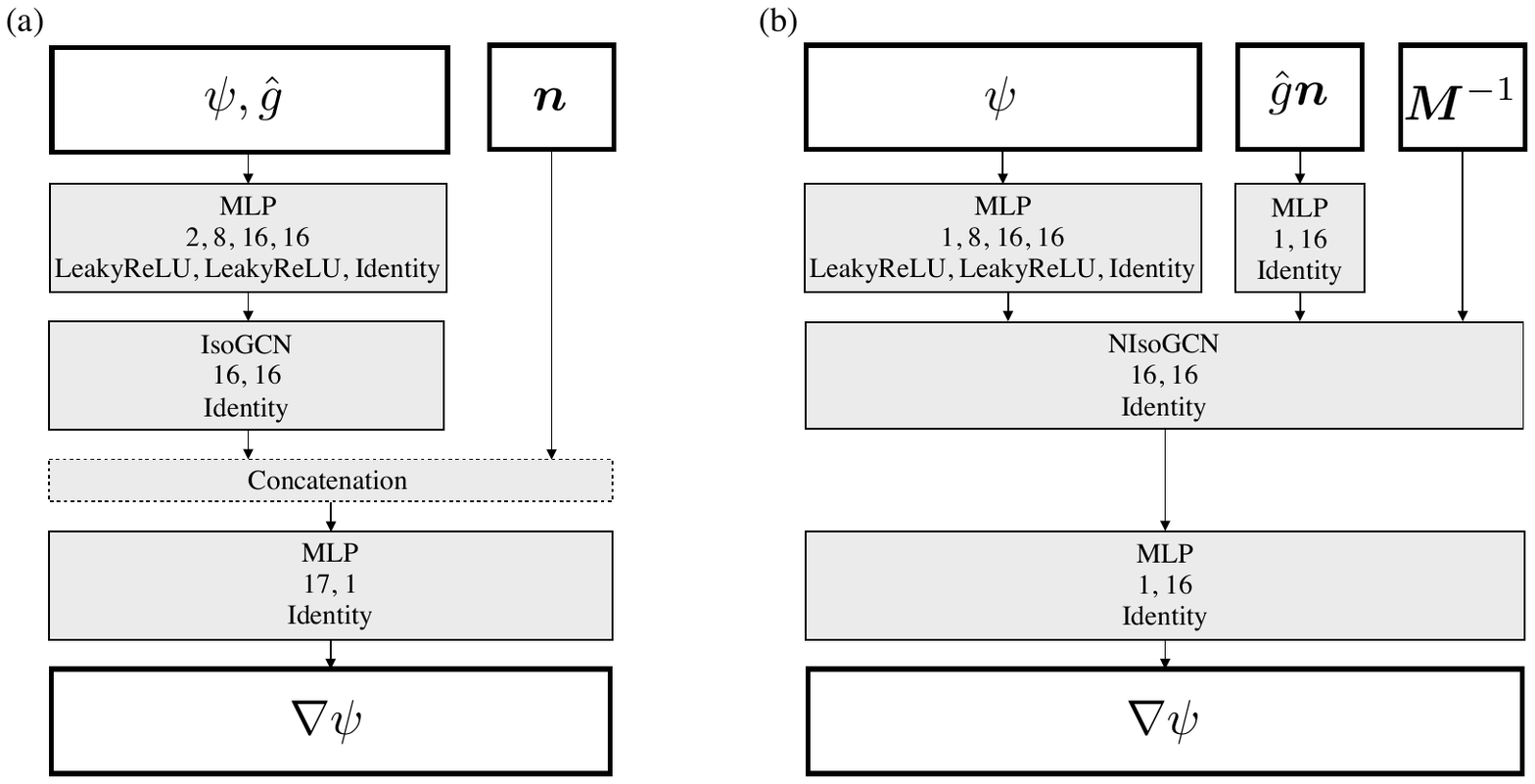}
  \caption{Architecture used for
  (a) original IsoGCN and
  (b) NIsoGCN
  training.
  In each cell, we put the number of units in each layer along with the activation functions used.
  }
  \label{fig:grad_architecture}
\end{figure}

\Figref{fig:grad_architecture} shows the architectures we used for the gradient dataset.
The dataset is uploaded online.\footnote{\url{https://savanna.ritc.jp/~horiem/penn_neurips2022/data/grad/grad_data.tar.gz}}
We followed the instruction of \citet{horie2021isometric} (in particular, Appendix D.1 of their paper) to make the features and models equivariant.
To facilitate a fair comparison, we made input information for both models equivalent, except for
$\mM^{-1}$
in Equation \Eqref{eq:moment_NIsoGCN}, which is a part of our novelty.
For both models, we used Adam~\citep{kingma2014adam} as an optimizer with the default setting.
Training for both models took around ten minutes using one GPU (NVIDIA A100 for NVLink 40GiB HBM2).
\Figref{fig:grad_architecture} shows model architectures used for the experiment.

\section{Experiment details: incompressible flow dataset}
\label{app:fluid}
\subsection{Governing equation}
The incompressible Navier--Stokes equations, the governing equations of
incompressible flow, are expressed as:
\begin{align}
  \pdiff{\vu}{t}
  &= - (\vu \cdot \nabla) \vu
  + \frac{1}{\mathrm{Re}} \nabla \cdot \nabla \vu
  - \nabla p
  & (t, \vx) \in (0, T) \times \Omega,
  \\
  \vu &= \hat{\vu}
  & (t, \vx) \in \partial\Omega_\mathrm{Dirichlet}^{(\vu)},
  \\
  \left[\nabla \vu + (\nabla \vu)^T\right]\vn &= \bm{0}
  & (t, \vx) \in \partial\Omega_\mathrm{Neumann}^{(\vu)}.
\end{align}
We also consider the following incompressible condition:
\begin{align}
  \nabla \cdot \vu = 0 \ \ \ \ \ \ \ \ (t, \vx) \in (0, T) \times \Omega,
\end{align}
which may be problematic when solving these equations numerically.
Therefore, it is common to divide the equations into two: one to obtain pressure and one to compute velocity.
There are many methods to make such a division;
for instance, the fractional step method derives the Poisson equation for pressure as follows:
\begin{align}
  \nabla \cdot \nabla p(t + \Delta t, \vx) = \frac{1}{\Delta t} (\nabla \cdot \tilde{\vu})(t, \vx),
  \label{eq:pressure_poisson}
\end{align}
where
\begin{align}
  \tilde{\vu} = \vu - \Delta t \left( \vu \cdot \nabla \vu - \frac{1}{\mathrm{Re}} \nabla \cdot \nabla \vu \right)
  \label{eq:intermediate_velocity}
\end{align}
is called the intermediate velocity.
Once we solve the equation, we can compute the time evolution of velocity as follows:
\begin{align}
  \vu(t + \Delta t, \vx) = \tilde{\vu}(t, \vx) - \Delta t \nabla p(t + \Delta t, \vx).
  \label{eq:fractional_step}
\end{align}

Because the fractional step method requires solving the Poisson equation for pressure, we also need the boundary conditions for pressure as well:
\begin{align}
  p &= 0
  & (t, \vx) \in \partial\Omega_\mathrm{Dirichlet}^{(p)},
  \\
  \nabla p \cdot \vn &= 0
  & (t, \vx) \in \partial\Omega_\mathrm{Neumann}^{(p)}.
\end{align}
Our machine learning task is also based on the same assumption: motivating pressure prediction in addition to velocity with boundary conditions of both.

\begin{figure}[t]
  \centering
  \includegraphics[trim={0cm 3cm 0cm 2cm},clip,width=0.99\linewidth]
  {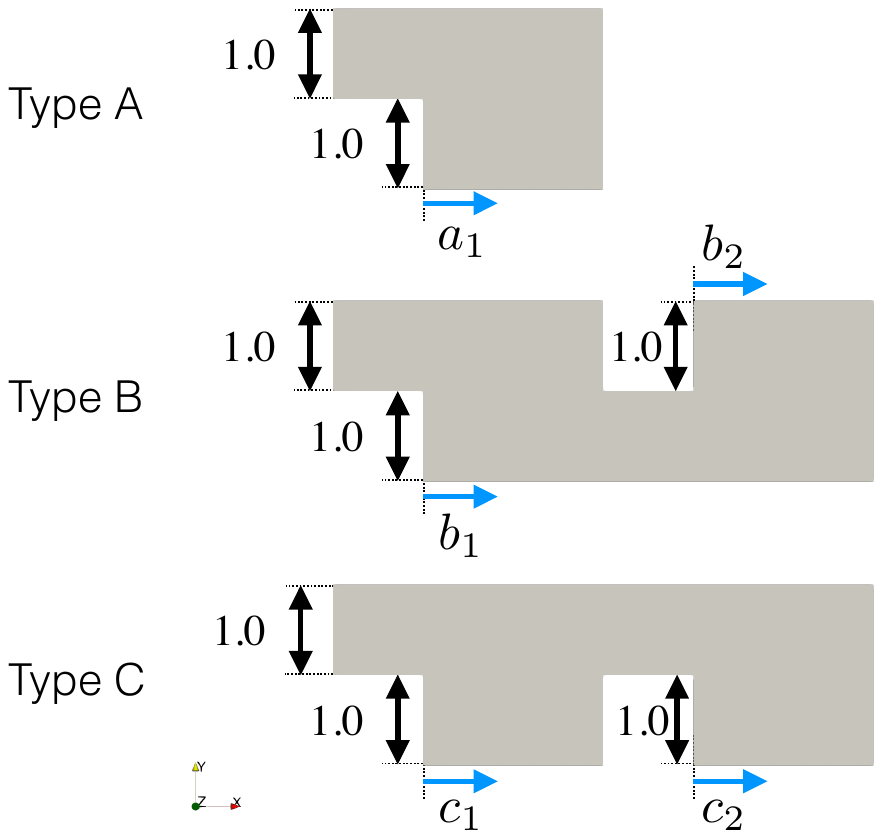}
  \caption{Three template shapes used to generate the dataset.
  $a_1$, $b_1$, $b_2$, $c_1$, and $c_2$ are the design parameters.
  }
  \label{fig:fluid_shapes}
\end{figure}
\begin{figure}[t]
  \centering
  \includegraphics[trim={0cm 4cm 0cm 2cm},clip,width=0.99\linewidth]
  {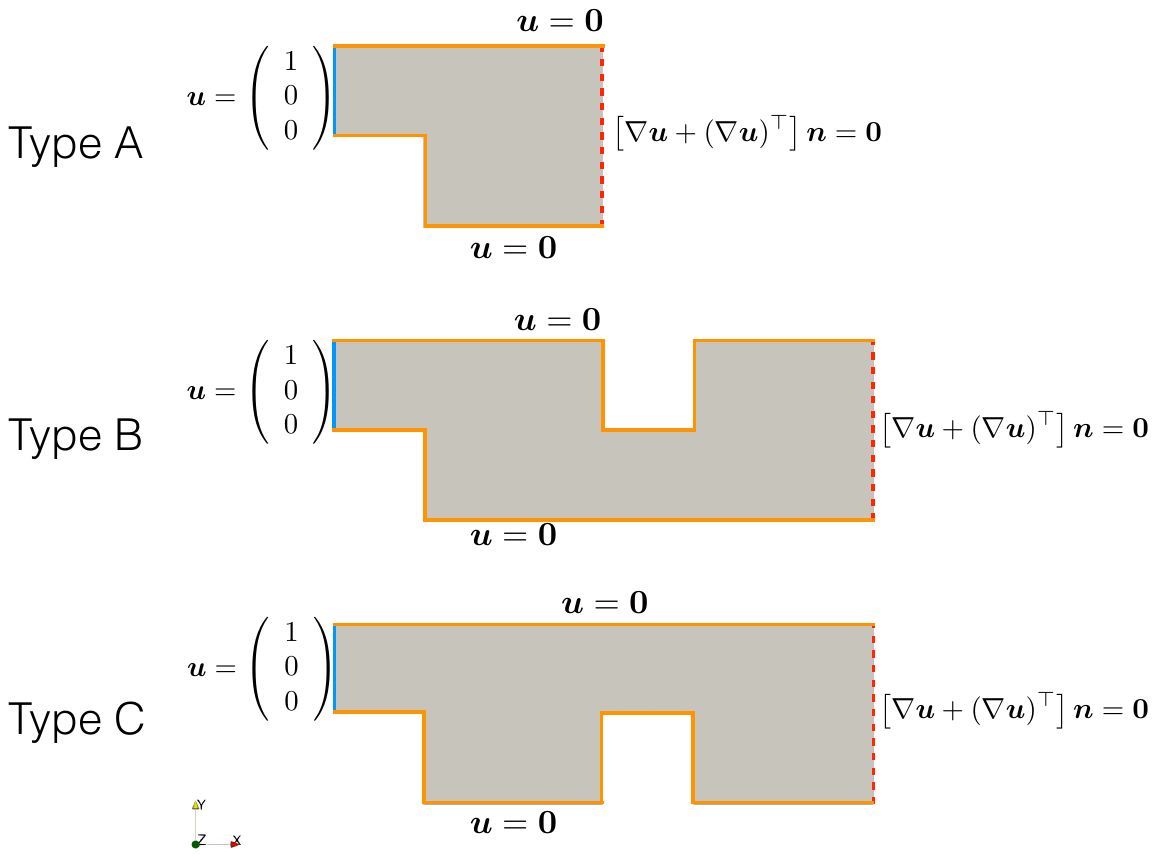}
  \caption{Boundary conditions of $\vu$ used to generate the dataset.
  The continuous lines and dotted lines correspond to Dirichlet and Neumann boundaries.
  }
  \label{fig:fluid_boundaries_u}
\end{figure}
\begin{figure}[t]
  \centering
  \includegraphics[trim={0cm 4cm 0cm 2cm},clip,width=0.99\linewidth]
  {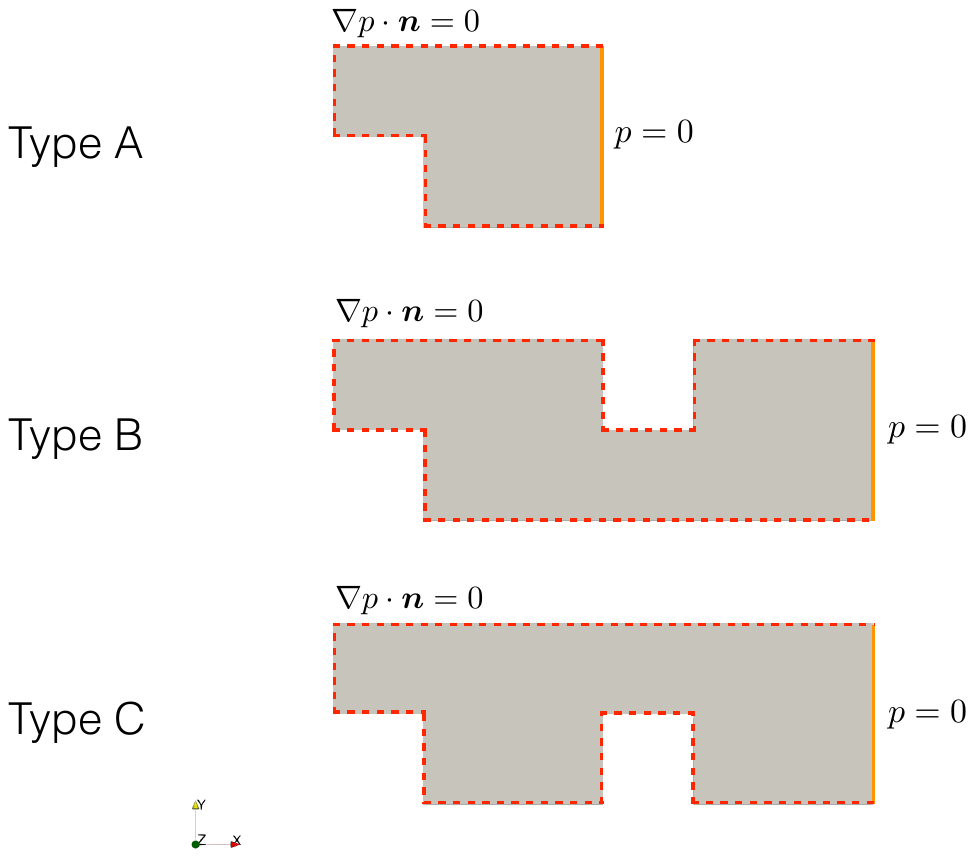}
  \caption{Boundary conditions of $p$ used to generate the dataset.
  The continuous lines and dotted lines correspond to Dirichlet and Neumann boundaries.
  }
  \label{fig:fluid_boundaries_p}
\end{figure}

\subsection{Dataset}
\label{app:fluid_dataset}
We generated numerical analysis results using various shapes of the computational domain, starting from three template shapes and changing their design parameters as shown in \Figref{fig:fluid_shapes}.
For each design parameter, we varied from 0 to 1.0 with a step size of 0.1, yielding 11 shapes for type A and 121 shapes for type B and C.
The boundary conditions were set as shown in Figures \ref{fig:fluid_boundaries_u} and \ref{fig:fluid_boundaries_p}.
These design and boundary conditions were chosen to have the characteristic length of 1.0 and flow speed of 1.0.
The viscosity was set to $10^{-3}$, resulting in Reynolds number $\mathrm{Re} \sim 10^{3}$.
The linear solvers used were generalized geometric-algebraic multi-grid for
$p$
and the smooth solver with the Gauss--Siedel smoother for
$\vu$.
Numerical analysis to generate each sample took up to one hour using CPU one core (Intel Xeon CPU E5-2695 v2@2.40GHz).
The dataset is uploaded online.\footnote{
  \url{https://savanna.ritc.jp/~horiem/penn_neurips2022/data/fluid/fluid_data.tar.gz.parta[a-e]}}

\subsection{Model architectures}
\label{app:fluid_model}
\begin{figure}[t]
  \centering
  \includegraphics[trim={0cm 7cm 14cm 2cm},clip,width=0.99\linewidth]
  {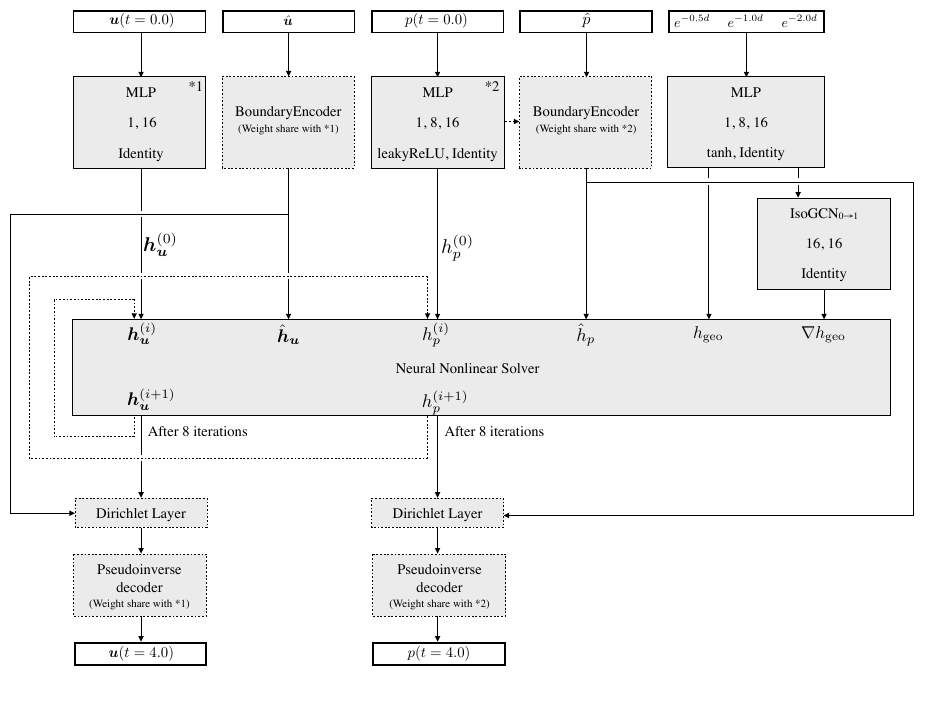}
  \caption{The overview of the PENN architecture for the incompressible flow dataset.
  Gray boxes with continuous (dotted) lines are trainable (untrainable) components.
  Arrows with dotted lines correspond to the loop.
  In each cell, we put the number of units in each layer along with the activation functions used.
  }
  \label{fig:fluid_network_overview}
\end{figure}

\begin{figure}[t]
  \centering
  \includegraphics[trim={0cm 2cm 13cm 2cm},clip,width=0.99\linewidth]
  {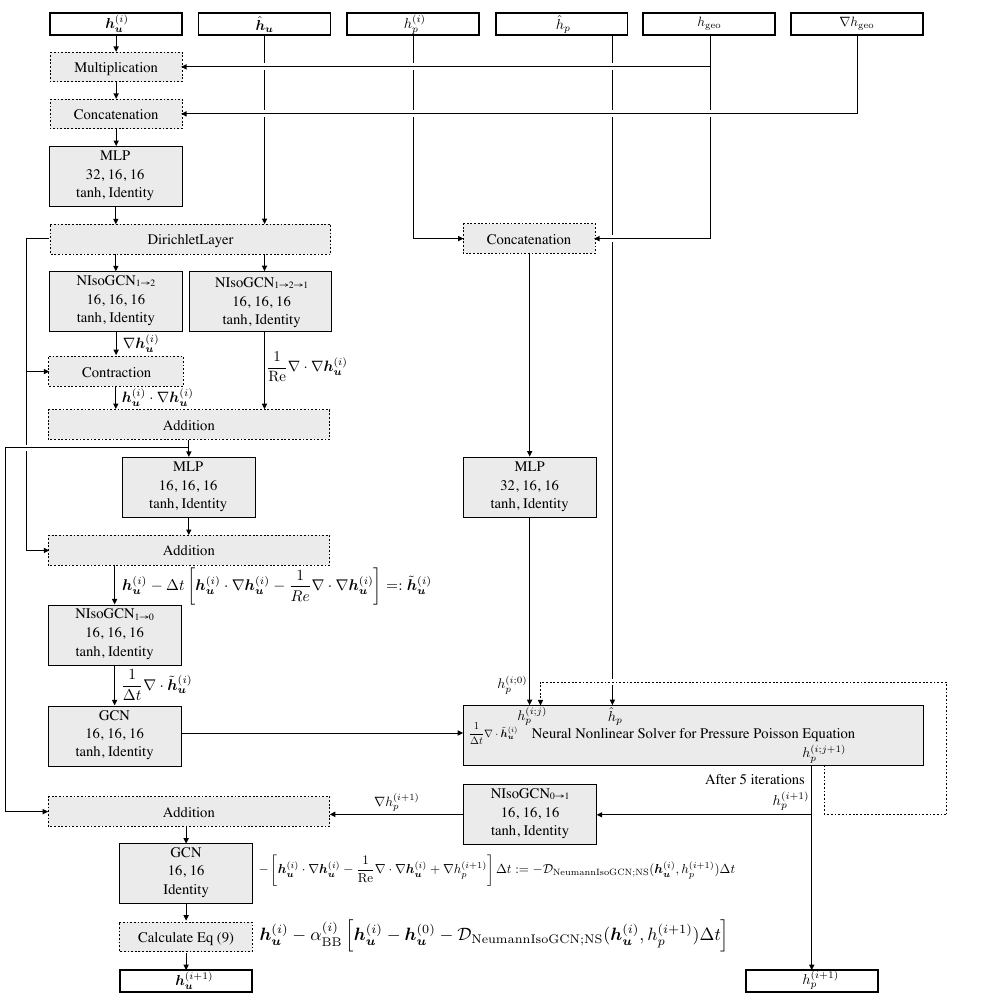}
  \caption{The neural nonlinear solver for velocity.
  Gray boxes with continuous (dotted) lines are trainable (untrainable) components.
  Arrows with dotted lines correspond to the loop.
  In each cell, we put the number of units in each layer along with the activation functions used.
  }
  \label{fig:fluid_network_group1}
\end{figure}

\begin{figure}[t]
  \centering
  \includegraphics[trim={0cm 11cm 18cm 2cm},clip,width=0.99\linewidth]
  {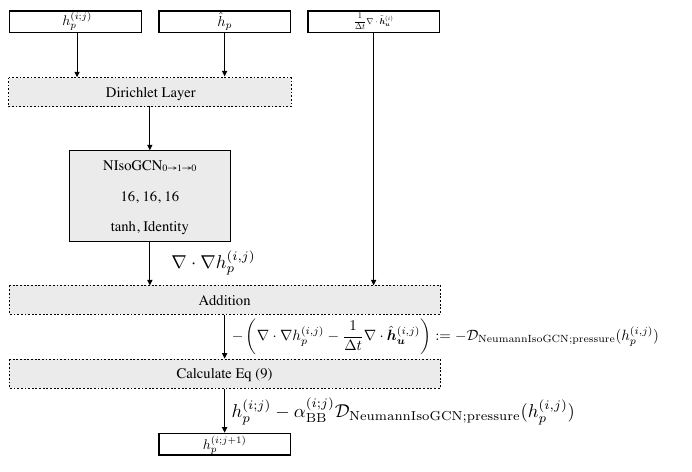}
  \caption{The neural nonlinear solver for pressure.
  Gray boxes with continuous (dotted) lines are trainable (untrainable) components.
  In each cell, we put the number of units in each layer along with the activation functions used.
  }
  \label{fig:fluid_network_group_poisson}
\end{figure}

The input features of the model are:
\begin{itemize}
  \item
    $\vu(t = 0.0)$: The initial velocity field, the solulsion of potential flow
  \item
    $\hat{\vu}$: The Dirichlet boundary condition for velocity
  \item
    $p(t = 0.0)$: The initial pressure field
  \item
    $\hat{p}$: The Dirichlet boundary condition for pressure
  \item
    $e^{-0.5 d}, e^{-1.0 d}, e^{-2.0 d}$: Features computed from $d$, the distance from the wall boundary condition
\end{itemize}
and the output features are:
\begin{itemize}
  \item
    $\vu(t = 4.0)$: The velocity field at $t = 4.0$
  \item
    $p(t = 4.0)$: The pressure field at $t = 4.0$
\end{itemize}

The strategy to construct PENN for the incompressible flow dataset is the following:
\begin{itemize}
  \item
    Consider the encoded version of the governing equation
  \item
    Apply the neural nonlinear solver containing the Dirichlet layer and the NIsoGCN to the encoded equation
  \item
    Decode the hidden feature using the pseudoinverse decoder.
\end{itemize}
Reflecting the fractional step method, we build PENN using spatial differential operators provided by NIsoGCN.
We use a simple linear encoder for the velocity and the associated Dirichlet boundary conditions.
For pressure and its Dirichlet constraint, we use a simple MLP with one hidden layer.
We encode each feature in a 16-dimensional space.
After features are encoded, we apply a neural nonlinear solver containing NeumanIsoGCNs and Dirichlet layers, reflecting the fractional step method (Equations \ref{eq:intermediate_velocity} and \ref{eq:fractional_step}).

The encoded equations are expressed as:
\begin{align}
  [
    \mathrm{NIsoGCN}_{1 \to 0}
    &\circ \mathrm{NIsoGCN}_{0 \to 1}\left( h_p \right)
  ](t + \Delta t, \vx)
  = \frac{1}{\Delta t} \left[\mathrm{NIsoGCN}_{1 \to 0}\left( \tilde{\vh}_\vu \right) \right](t, \vx),
  \label{eq:encoded_pressure_poisson}
  \\
  \tilde{\vh}_\vu
  :&= \vh_\vu - \Delta t \left[
    \vh_\vu \cdot \mathrm{NIsoGCN}_{1 \to 2} \left( \vh_\vu \right)
    - \frac{1}{\mathrm{Re}} \mathrm{NIsoGCN}_{2 \to 1} \circ \mathrm{NIsoGCN}_{1 \to 2} \left( \vh_\vu \right)
  \right],
  \label{eq:encoded_intermediate_velocity}
  \\
  \vh_\vu(t + \Delta t, \vx)
  &= \tilde{\vh}_\vu (t, \vx)
  - \Delta t \ \mathrm{NIsoGCN}_{0 \to 1} \left( h_p \right) (t + \Delta t, \vx),
  \label{eq:encoded_fractional_step}
\end{align}
where
$\vh_\vu$ is the encoded $\vu$ and
$h_p$ is the encoded $p$.
Note that these equations correspond to Equations \ref{eq:pressure_poisson}, \ref{eq:intermediate_velocity}, and \ref{eq:fractional_step}, by regarding IsoGCNs as spatial derivative operators.
The corresponding neural nonlinear solvers are expressed as:
\begin{align}
  &\vh_\vu^{(i+1)}
  =
  \vh_\vu^{(i)} - \alpha_\mathrm{BB}^{(i)} \left[ \vh_\vu^{(i)} - \vh_\vu^{(0)} - \mathcal{D}_\mathrm{NIsoGCN; NS} \left( \vh_\vu^{(i)}, h_p^{(i+1)} \right) \Delta t
  \right],
  \\[10pt]
  \mathcal{D}&_\mathrm{NIsoGCN; NS} \left( \vh_\vu^{(i)}, h_p^{(i+1)} \right)
  \nonumber
  \\
  :&=
  \left[
    \vh_\vu^{(i)} \cdot \mathrm{NIsoGCN}_{1 \to 2}\left( \vh_\vu^{(i)} \right)
    - \frac{1}{\mathrm{Re}} \mathrm{NIsoGCN}_{2 \to 1} \circ \mathrm{NIsoGCN}_{1 \to 2} \left( \vh_\vu^{(i)} \right)
    + \mathrm{NIsoGCN} \left( h_p^{(i+1)} \right)
  \right],
\end{align}
for $\vh_\vu$ and
\begin{align}
  h_p^{(i; j+1)}
  &=
  h_p^{(i; j)} - \alpha_\mathrm{BB}^{(i;j)} \mathcal{D}_\mathrm{NIsoGCN; pressure}(h_p^{(i; j)}),
  \\[10pt]
  \mathcal{D}_\mathrm{NIsoGCN; pressure} \left( h_p^{(i; j)} \right)
  :&=
  \left(
    \mathrm{NIsoGCN}_{1 \to 0} \circ \mathrm{NIsoGCN}_{0 \to 1}\left( h_p^{(;, j)} \right)
    - \frac{1}{\Delta t} \mathrm{NIsoGCN}_{1 \to 0} \left( \hat{\vh}_\vu^{(i)} \right)
  \right),
\end{align}
for $h_p$, where
$\vh_\vu^{(0)} = \vh_\vu(t, \cdot)$,
$h_p^{(0)} = h_p(t, \cdot)$, and
$h_p^{(i; 0)} = h_p^{(i)}$.
For notation regarding NIsoGCNs, please see \Appref{app:NIsoGCN}.
Figures \ref{fig:fluid_network_overview}, \ref{fig:fluid_network_group1}, and \ref{fig:fluid_network_group_poisson}
present the PENN model architecture used for the incompressible flow dataset.

As seen in \Figref{fig:fluid_network_group1}, we have a subloop that solves the Poisson equation for pressure in the nonlinear solver's loop for velocity.
We looped the solver for pressure five times and eight times for velocity.
After these loops stopped, we decoded the hidden features to obtain predictions for velocity and pressure, using the corresponding pseudoinverse decoders.

\subsection{Implementation details}
As discussed in \citet{horie2021isometric}, nonlinearity can be applied to the scalar but cannot be applied to the tensors with a rank equal to or greater than one. For such a tensor, nonlinearity can be applied to its norm as:
\begin{align}
  \mathrm{MLP}_\mathrm{tensor}(\vv) := \mathrm{MLP}(\Vert\vv\Vert) \vv.
\end{align}
This strategy to apply nonlinearity is used not only in the MLP blocks but also NIsoGCN blocks.
To facilitate the smoothness of pressure and velocity fields, we apply GCN layers corresponding to numerical viscosity in the standard numerical analysis method.
Here, please note that the PENN model consists of components that accept arbitrary input lengths, e.g., pointwise MLPs, deep sets, and NIsoGCNs. Thanks to the model's flexibility, we can apply the same model to arbitrary meshes similar to other GNNs.

\subsection{Training details}
Because the neural nonlinear solver applies the same layers many times during the loop, the model behaved somehow similar to recurrent neural networks during training, which could cause instability.
To avoid such unwanted behavior, we simply retried training by reducing the learning rate of the Adam optimizer by a factor of 0.5.
We found our way of training useful compared to using the learning rate schedule because sometimes the loss value of PENN can be extremely high, resulting in difficulty to reach convergence with a lower learning rate after such an explosion.
Therefore, we applied early stopping and restarted training using a lower learning rate from the epoch with the best validation loss.
Our initial learning rate was $5.0 \times 10^{-4}$, and we restarted the training twice, which was done automatically, within the 24-hour training period of PENN.
For the ablation study, we used the same setting for all models.
For PENN and ablation models, we used Adam~\citep{kingma2014adam} as an optimizer.
For MP-PDE solvers, we used the default setting written in the paper and the code.

\bgroup
\def\arraystretch{1.3}
\begin{table}[bt]
  \caption{MSE loss ($\pm$ the standard error of the mean)
  on test dataset of incompressible flow.
  If "Trans." is "Yes", it means evaluation on randomly
  rotated and transformed test dataset.
  $n$ denotes the number of hidden features,
  $r$ denotes the number of iterations in the neural nonlinear solver used in PENN models, and
  $\mathrm{TW}$ denotes the time window size used in MP-PDE models.
  }
  \label{tab:fluid_results_detailed}
  \centering
  \scalebox{1.0}{
    \begin{tabular}{llrrrr}
      \\[-8pt]
      \toprule
      Method
      &
      Trans.
      & \makecell{$\vu$\\$(\times 10^{-4})$}
      & \makecell{$p$\\$(\times 10^{-3})$}
      & \makecell{$\hat{\vu}_\mathrm{Dirichlet}$\\$(\times 10^{-4})$}
      & \makecell{$\hat{p}_\mathrm{Dirichlet}$\\$(\times 10^{-3})$}
      \\
      \hline
\multirow{2}{*}{\makecell[l]{PENN\\$n=16, r=8$}} &
No &
$4.36 \pm 0.03$ &
$1.17 \pm 0.01$ &
$0.00 \pm 0.00$ &
$0.00 \pm 0.00$
\\
& Yes &
$4.36 \pm 0.03$ &
$1.17 \pm 0.01$ &
$0.00 \pm 0.00$ &
$0.00 \pm 0.00$
\\
\multirow{2}{*}{\makecell[l]{PENN\\$n=16, r=4$}} &
No &
$29.09 \pm 0.17$ &
$11.35 \pm 0.04$ &
$0.00 \pm 0.00$ &
$0.00 \pm 0.00$
\\
& Yes &
$29.09 \pm 0.17$ &
$11.35 \pm 0.04$ &
$0.00 \pm 0.00$ &
$0.00 \pm 0.00$
\\
\multirow{2}{*}{\makecell[l]{PENN\\$n=8, r=8$}} &
No &
$177.42 \pm 0.93$ &
$35.70 \pm 0.12$ &
$0.00 \pm 0.00$ &
$0.00 \pm 0.00$
\\
& Yes &
$177.42 \pm 0.93$ &
$35.70 \pm 0.12$ &
$0.00 \pm 0.00$ &
$0.00 \pm 0.00$
\\
\multirow{2}{*}{\makecell[l]{PENN\\$n=8, r=4$}} &
No &
$26.82 \pm 0.16$ &
$7.86 \pm 0.03$ &
$0.00 \pm 0.00$ &
$0.00 \pm 0.00$
\\
& Yes &
$26.82 \pm 0.16$ &
$7.86 \pm 0.03$ &
$0.00 \pm 0.00$ &
$0.00 \pm 0.00$
\\
\multirow{2}{*}{\makecell[l]{PENN\\$n=4, r=8$}} &
No &
$92.80 \pm 0.52$ &
$31.47 \pm 0.13$ &
$0.00 \pm 0.00$ &
$0.00 \pm 0.00$
\\
& Yes &
$92.80 \pm 0.52$ &
$31.47 \pm 0.13$ &
$0.00 \pm 0.00$ &
$0.00 \pm 0.00$
\\
\multirow{2}{*}{\makecell[l]{PENN\\$n=4, r=4$}} &
No &
$120.35 \pm 0.65$ &
$35.53 \pm 0.12$ &
$0.00 \pm 0.00$ &
$0.00 \pm 0.00$
\\
& Yes &
$120.35 \pm 0.65$ &
$35.53 \pm 0.12$ &
$0.00 \pm 0.00$ &
$0.00 \pm 0.00$
\\
\multirow{2}{*}{\makecell[l]{MP-PDE\\$n=128, \mathrm{TW}=20$}} &
No &
$1.30 \pm 0.01$ &
$1.32 \pm 0.01$ &
$0.45 \pm 0.01$ &
$0.28 \pm 0.02$
\\
& Yes &
$1953.62 \pm 7.62$ &
$281.86 \pm 0.78$ &
$924.73 \pm 6.14$ &
$202.97 \pm 3.81$
\\
\multirow{2}{*}{\makecell[l]{MP-PDE\\$n=128, \mathrm{TW}=10$}} &
No &
$12.08 \pm 0.11$ &
$6.49 \pm 0.03$ &
$1.36 \pm 0.01$ &
$2.57 \pm 0.05$
\\
& Yes &
$1468.12 \pm 5.75$ &
$192.97 \pm 0.57$ &
$767.17 \pm 4.36$ &
$51.87 \pm 1.07$
\\
\multirow{2}{*}{\makecell[l]{MP-PDE\\$n=128, \mathrm{TW}=4$}} &
No &
$32.07 \pm 0.33$ &
$6.22 \pm 0.05$ &
$0.85 \pm 0.01$ &
$0.92 \pm 0.03$
\\
& Yes &
$2068.99 \pm 8.30$ &
$180.54 \pm 0.57$ &
$284.72 \pm 1.69$ &
$59.21 \pm 1.32$
\\
\multirow{2}{*}{\makecell[l]{MP-PDE\\$n=128, \mathrm{TW}=2$}} &
No &
$58.88 \pm 0.60$ &
$9.62 \pm 0.07$ &
$1.02 \pm 0.02$ &
$2.83 \pm 0.10$
\\
& Yes &
$1853.27 \pm 7.89$ &
$219.59 \pm 0.53$ &
$965.90 \pm 28.61$ &
$358.53 \pm 2.13$
\\
\multirow{2}{*}{\makecell[l]{MP-PDE\\$n=64, \mathrm{TW}=20$}} &
No &
$6.09 \pm 0.05$ &
$5.39 \pm 0.03$ &
$1.65 \pm 0.02$ &
$2.16 \pm 0.08$
\\
& Yes &
$1969.34 \pm 7.50$ &
$388.54 \pm 1.12$ &
$720.35 \pm 5.15$ &
$218.06 \pm 8.01$
\\
\multirow{2}{*}{\makecell[l]{MP-PDE\\$n=64, \mathrm{TW}=10$}} &
No &
$38.54 \pm 0.32$ &
$31.33 \pm 0.09$ &
$2.04 \pm 0.02$ &
$5.87 \pm 0.09$
\\
& Yes &
$2738.84 \pm 9.37$ &
$171.32 \pm 0.60$ &
$417.57 \pm 2.49$ &
$28.34 \pm 0.92$
\\
\multirow{2}{*}{\makecell[l]{MP-PDE\\$n=64, \mathrm{TW}=2$}} &
No &
$125.09 \pm 1.11$ &
$21.93 \pm 0.09$ &
$2.27 \pm 0.03$ &
$5.92 \pm 0.16$
\\
& Yes &
$1402.01 \pm 6.03$ &
$435.75 \pm 2.41$ &
$384.30 \pm 4.13$ &
$57.26 \pm 1.90$
\\
\multirow{2}{*}{\makecell[l]{MP-PDE\\$n=32, \mathrm{TW}=20$}} &
No &
$32.46 \pm 0.24$ &
$17.40 \pm 0.07$ &
$5.92 \pm 0.05$ &
$5.94 \pm 0.17$
\\
& Yes &
$2201.16 \pm 7.59$ &
$351.66 \pm 0.82$ &
$429.30 \pm 3.27$ &
$562.16 \pm 11.62$
\\
\multirow{2}{*}{\makecell[l]{MP-PDE\\$n=32, \mathrm{TW}=10$}} &
No &
$115.30 \pm 1.01$ &
$34.97 \pm 0.15$ &
$10.26 \pm 0.09$ &
$6.84 \pm 0.14$
\\
& Yes &
$2824.76 \pm 8.60$ &
$496.33 \pm 1.33$ &
$2276.11 \pm 10.57$ &
$488.50 \pm 5.01$
\\
\multirow{2}{*}{\makecell[l]{MP-PDE\\$n=32, \mathrm{TW}=4$}} &
No &
$272.73 \pm 2.07$ &
$94.27 \pm 0.45$ &
$11.50 \pm 0.12$ &
$35.76 \pm 0.29$
\\
& Yes &
$1973.35 \pm 8.29$ &
$554.69 \pm 4.26$ &
$647.31 \pm 7.40$ &
$157.85 \pm 8.41$
\\
\multirow{2}{*}{\makecell[l]{MP-PDE\\$n=32, \mathrm{TW}=2$}} &
No &
$794.90 \pm 4.68$ &
$82.61 \pm 0.40$ &
$50.23 \pm 0.91$ &
$31.41 \pm 1.88$
\\
& Yes &
$3240.69 \pm 21.91$ &
$443.10 \pm 2.56$ &
$2885.30 \pm 41.17$ &
$562.08 \pm 19.28$
\\
    \bottomrule
    \end{tabular}
  }
\end{table}
\egroup

\subsection{Result details}
\Tabref{tab:fluid_results_detailed} presents the detailed results of the comparison between MP-PDE and PENN.
Interestingly, the performance of MP-PDE gets better as the time window size increases. Therefore, our future direction may be to incorporate MP-PDE's temporal bundling and pushforward trick into PENN to enable us to predict the state after a far longer time than we do in the present work.

Tables \ref{tab:fluid_compromise_penn} and \ref{tab:fluid_compromise_mppde} show the speed and accuracy of the machine learning models tested.
PENN models show excellent performance with a lot smaller number of parameters compared to MP-PDE models.
It is achieved due to efficient parameter sharing in the proposed model, e.g., the same weights are used repeatedly in the neural nonlinear encoder.
Also, as pointed out in
\citet{pmlr-v70-ravanbakhsh17a},
there is a strong connection between parameter sharing and equivariance.
PENN has equivariance in, e.g., permutation, time translation, and $\mathrm{E}(n)$ through parameter sharing, which is in line with them.

\Tabref{tab:fluid_compromise_openfoam} presents the speed and accuracy with various settings of OpenFOAM to seek a speed-accuracy tradeoff.
We tested three configurations of linear solvers:
\begin{itemize}
  \item
    Generalized geometric-algebraic multi-grid (GAMG) for $p$ and the smooth solver for $\vu$
  \item
    Generalized geometric-algebraic multi-grid (GAMG) for both $p$ and $\vu$
  \item
    The smooth solver for $p$ and $\vu$
\end{itemize}
In addition, we tested different resolutions for space and time by changing:
\begin{itemize}
  \item
    The number of divisions per unit length: 22.5, 45.0, 90.0
  \item
    Time step size: 0.001, 0.005, 0.010, 0.050
\end{itemize}
Ground truth is computed using the number of divisions per unit length of 90.0 and time step size of 0.001; thus, this combination is eliminated from the comparison because the MSE error is underestimated (in particular, zero).

\bgroup
\def\arraystretch{1.3}
\begin{table}[bt]
  \caption{MSE loss ($\pm$ the standard error of the mean) of PENN models
  on test dataset of incompressible flow.
  }
  \label{tab:fluid_compromise_penn}
  \centering
  \scalebox{1.0}{
    \begin{tabular}{ccrrr}
      \\[-8pt]
      \toprule
      \makecell{
        \# hidden
        \\
        feature
      }
      &
      \makecell{
        \# iteration in
        \\
        the neural nonlinear solver
      }
      &
      \# parameter
      &
      \makecell{Total MSE\\$(\times 10^{-3})$}
      &
      \makecell{Total time [s]}
      \\
      \hline
16 &8 &8,432 &$1.61 \pm 0.01$ &$5.33 \pm 0.13$
\\
16 &4 &8,432 &$14.26 \pm 0.03$ &$2.52 \pm 0.06$
\\
8 &8 &2,100 &$53.44 \pm 0.11$ &$3.54 \pm 0.08$
\\
8 &4 &2,100 &$10.54 \pm 0.03$ &$2.16 \pm 0.04$
\\
4 &8 &596 &$40.75 \pm 0.10$ &$2.86 \pm 0.06$
\\
4 &4 &596 &$47.57 \pm 0.10$ &$1.35 \pm 0.04$
\\
    \bottomrule
    \end{tabular}
  }
\end{table}
\egroup

\bgroup
\def\arraystretch{1.3}
\begin{table}[bt]
  \caption{MSE loss ($\pm$ the standard error of the mean) of MP-PDE models
  on test dataset of incompressible flow.
  }
  \label{tab:fluid_compromise_mppde}
  \centering
  \scalebox{1.0}{
    \begin{tabular}{ccrrrr}
      \\[-8pt]
      \toprule
      \makecell{
        \# hidden
        \\
        feature
      }
      &
      Time window size
      &
      \# parameter
      &
      \makecell{Total MSE\\$(\times 10^{-3})$}
      &
      \makecell{Total MSE (Trans.)\\$(\times 10^{-3})$}
      &
      \makecell{Total time [s]}
      \\
      \hline
128 &20 &709,316 &$1.45 \pm 0.01$ &$477.23 \pm 0.77$ &$51.61 \pm 1.41$
\\
128 &10 &673,484 &$7.70 \pm 0.02$ &$339.78 \pm 0.57$ &$94.01 \pm 2.66$
\\
128 &4 &651,972 &$9.43 \pm 0.04$ &$387.44 \pm 0.71$ &$137.32 \pm 3.91$
\\
128 &2 &644,548 &$15.51 \pm 0.07$ &$404.92 \pm 0.67$ &$57.28 \pm 1.91$
\\
64 &20 &204,004 &$6.00 \pm 0.02$ &$585.48 \pm 0.95$ &$13.62 \pm 0.38$
\\
64 &10 &185,356 &$35.19 \pm 0.07$ &$445.20 \pm 0.79$ &$23.73 \pm 0.67$
\\
64 &2 &174,740 &$34.44 \pm 0.10$ &$575.95 \pm 1.76$ &$32.61 \pm 1.02$
\\
32 &20 &63,964 &$20.64 \pm 0.05$ &$571.77 \pm 0.79$ &$7.64 \pm 0.24$
\\
32 &10 &55,348 &$46.50 \pm 0.13$ &$778.80 \pm 1.12$ &$12.93 \pm 0.39$
\\
32 &4 &49,948 &$121.55 \pm 0.35$ &$752.03 \pm 3.07$ &$13.99 \pm 0.41$
\\
32 &2 &47,924 &$162.10 \pm 0.44$ &$767.17 \pm 2.38$ &$4.55 \pm 0.13$
\\
    \bottomrule
    \end{tabular}
  }
\end{table}
\egroup

\bgroup
\def\arraystretch{1.3}
\begin{table}[bt]
  \caption{MSE loss ($\pm$ the standard error of the mean)
  of OpenFOAM computations
  on test dataset of incompressible flow.
  }
  \label{tab:fluid_compromise_openfoam}
  \centering
  \scalebox{1.0}{
    \begin{tabular}{ccccrr}
      \\[-8pt]
      \toprule
      Solver for $\vu$
      &
      Solver for $p$
      &
      \makecell{\# division\\per unit length}
      &
      $\Delta t$
      &
      Total MSE ($\times 10^{-3}$)
      &
      Total time [s]
      \\
      \hline
GAMG & Smooth & 22.5 & 0.050 & Divergent & Divergent
\\
GAMG & Smooth & 22.5 & 0.010 & $6.09 \pm 0.02$ & $6.08 \pm 0.17$
\\
GAMG & Smooth & 22.5 & 0.005 & $6.04 \pm 0.02$ & $11.57 \pm 0.32$
\\
GAMG & Smooth & 22.5 & 0.001 & $4.80 \pm 0.02$ & $51.43 \pm 1.39$
\\
GAMG & Smooth & 45.0 & 0.050 & Divergent & Divergent
\\
GAMG & Smooth & 45.0 & 0.010 & $0.46 \pm 0.00$ & $25.12 \pm 0.81$
\\
GAMG & Smooth & 45.0 & 0.005 & $0.78 \pm 0.00$ & $46.71 \pm 1.53$
\\
GAMG & Smooth & 45.0 & 0.001 & $1.04 \pm 0.00$ & $201.11 \pm 6.29$
\\
GAMG & Smooth & 90.0 & 0.050 & Divergent & Divergent
\\
GAMG & Smooth & 90.0 & 0.010 & Divergent & Divergent
\\
GAMG & Smooth & 90.0 & 0.005 & $0.15 \pm 0.00$ & $231.18 \pm 10.38$
\\
GAMG & GAMG & 22.5 & 0.050 & Divergent & Divergent
\\
GAMG & GAMG & 22.5 & 0.010 & $6.05 \pm 0.02$ & $6.41 \pm 0.18$
\\
GAMG & GAMG & 22.5 & 0.005 & $6.00 \pm 0.02$ & $12.21 \pm 0.34$
\\
GAMG & GAMG & 22.5 & 0.001 & $4.80 \pm 0.02$ & $55.51 \pm 1.52$
\\
GAMG & GAMG & 45.0 & 0.050 & Divergent & Divergent
\\
GAMG & GAMG & 45.0 & 0.010 & $0.46 \pm 0.00$ & $26.00 \pm 0.85$
\\
GAMG & GAMG & 45.0 & 0.005 & $0.77 \pm 0.00$ & $48.78 \pm 1.57$
\\
GAMG & GAMG & 45.0 & 0.001 & $1.03 \pm 0.00$ & $214.29 \pm 6.62$
\\
GAMG & GAMG & 90.0 & 0.050 & Divergent & Divergent
\\
GAMG & GAMG & 90.0 & 0.010 & Divergent & Divergent
\\
GAMG & GAMG & 90.0 & 0.005 & $0.14 \pm 0.00$ & $238.94 \pm 10.70$
\\
Smooth & Smooth & 22.5 & 0.050 & Divergent & Divergent
\\
Smooth & Smooth & 22.5 & 0.010 & $5.59 \pm 0.02$ & $85.50 \pm 3.05$
\\
Smooth & Smooth & 22.5 & 0.005 & $5.41 \pm 0.02$ & $164.36 \pm 7.57$
\\
Smooth & Smooth & 22.5 & 0.001 & $4.19 \pm 0.02$ & $765.50 \pm 29.65$
\\
Smooth & Smooth & 45.0 & 0.050 & Divergent & Divergent
\\
Smooth & Smooth & 45.0 & 0.010 & $51.10 \pm 0.05$ & $426.07 \pm 22.51$
\\
Smooth & Smooth & 45.0 & 0.005 & $2.09 \pm 0.00$ & $824.71 \pm 39.90$
\\
Smooth & Smooth & 45.0 & 0.001 & $1.12 \pm 0.00$ & $3960.88 \pm 151.93$
\\
Smooth & Smooth & 90.0 & 0.050 & Divergent & Divergent
\\
Smooth & Smooth & 90.0 & 0.010 & Divergent & Divergent
\\
Smooth & Smooth & 90.0 & 0.005 & $4493.78 \pm 1.88$ & $3566.05 \pm 183.75$
\\
    \bottomrule
    \end{tabular}
  }
\end{table}
\egroup

\bgroup
\def\arraystretch{1.3}
\begin{table}[bt]
  \caption{Ablation study on
  2D incompressible flow dataset.
  The value represents MSE loss ($\pm$ standard error of the mean) on the
  test dataset.
  "Divergent" means the implicit solver does not converge
  and the loss gets extreme value ($\sim 10^{14}$).
  This presents the same results as \Tabref{tab:fluid_ablation}.
  }
  \centering
  \label{tab:fluid_ablation_appendix}
  \scalebox{1.0}{
    \begin{tabular}{lrrrr}
      \\[-8pt]
      \toprule
      Method
      & \makecell{$\vu$\\$(\times 10^{-4})$}
      & \makecell{$p$\\$(\times 10^{-3})$}
      & \makecell{$\hat{\vu}_\mathrm{Dirichlet}$\\$(\times 10^{-4})$}
      & \makecell{$\hat{p}_\mathrm{Dirichlet}$\\$(\times 10^{-3})$}
      \\
      \hline
      (A) Without encoded boundary &
      Divergent &
      Divergent &
      Divergent &
      Divergent
      \\[3pt]
      \makecell[l]{
        (B) Without boundary condition
        \\
        in the neural nonlinear solver
      } &
$65.10 \pm 0.38$ &
$21.70 \pm 0.09$ &
$0.00 \pm 0.00$ &
$0.00 \pm 0.00$
      \\[3pt]
      (C) Without neural nonlinear solver &
$31.03 \pm 0.19$ &
$9.81 \pm 0.04$ &
$\boldsymbol{0.00} \pm 0.00$ &
$\boldsymbol{0.00} \pm 0.00$
      \\[3pt]
      (D) Without boundary condition input &
$20.08 \pm 0.21$ &
$3.61 \pm 0.02$ &
$59.60 \pm 0.89$ &
$1.43 \pm 0.05$
      \\[3pt]
      (E) Without Dirichlet layer &
$8.22 \pm 0.07$ &
$1.41 \pm 0.01$ &
$18.20 \pm 0.28$ &
$0.38 \pm 0.01$
      \\[3pt]
      (F) Without pseudoinverse decoder
      &
$8.91 \pm 0.06$ &
$2.36 \pm 0.02$ &
$1.97 \pm 0.06$ &
$\boldsymbol{0.00} \pm 0.00$
      \\[3pt]
      \makecell[l]{
        (G) Without pseudoinverse decoder
        \\
        with Dirichlet layer after decoding
      } &
$6.65 \pm 0.05$ &
$1.71 \pm 0.01$ &
$\boldsymbol{0.00} \pm 0.00$ &
$\boldsymbol{0.00} \pm 0.00$
      \\[5pt]
      \textbf{PENN} &
      $\boldsymbol{4.36} \pm 0.03$ &
      $\boldsymbol{1.17} \pm 0.01$ &
      $\boldsymbol{0.00} \pm 0.00$ &
      $\boldsymbol{0.00} \pm 0.00$
      \\
      \bottomrule
    \end{tabular}
  }
\end{table}
\egroup

\begin{figure}[bt]
  \centering

  \stackunder[-60pt]
  {\includegraphics[trim={200cm 19cm 0cm 19cm},clip,width=0.05\textwidth]
    {figs/fluid/mp_pde/answer_u.png}}
  {(i)}
  \stackunder[3pt]
  {\includegraphics[trim={27cm 19cm 27cm 19cm},clip,width=0.4\textwidth]
    {figs/fluid/mp_pde/answer_u.png}}
  {}
  \hspace{5pt}
  \stackunder[3pt]
  {\includegraphics[trim={27cm 19cm 27cm 19cm},clip,width=0.4\textwidth]
    {figs/fluid/mp_pde/answer_p.png}}
  {}
  \\
  \stackunder[-60pt]
  {\includegraphics[trim={200cm 19cm 0cm 19cm},clip,width=0.05\textwidth]
    {figs/fluid/mp_pde/answer_u.png}}
  {(ii)}
  {\includegraphics[trim={27cm 19cm 27cm 19cm},clip,width=0.4\textwidth]
    {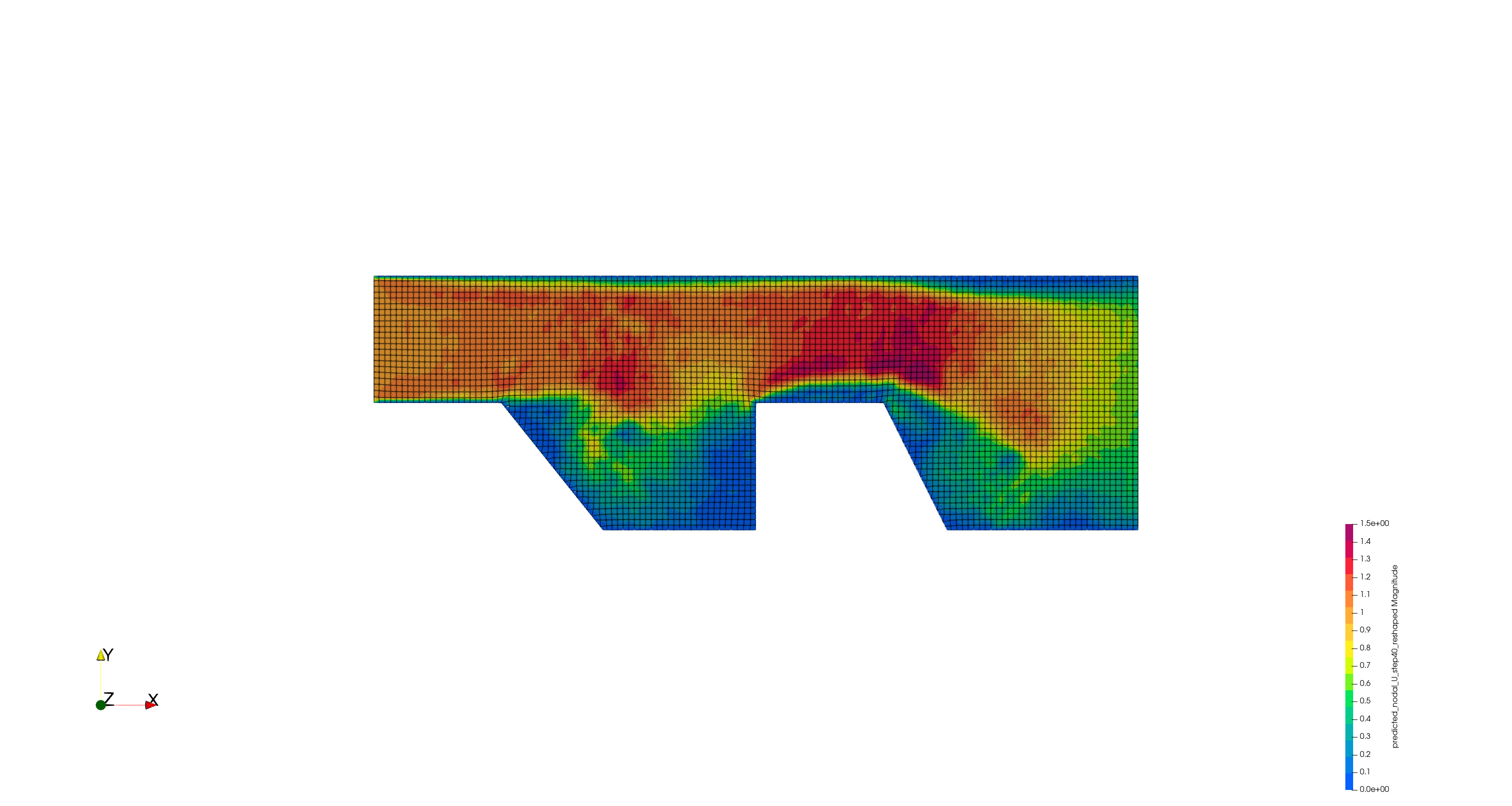}}
  {}
  \hspace{5pt}
  \stackunder[3pt]
  {\includegraphics[trim={27cm 19cm 27cm 19cm},clip,width=0.4\textwidth]
    {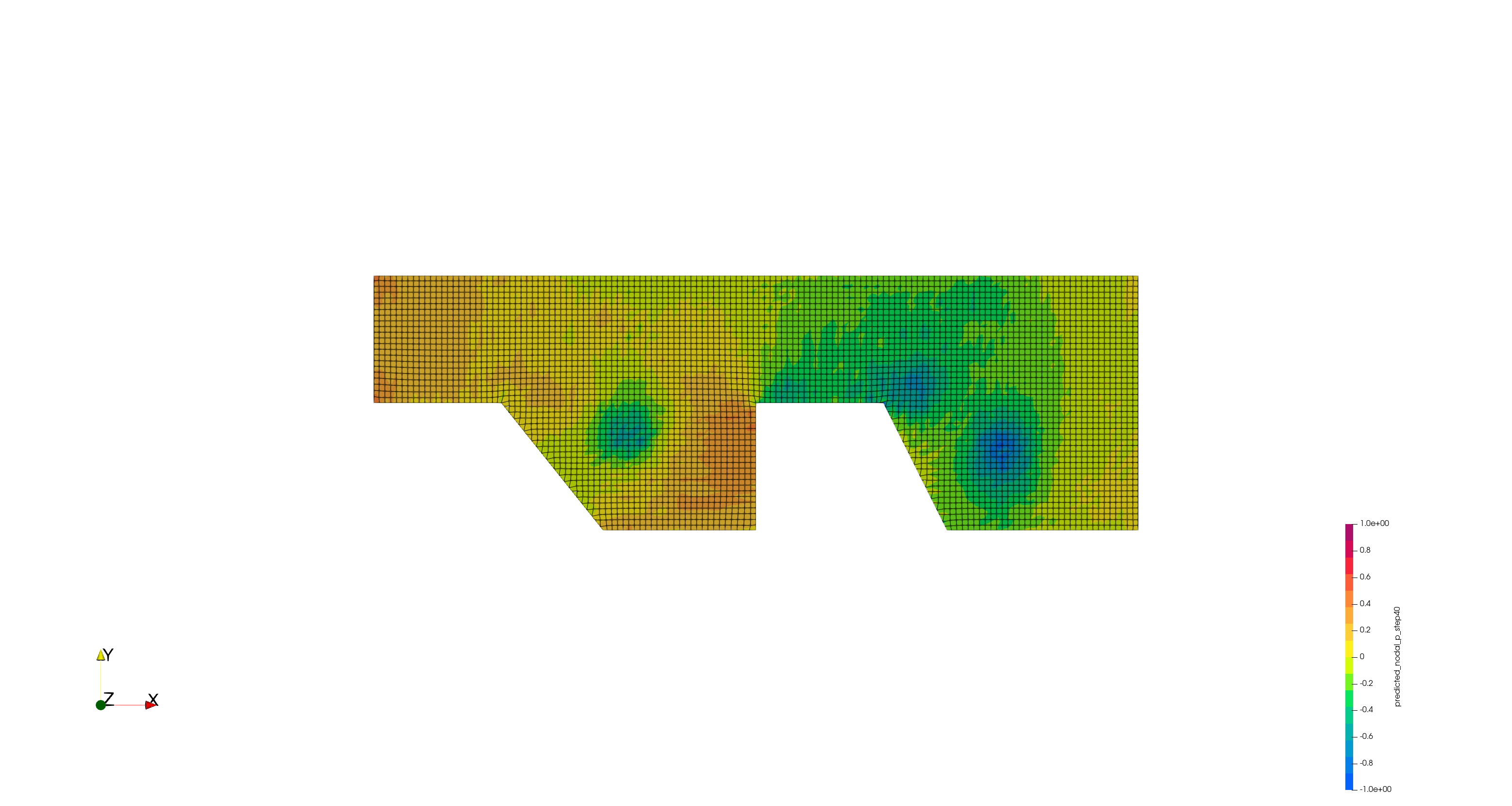}}
  {}
  \\
  \stackunder[-60pt]
  {\includegraphics[trim={200cm 19cm 0cm 19cm},clip,width=0.05\textwidth]
    {figs/fluid/mp_pde/answer_u.png}}
  {(iii)}
  {\includegraphics[trim={27cm 19cm 27cm 19cm},clip,width=0.4\textwidth]
    {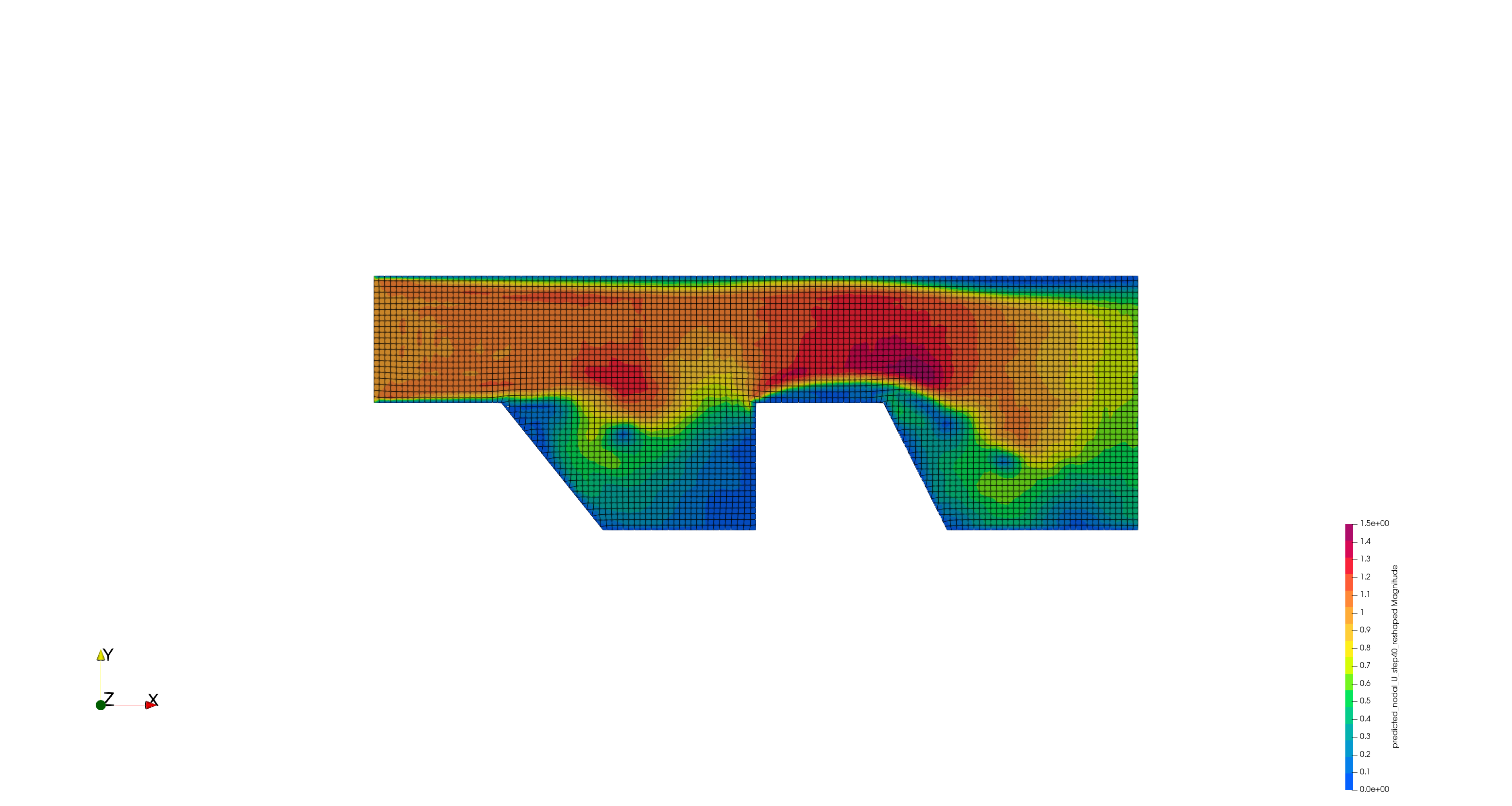}}
  {}
  \hspace{5pt}
  \stackunder[3pt]
  {\includegraphics[trim={27cm 19cm 27cm 19cm},clip,width=0.4\textwidth]
    {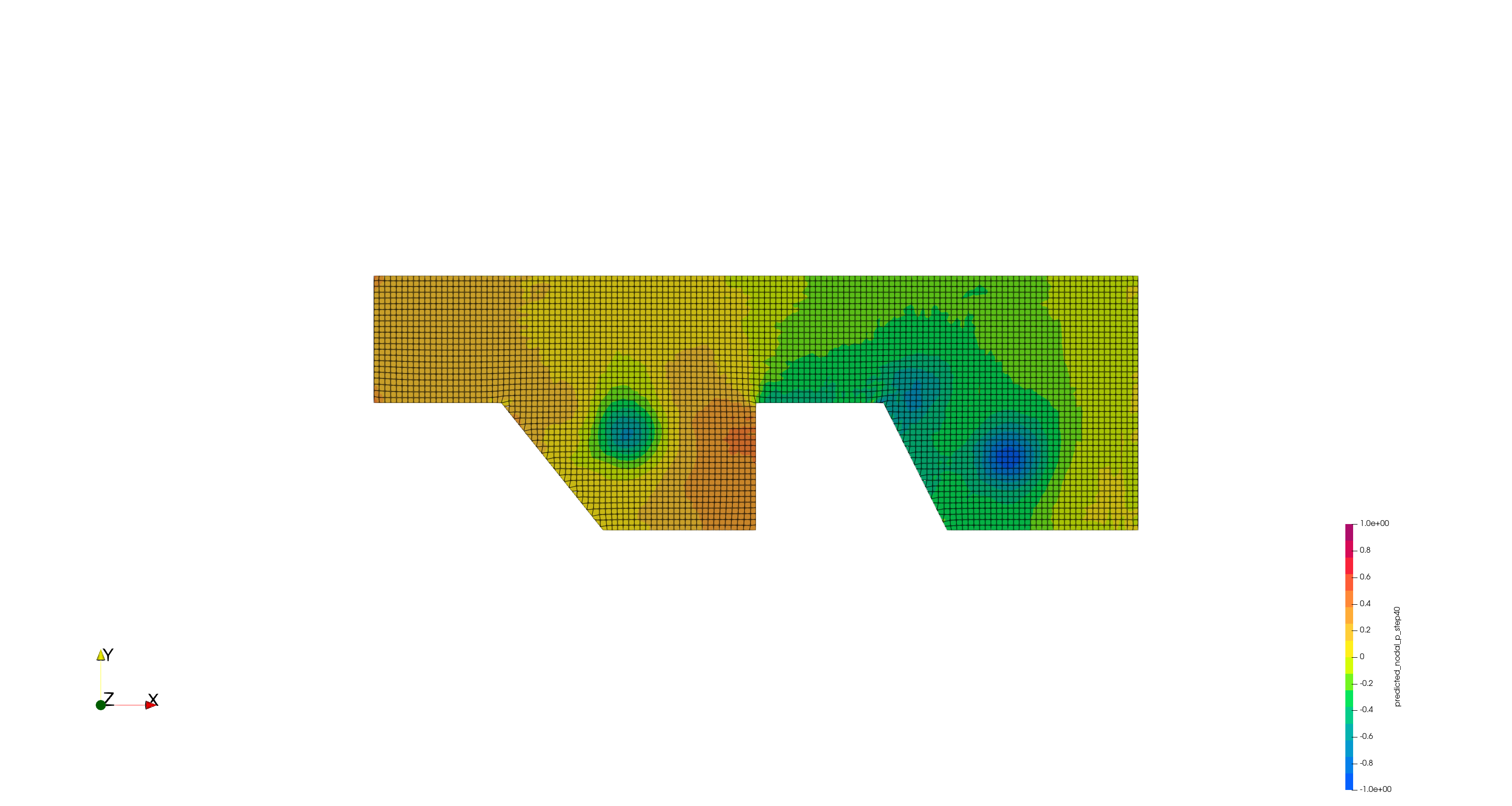}}
  {}
  \\
  \stackunder[-60pt]
  {\includegraphics[trim={200cm 19cm 0cm 19cm},clip,width=0.05\textwidth]
    {figs/fluid/mp_pde/answer_u.png}}
  {(iv)}
  {\includegraphics[trim={27cm 19cm 27cm 19cm},clip,width=0.4\textwidth]
    {figs/fluid/penn/pred_u.png}}
  {}
  \hspace{5pt}
  \stackunder[3pt]
  {\includegraphics[trim={27cm 19cm 27cm 19cm},clip,width=0.4\textwidth]
    {figs/fluid/penn/pred_p.png}}
  {}
  \\
  \stackunder[-60pt]
  {\includegraphics[trim={200cm 19cm 0cm 19cm},clip,width=0.05\textwidth]
    {figs/fluid/mp_pde/answer_u.png}}
  {}
  \stackunder[3pt]
  {\includegraphics[trim={0cm 0cm 0cm 0cm},clip,width=0.4\textwidth]
    {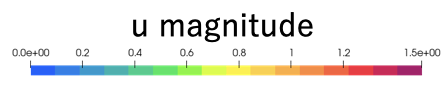}}
  {}
  \hspace{5pt}
  \stackunder[3pt]
  {\includegraphics[trim={0cm 0cm 0cm 0cm},clip,width=0.4\textwidth]
    {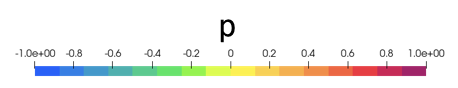}}
  {}
  \caption{Visual comparison of the ablation study of
  (i) ground truth,
  (ii) the model without the neural nonlinear solver (Model (C)),
  (iii) the model without pseudoinverse decoder with Dirichlet layer after decoding (Model (G)), and
  (iv) PENN.
  It can be observed that PENN improves the prediction smoothness, especially for the velocity field.
  }
  \label{fig:fluid_ablation}
\end{figure}

\subsection{Ablation study details}
To validate the effectiveness of our model through an ablation study on the following settings:
\begin{enumerate}[label=(\Alph*)]
  \item
    Without encoded boundary: In the nonlinear loop, we decode features to apply boundary conditions to fulfill Dirichlet conditions in the original physical space
  \item
    Without boundary condition in the neural nonlinear solver: We removed the Dirichlet layer in the nonlinear loop. Instead, we added the Dirichlet layer after the (non-pseudoinverse) decoder.
  \item
    Without neural nonlinear solver: We removed the nonlinear solver from the model and used the explicit time-stepping instead
  \item
    Without boundary condition input: We removed the boundary condition from input features
  \item
    Without Dirichlet layer: We removed the Dirichlet layer. Instead, we let the model learn to satisfy boundary conditions during training.
  \item
    Without pseudoinverse decoder: We removed the pseudoinverse decoder and used simple MLPs for decoders.
  \item
    Without pseudoinverse decoder with Dirichlet boundary layer after decoding: Same as above, but with Dirichlet layer after decoding.
\end{enumerate}

We again put the results of the ablation study in \Tabref{tab:fluid_ablation_appendix},
which is already presented in \Tabref{tab:fluid_ablation}, for the convenience of the readers.

Comparison with Model (A) shows that the nonlinear loop in the encoded space is inevitable for machine learning.
This result is quite convincing because if the loop is made in the original space, the advantage of the expressive power of the neural networks cannot be leveraged.
Comparison with Model (C) confirms that the concept of the solver is effective compared to simply stacking GNNs, corresponding to the explicit method.

If the boundary condition input is excluded (Model (D)), the performance degrades in line with \citet{brandstetter2022message}.
That model also has an error on the Dirichlet boundaries.
Model (E) shows a similar result, improving performance using the information of the boundary conditions.
If the pseudoinverse decoder is excluded (Model (F)), the output may not satisfy the Dirichlet boundary conditions as well.
Besides, the decoder has more effect than expected because PENN is better than Model (G).
Both models satisfy the Dirichlet boundary condition, while PENN has significant improvement.
This may be because the pseudoinverse decoder facilitates the  spatial continuity of the outputs in addition to the fulfillment of the Dirichlet boundary condition.
In other words, using a simple decoder and the Dirichlet layer after that may cause  spatial discontinuity of outputs.
Visual comparison of part of the ablation study is shown in \Figref{fig:fluid_ablation}.

\clearpage

\section{Experiment details: advection-diffusion dataset}
\label{app:ad}
To test the generalization ability of PENNs regarding PDE's parameters and time series, we run an experiment with the advection-diffusion dataset.
The governing equation regarding the temperature field
$T$
used for the experiment is expressed as:
\begin{align}
  \pdiff{T}{t}
  &= - c \left(
    \begin{array}{c}
      1
      \\
      0
      \\
      0
    \end{array}
  \right)
  \cdot \nabla T + D \nabla \cdot \nabla T
  &(t, \vx)
  &\in (0, 1) \times \Omega,
  \\
  T(t = 0, \vx)
  &= 0
  &\vx
  &\in \Omega,
  \\
  T
  &= \hat{T}
  &(t, \vx)
  &\in \partial\Omega_\mathrm{Dirichlet},
  \\
  \nabla T \cdot \vn
  &= 0
  &(t, \vx)
  &\in \partial\Omega_\mathrm{Neumann},
\end{align}
where
$c \in \mathbb{R}$
is the magnitude of a known velocity field, and
$D \in \mathbb{R}$ is the diffusion coefficient.
We set
$\Omega = \{\vx \in \mathbb{R}^3 \ | \ 0 < x_1 < 1 \land 0 < x_2 < 1 \land 0 < x_3 < 0.01 \}$
,
$\partial\Omega_\mathrm{Dirichlet} = \{\vx \in \partial \Omega \ | \ x_1 = 0 \}$
and
$\partial\Omega_\mathrm{Neumann} = \partial\Omega \setminus \partial\Omega_\mathrm{Dirichlet}$
.

\subsection{Dataset}
We varied
$c$
and
$D$
from 0.0 to 1.0, eliminating the condition
$c = D = 0.0$
because nothing drives the phenomena, and
and varied
$\hat{T}$
from 0.1 to 1.0.
Like the incompressible flow dataset, we generated fine meshes, ran computation with OpenFOAM, and interpolated the obtained temperature fields onto coarser meshes. We split the generated data into training, validation, and test dataset containing 960, 120, and 120 samples.
The dataset is uploaded online.\footnote{\url{https://savanna.ritc.jp/~horiem/penn_neurips2022/data/ad/ad_preprocessed.tar.gz}}

\subsection{Model architecture}
The strategy to construct PENN for the advection-diffusion dataset is consistent with one for the incompressible flow dataset (see \Appref{app:fluid_model}).
The input features of the model are:
\begin{itemize}
  \item
    $T(t = 0.0)$: The initial temperature field
  \item
    $\hat{T}$: The Dirichlet boundary condition for the temperature field
  \item
    $(c, 0, 0)^\top$: The velocity field
  \item
    $c$: The magnitude of the velocity
  \item
    $D$: The diffusion coefficient
  \item
    $e^{-0.5 d}, e^{-1.0 d}, e^{-2.0 d}$: Features computed from $d$, the distance from the Dirichlet boundary
\end{itemize}
and the output features are:
\begin{itemize}
  \item
    $T(t = 0.25)$: The temperature field at $t = 0.25$
  \item
    $T(t = 0.50)$: The temperature field at $t = 0.50$
  \item
    $T(t = 0.75)$: The temperature field at $t = 0.75$
  \item
    $T(t = 1.00)$: The temperature field at $t = 1.00$
\end{itemize}

The encoded governing equation is expressed as:
\begin{align}
  h_T(t + \Delta t, \vx)
  &=
  h_T(t, \vx) + \mathcal{D}_\mathrm{NIsoGCN; \AD}\left(h_T\right) (t + \Delta t, \vx)
  \\
  \mathcal{D}_\mathrm{NIsoGCN; \AD}\left(h_T\right)
  :&=
  - \vh_\vc \cdot \mathrm{NIsoGCN}_{0 \to 1} (h_T)
  + h_D \ \mathrm{NIsoGCN}_{0 \to 1 \to 0} (h_T)
\end{align}
The corresponding neural nonlinear solver is:
\begin{align}
  h_T\supp{i+1}
  = h_T\supp{i} - \alpha_\mathrm{BB}\supp{i}
    \left[
      h_T\supp{i} - h_T\supp{0} - \mathcal{D}_\mathrm{NIsoGCN; \AD}(h_T\supp{i})\Delta t
    \right],
\end{align}

\begin{figure}[tb]
  \centering
  \includegraphics[trim={0cm 13cm 19cm 2cm},clip,width=0.8\linewidth]
  {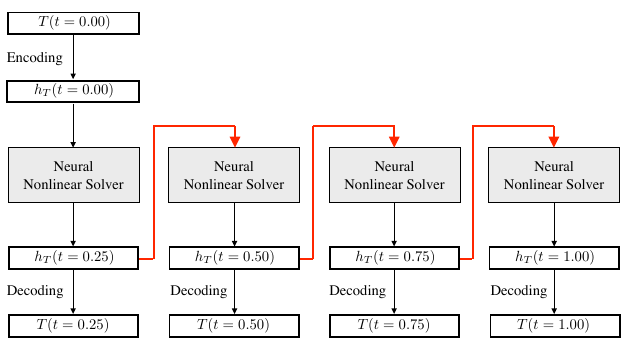}
  \caption{
    The concept of the neural nonlinear solver for time series data with autoregressive architecture. The solver's output is fed to the same solver to obtain the state at the next time step (bold red arrow).
    Please note that this architecture can be applied to arbitrary time series lengths.}
  \label{fig:ad_network_temporal}
\end{figure}

\begin{figure}[tb]
  \centering
  \includegraphics[trim={0cm 9cm 17cm 2cm},clip,width=0.99\linewidth]
  {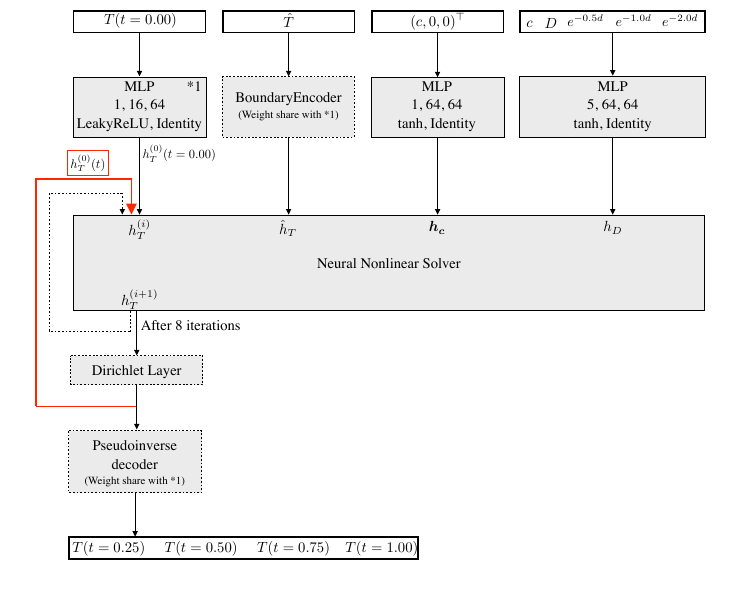}
  \caption{The overview of the PENN architecture for the advection-diffusion dataset.
  Gray boxes with continuous (dotted) lines are trainable (untrainable) components.
  Arrows with dotted lines correspond to the loop.
  In each cell, we put the number of units in each layer along with the activation functions used.
  The bold red arrow corresponds to the one in \Figref{fig:ad_network_temporal}.
  }
  \label{fig:ad_network_overview}
\end{figure}

\begin{figure}[tb]
  \centering
  \includegraphics[trim={0cm 4cm 14cm 2cm},clip,width=0.99\linewidth]
  {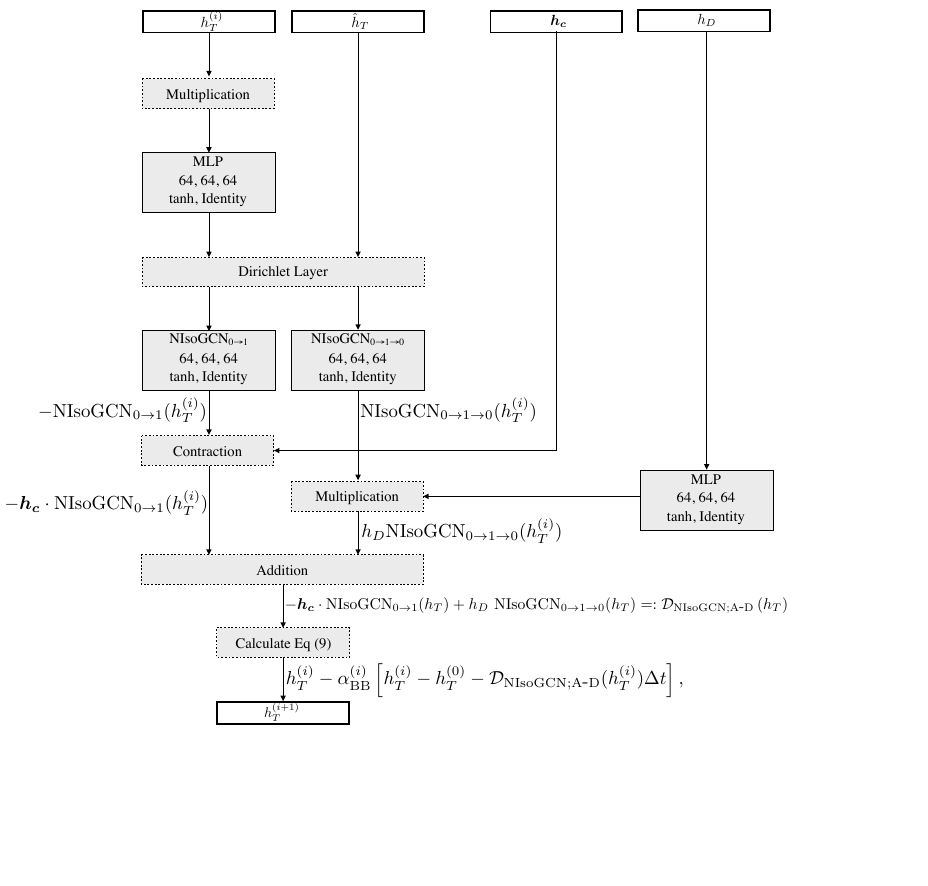}
  \caption{The overview of the PENN architecture for the advection-diffusion dataset.
  Gray boxes with continuous (dotted) lines are trainable (untrainable) components.
  In each cell, we put the number of units in each layer along with the activation functions used.
  }
  \label{fig:ad_network_group1}
\end{figure}

Because the task is to predict time series data, we adopt autoregressive architecture for the nonlinear neural solver, i.e., input the output of the solver of the previous step (which is in the encoded space) to predict the encoded feature of the next step (see \Figref{fig:ad_network_temporal}).
Figures \ref{fig:ad_network_overview} and \ref{fig:ad_network_group1} present the detailed architecture of the PENN model
for the advection-diffusion dataset experiment.

To confirm the PENN's effectiveness, we ran the ablation study similar to that in the incompressible flow dataset. The training is performed for up to ten hours using the Adam optimizer for each setting.

\subsection{Results}
\Tabref{tab:ad} presents the results of the ablation study. As well as the incompressible flow dataset, we found that the PENN model with all the proposed components achieved the best performance. Because the boundary condition applied is relatively simple compared to the incompressible flow dataset, the configuration without the Dirichlet layer (Model (E)) showed the second best performance; however, the fulfillment of the Dirichlet condition of that model is not rigorous.

\begin{figure}[bt]
  \centering

  \stackunder[-45pt]
  {\includegraphics[trim={200cm 10cm 0cm 0cm},clip,width=0.05\textwidth]
  {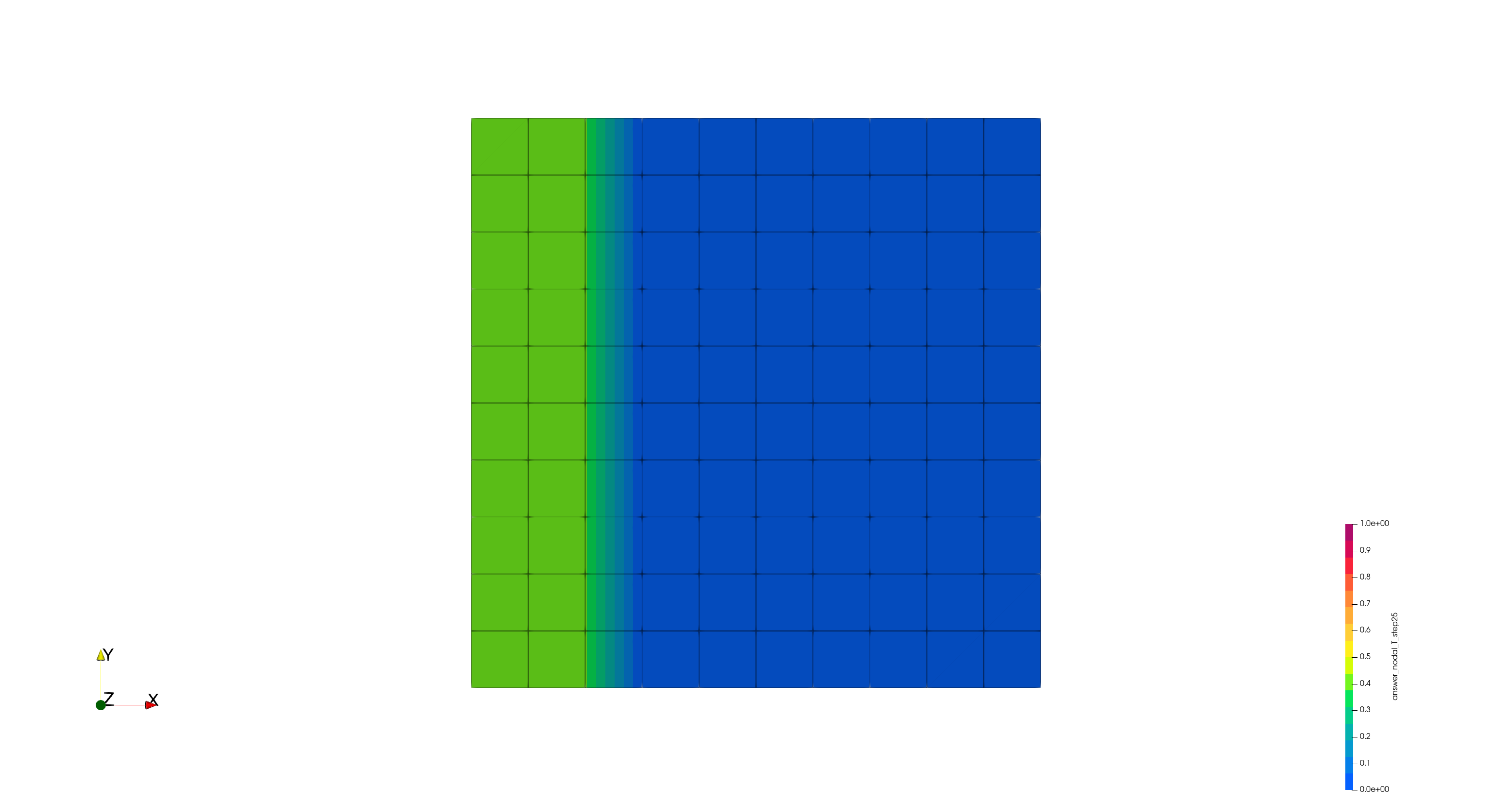}}
  {$t = 0.25$}
  \stackunder[3pt]
  {\includegraphics[trim={27cm 10cm 27cm 0cm},clip,width=0.3\textwidth]
  {figs/ad/penn/u0.9_d0.0_t0.4/0/answer_step25.png}}
  {}
  \hspace{5pt}
  \stackunder[3pt]
  {\includegraphics[trim={27cm 10cm 27cm 0cm},clip,width=0.3\textwidth]
  {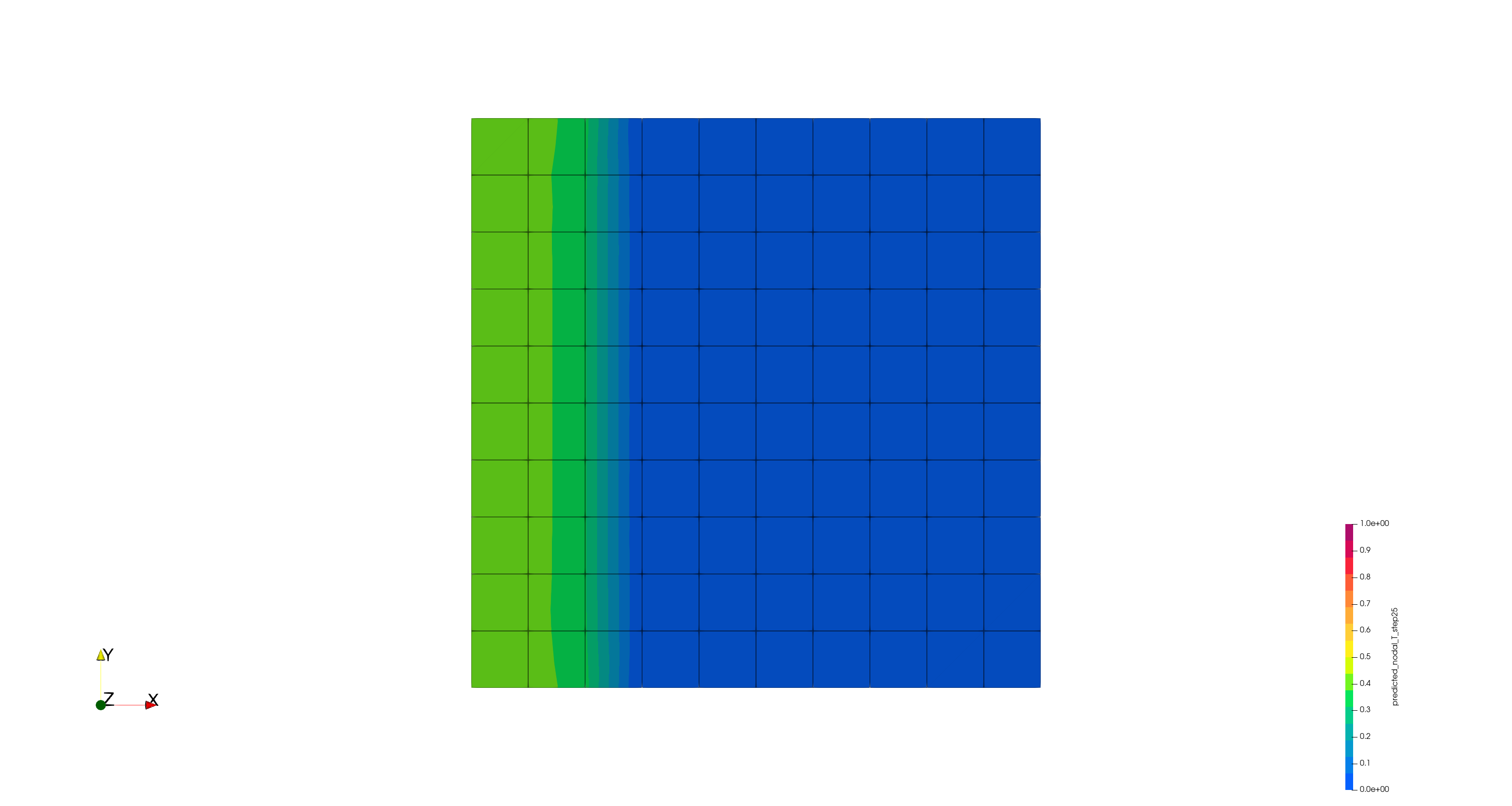}}
  {}
  \\
  \stackunder[-45pt]
  {\includegraphics[trim={200cm 10cm 0cm 0cm},clip,width=0.05\textwidth]
  {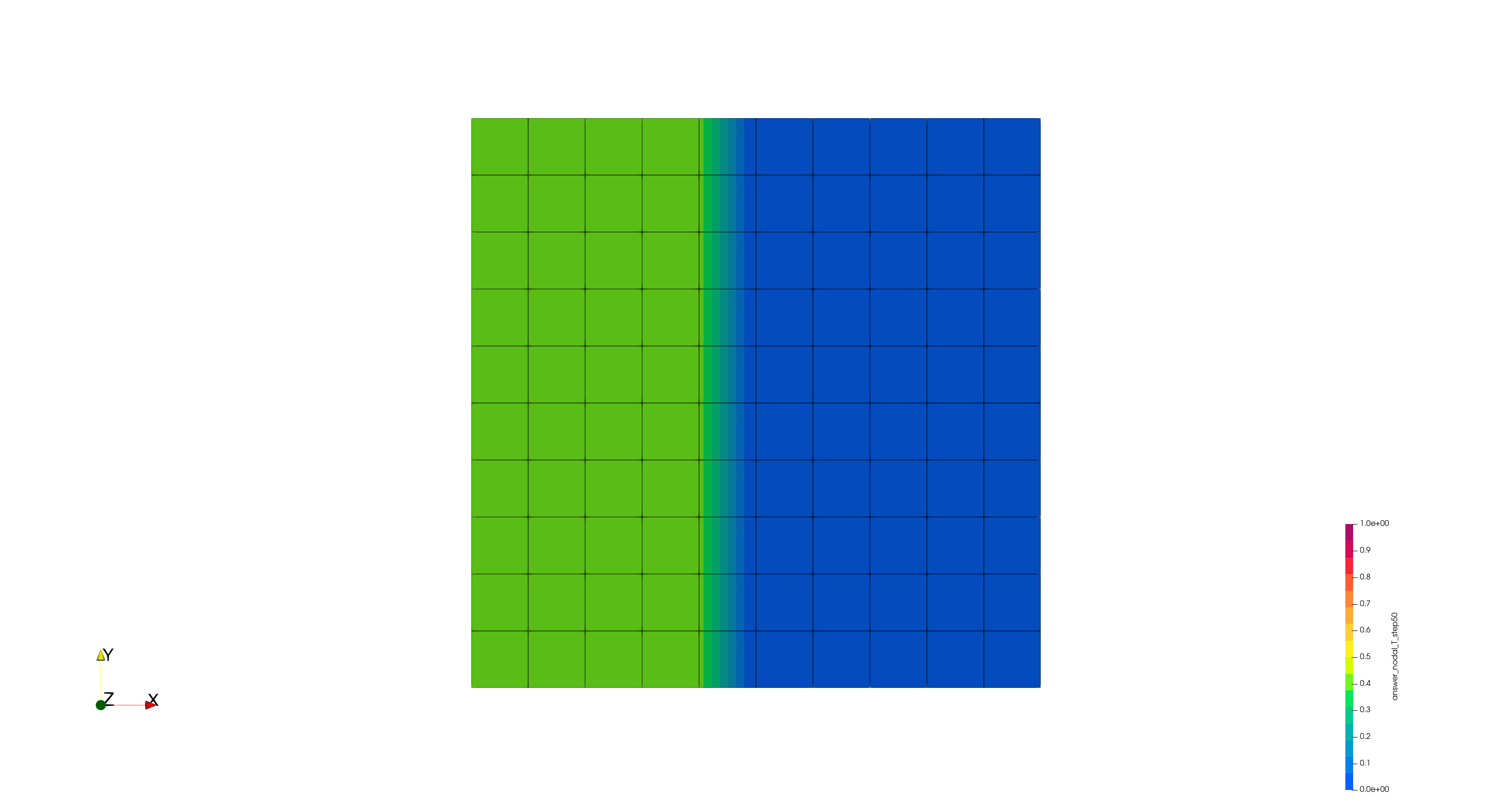}}
  {$t = 0.50$}
  \stackunder[3pt]
  {\includegraphics[trim={27cm 10cm 27cm 0cm},clip,width=0.3\textwidth]
  {figs/ad/penn/u0.9_d0.0_t0.4/0/answer_step50.png}}
  {}
  \hspace{5pt}
  \stackunder[3pt]
  {\includegraphics[trim={27cm 10cm 27cm 0cm},clip,width=0.3\textwidth]
  {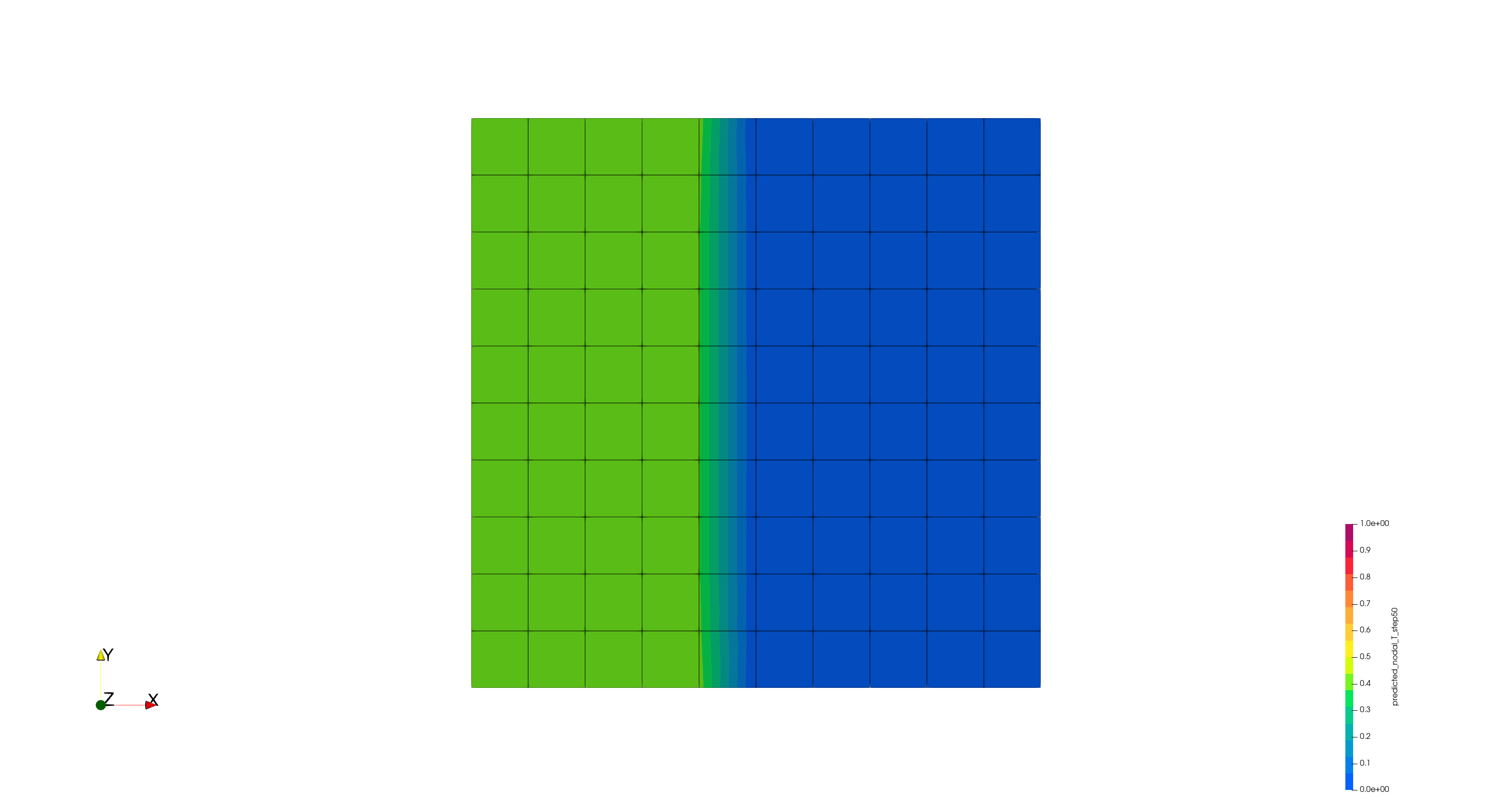}}
  {}
  \\
  \stackunder[-45pt]
  {\includegraphics[trim={200cm 10cm 0cm 0cm},clip,width=0.05\textwidth]
  {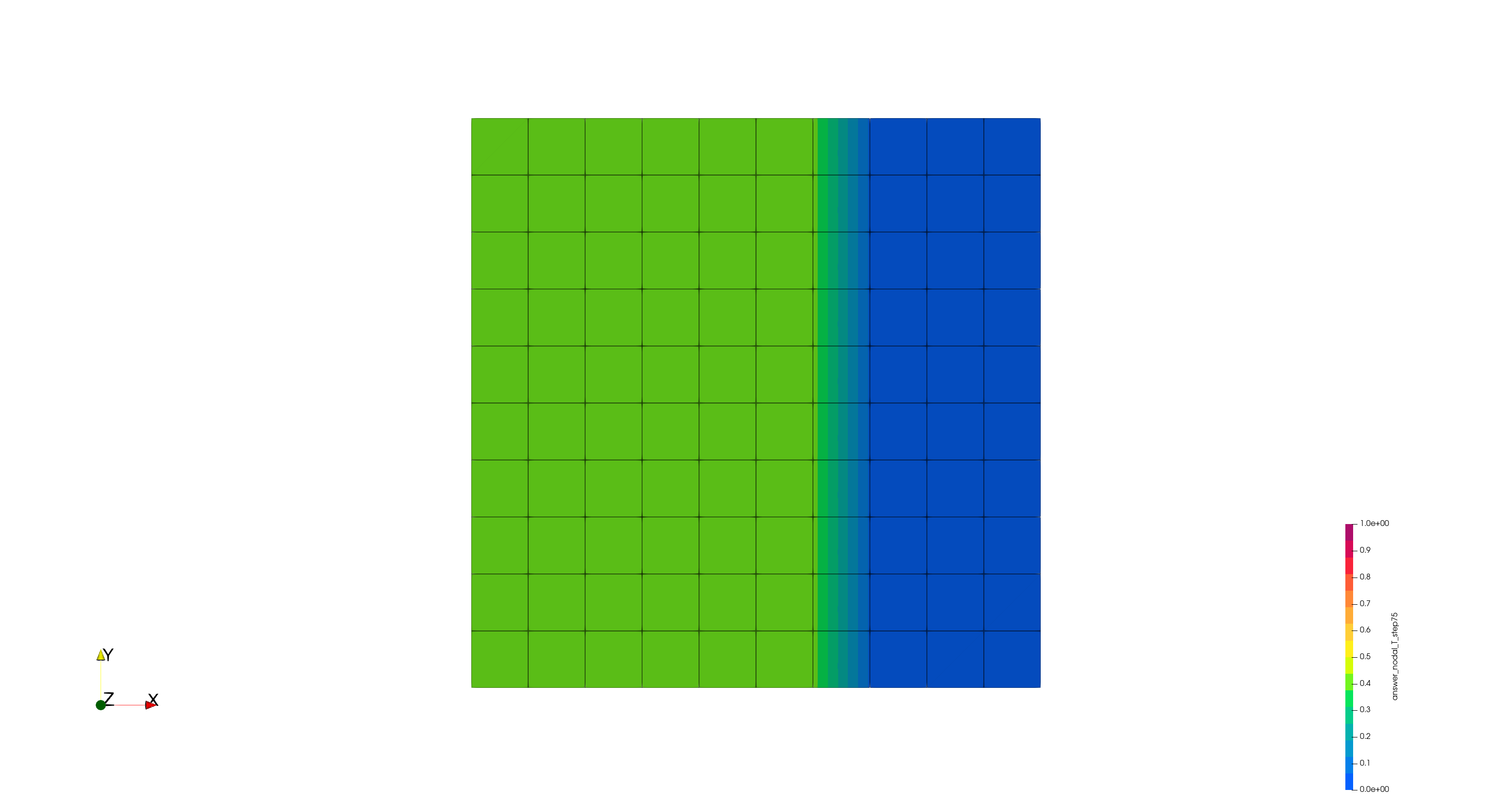}}
  {$t = 0.75$}
  \stackunder[3pt]
  {\includegraphics[trim={27cm 10cm 27cm 0cm},clip,width=0.3\textwidth]
  {figs/ad/penn/u0.9_d0.0_t0.4/0/answer_step75.png}}
  {}
  \hspace{5pt}
  \stackunder[3pt]
  {\includegraphics[trim={27cm 10cm 27cm 0cm},clip,width=0.3\textwidth]
  {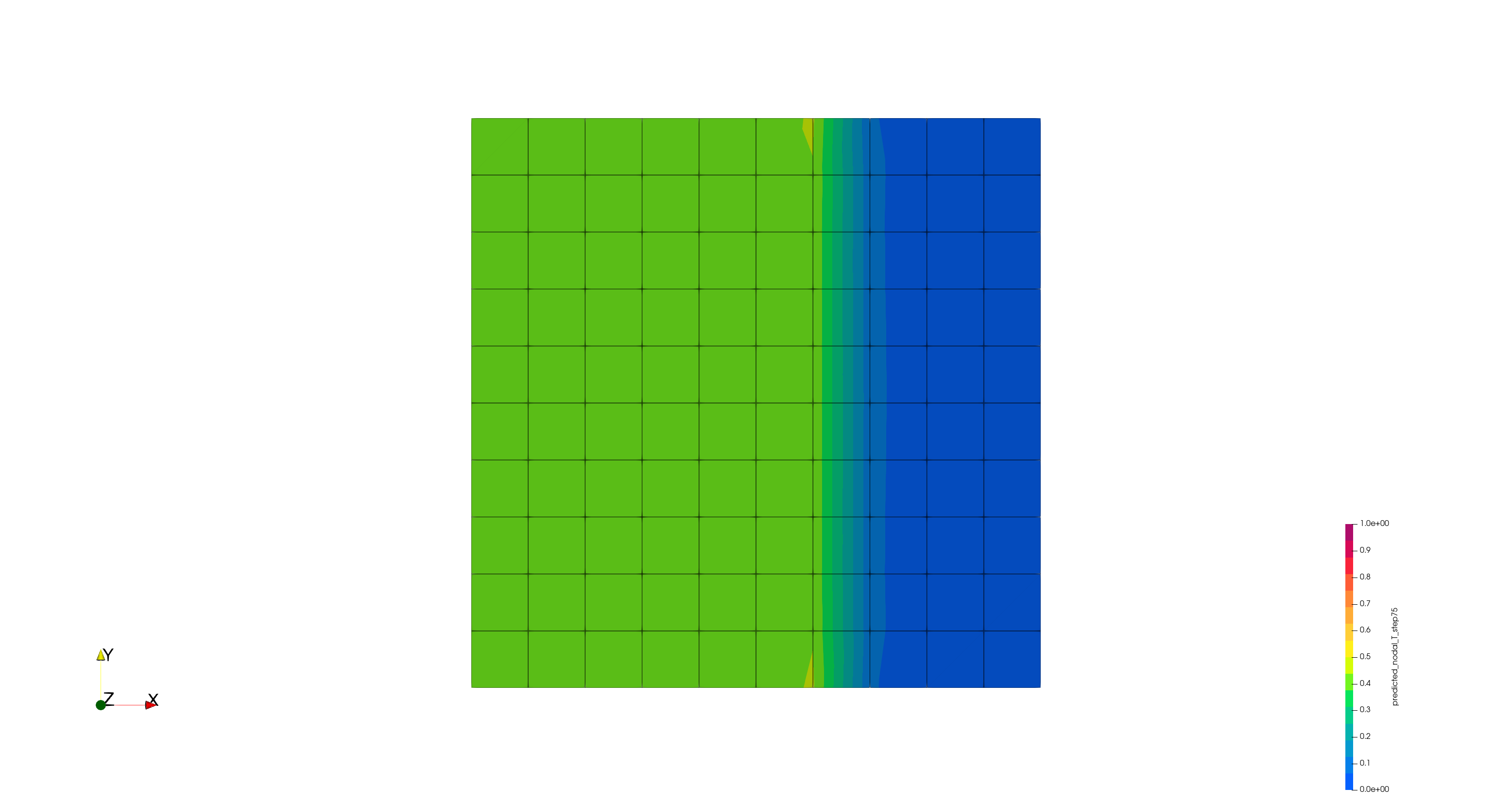}}
  {}
  \\
  \stackunder[-45pt]
  {\includegraphics[trim={200cm 10cm 0cm 0cm},clip,width=0.05\textwidth]
  {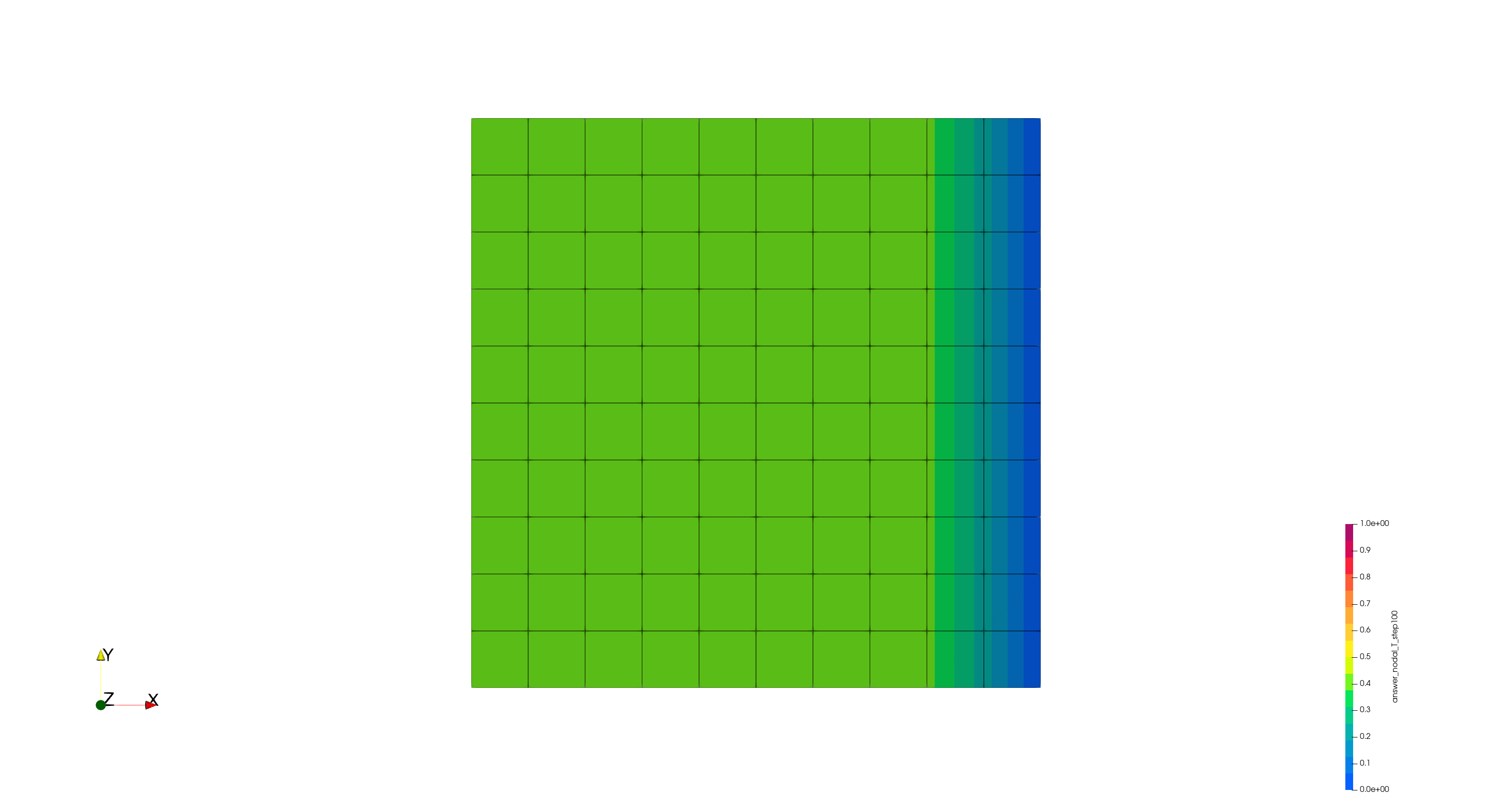}}
  {$t = 1.00$}
  \stackunder[3pt]
  {\includegraphics[trim={27cm 10cm 27cm 0cm},clip,width=0.3\textwidth]
  {figs/ad/penn/u0.9_d0.0_t0.4/0/answer_step100.png}}
  {}
  \hspace{5pt}
  \stackunder[3pt]
  {\includegraphics[trim={27cm 10cm 27cm 0cm},clip,width=0.3\textwidth]
  {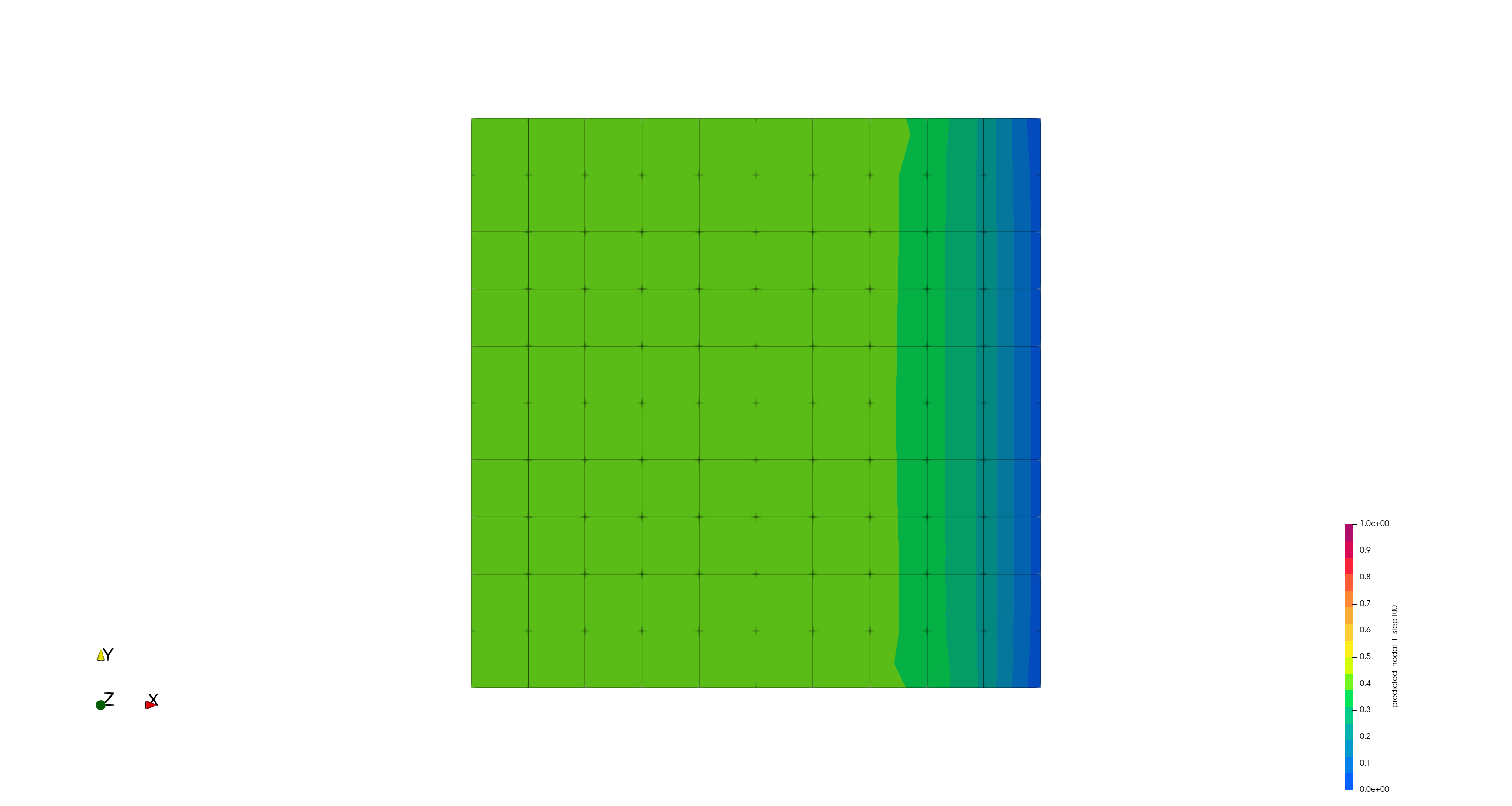}}
  {}
  \\
  {\includegraphics[trim={200cm 10cm 0cm 0cm},clip,width=0.05\textwidth]
  {figs/ad/penn/u0.9_d0.0_t0.4/0/answer_step25.png}}
  {}
  {\includegraphics[trim={0cm 0cm 0cm 0cm},clip,width=0.2\textwidth]
  {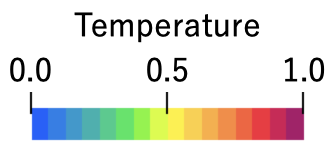}}
  \caption{Visual comparison on a test sample between
  (left) ground truth obtained from OpenFOAM computation with fine spatial-temporal resolution and
  (right) prediction by PENN.
  Here, $c =0.9$, $D = 0.0$, and $\hat{T} = 0.4$.
  }
  \label{fig:ad_advection}
\end{figure}

\begin{figure}[bt]
  \centering

  \stackunder[-45pt]
  {\includegraphics[trim={200cm 10cm 0cm 0cm},clip,width=0.05\textwidth]
  {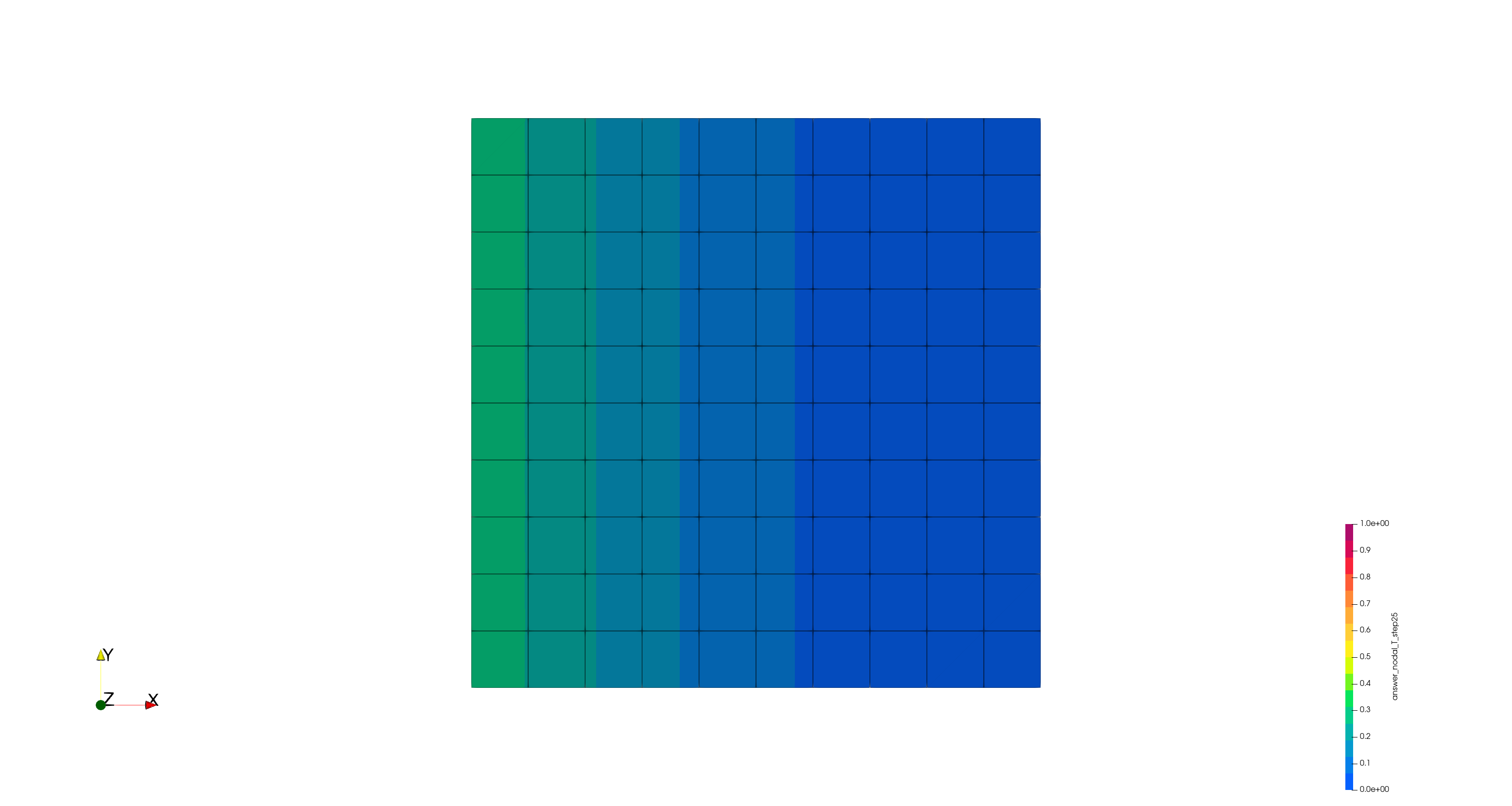}}
  {$t = 0.25$}
  \stackunder[3pt]
  {\includegraphics[trim={27cm 10cm 27cm 0cm},clip,width=0.3\textwidth]
  {figs/ad/penn/u0.0_d0.4_t0.3/0/answer_step25.png}}
  {}
  \hspace{5pt}
  \stackunder[3pt]
  {\includegraphics[trim={27cm 10cm 27cm 0cm},clip,width=0.3\textwidth]
  {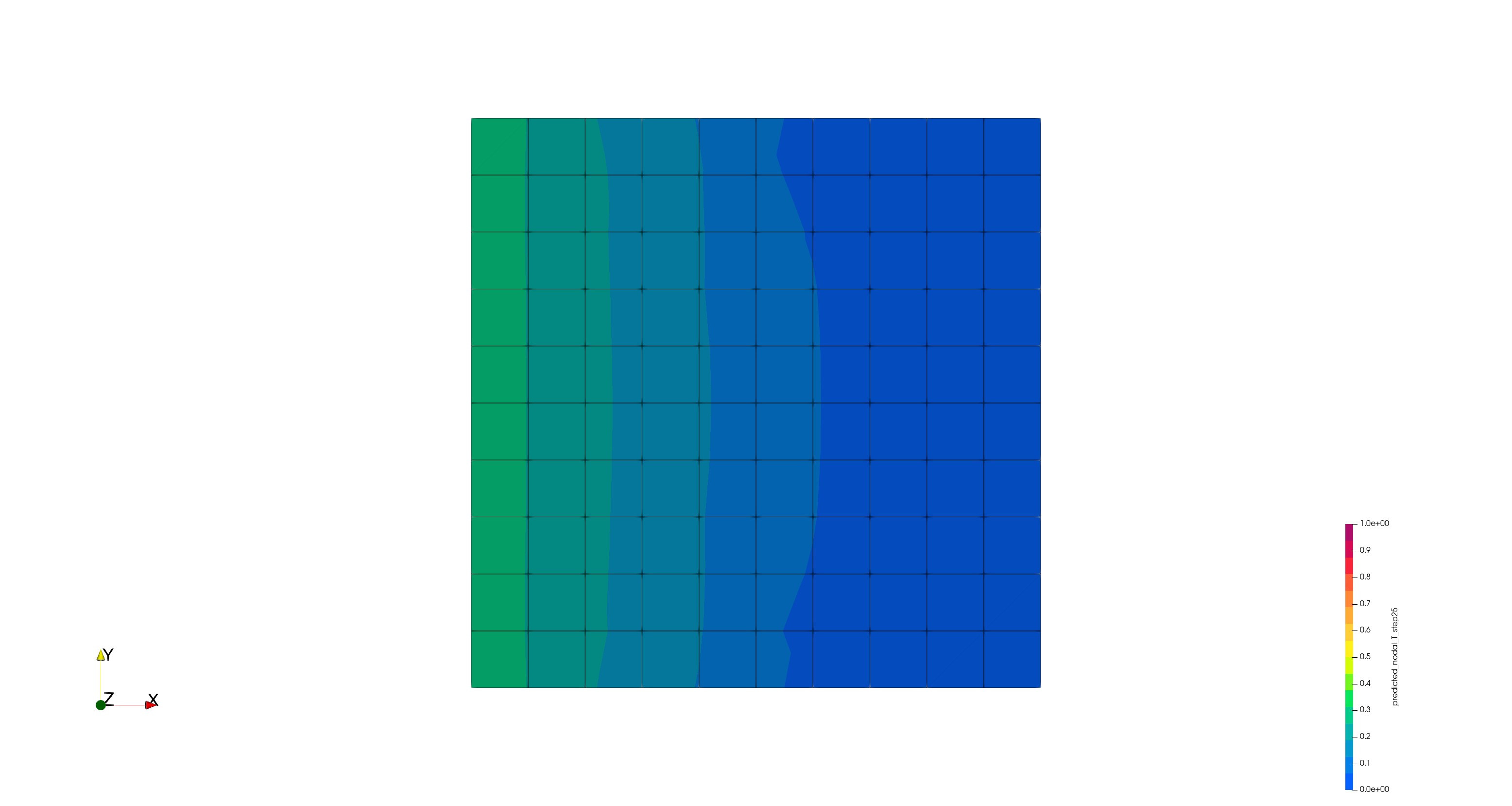}}
  {}
  \\
  \stackunder[-45pt]
  {\includegraphics[trim={200cm 10cm 0cm 0cm},clip,width=0.05\textwidth]
  {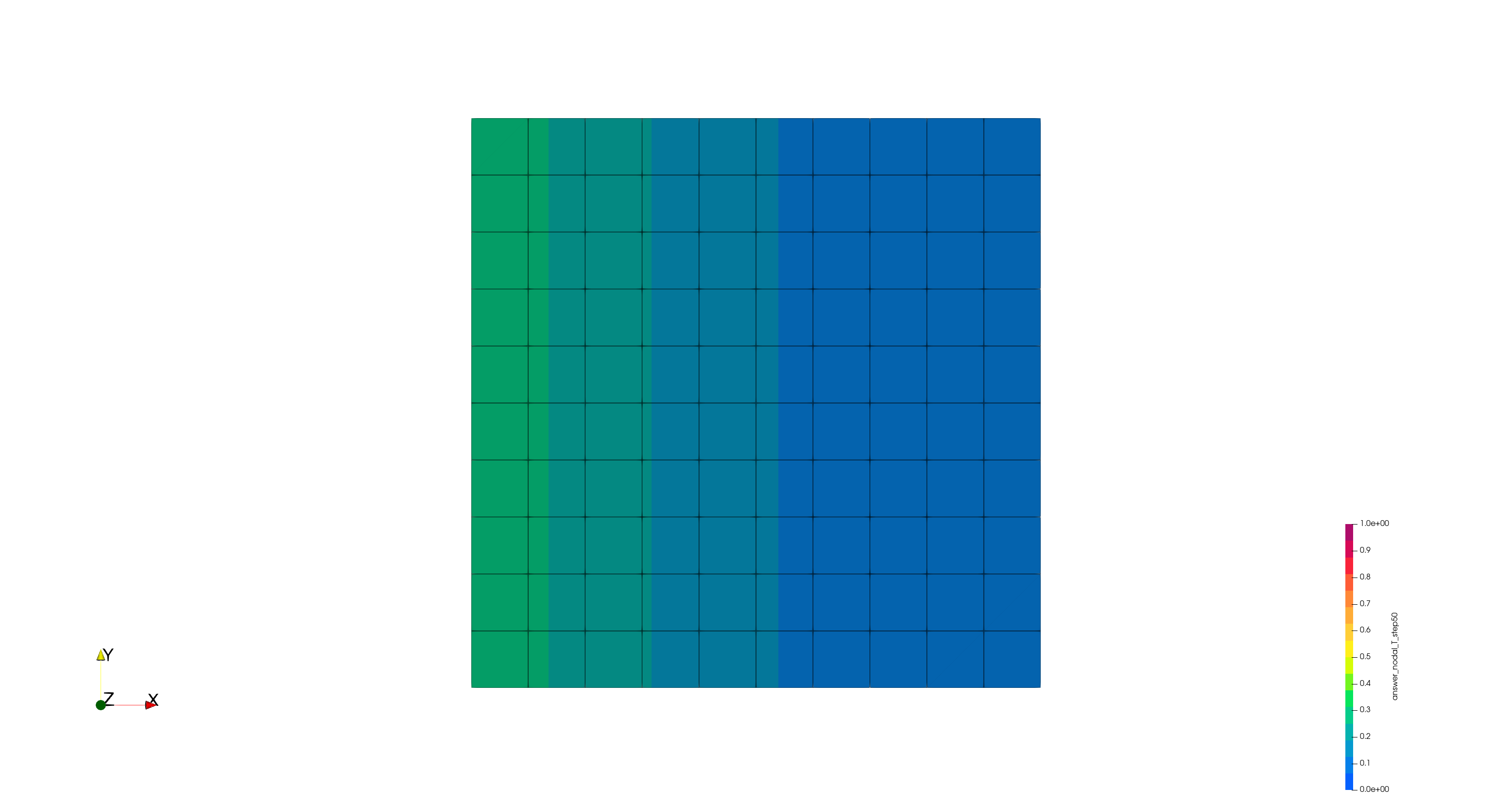}}
  {$t = 0.50$}
  \stackunder[3pt]
  {\includegraphics[trim={27cm 10cm 27cm 0cm},clip,width=0.3\textwidth]
  {figs/ad/penn/u0.0_d0.4_t0.3/0/answer_step50.png}}
  {}
  \hspace{5pt}
  \stackunder[3pt]
  {\includegraphics[trim={27cm 10cm 27cm 0cm},clip,width=0.3\textwidth]
  {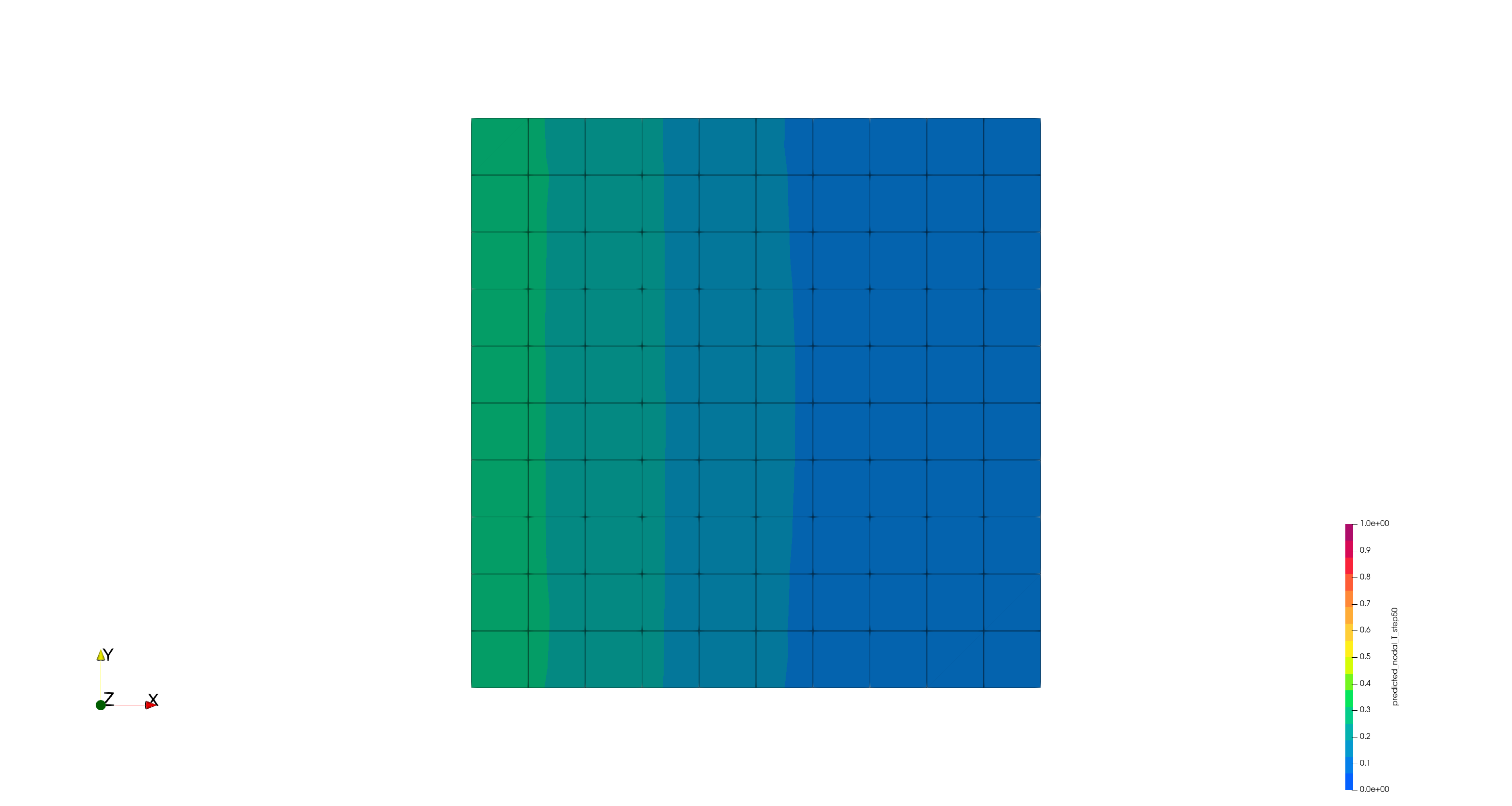}}
  {}
  \\
  \stackunder[-45pt]
  {\includegraphics[trim={200cm 10cm 0cm 0cm},clip,width=0.05\textwidth]
  {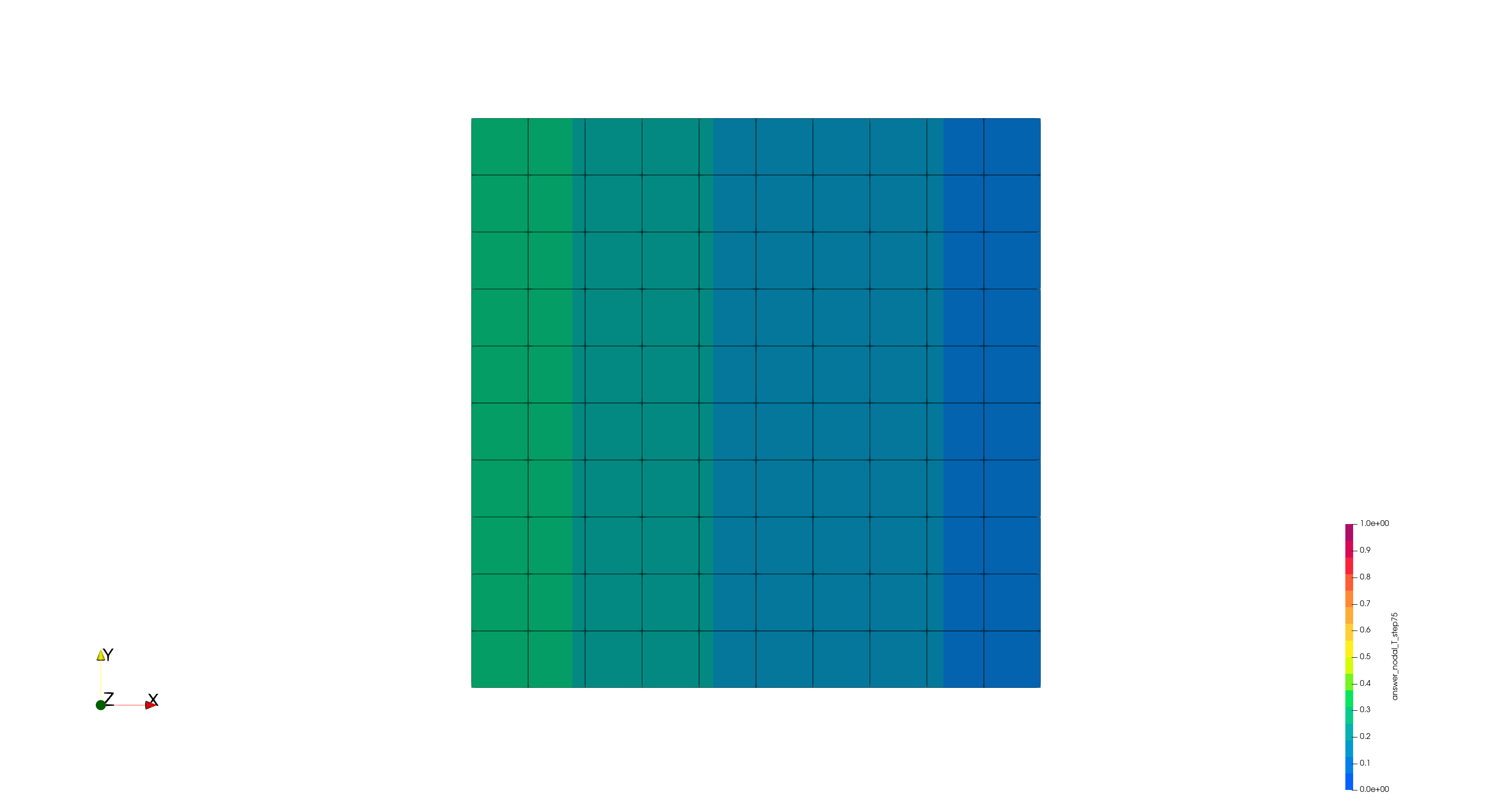}}
  {$t = 0.75$}
  \stackunder[3pt]
  {\includegraphics[trim={27cm 10cm 27cm 0cm},clip,width=0.3\textwidth]
  {figs/ad/penn/u0.0_d0.4_t0.3/0/answer_step75.png}}
  {}
  \hspace{5pt}
  \stackunder[3pt]
  {\includegraphics[trim={27cm 10cm 27cm 0cm},clip,width=0.3\textwidth]
  {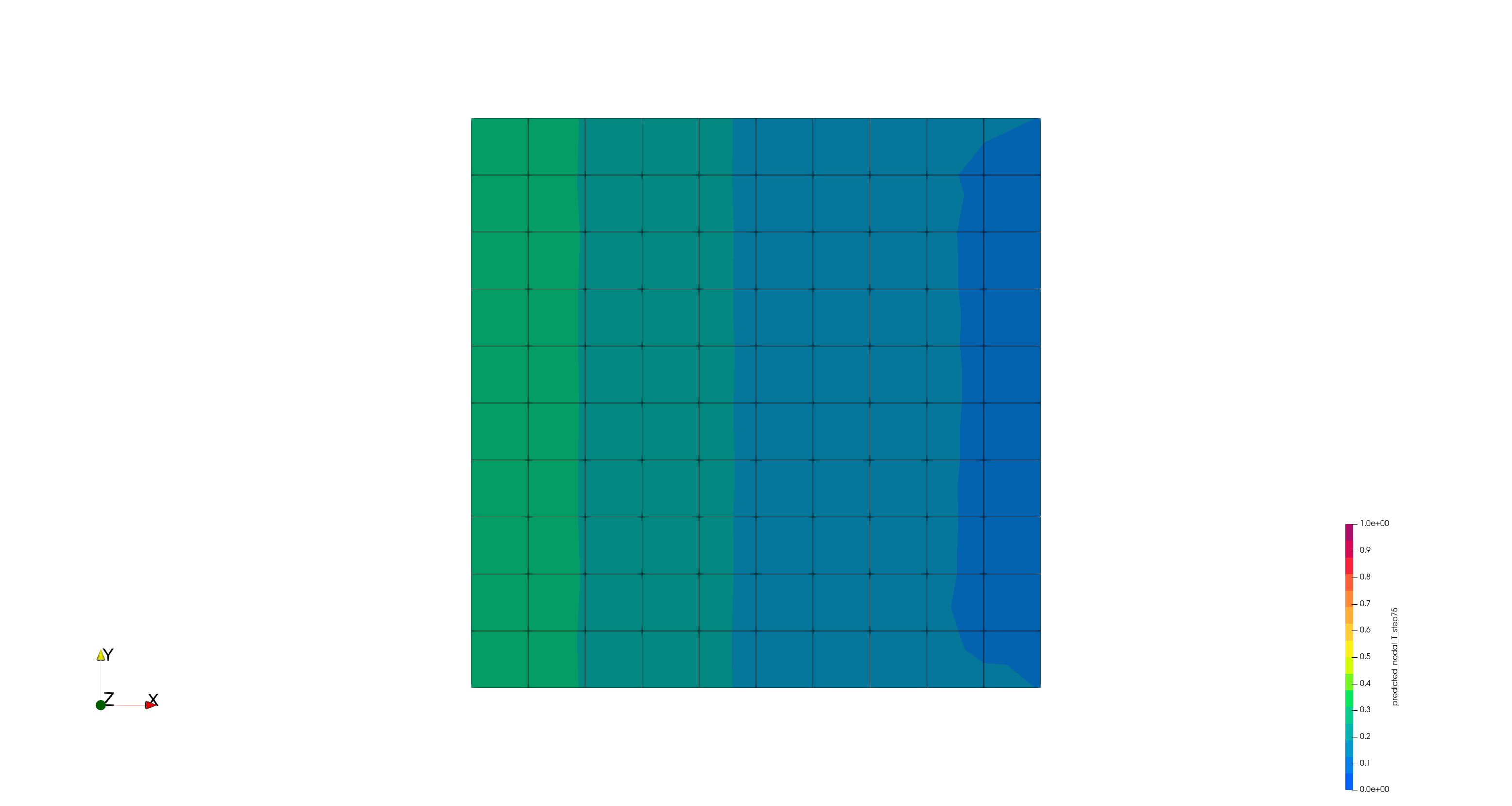}}
  {}
  \\
  \stackunder[-45pt]
  {\includegraphics[trim={200cm 10cm 0cm 0cm},clip,width=0.05\textwidth]
  {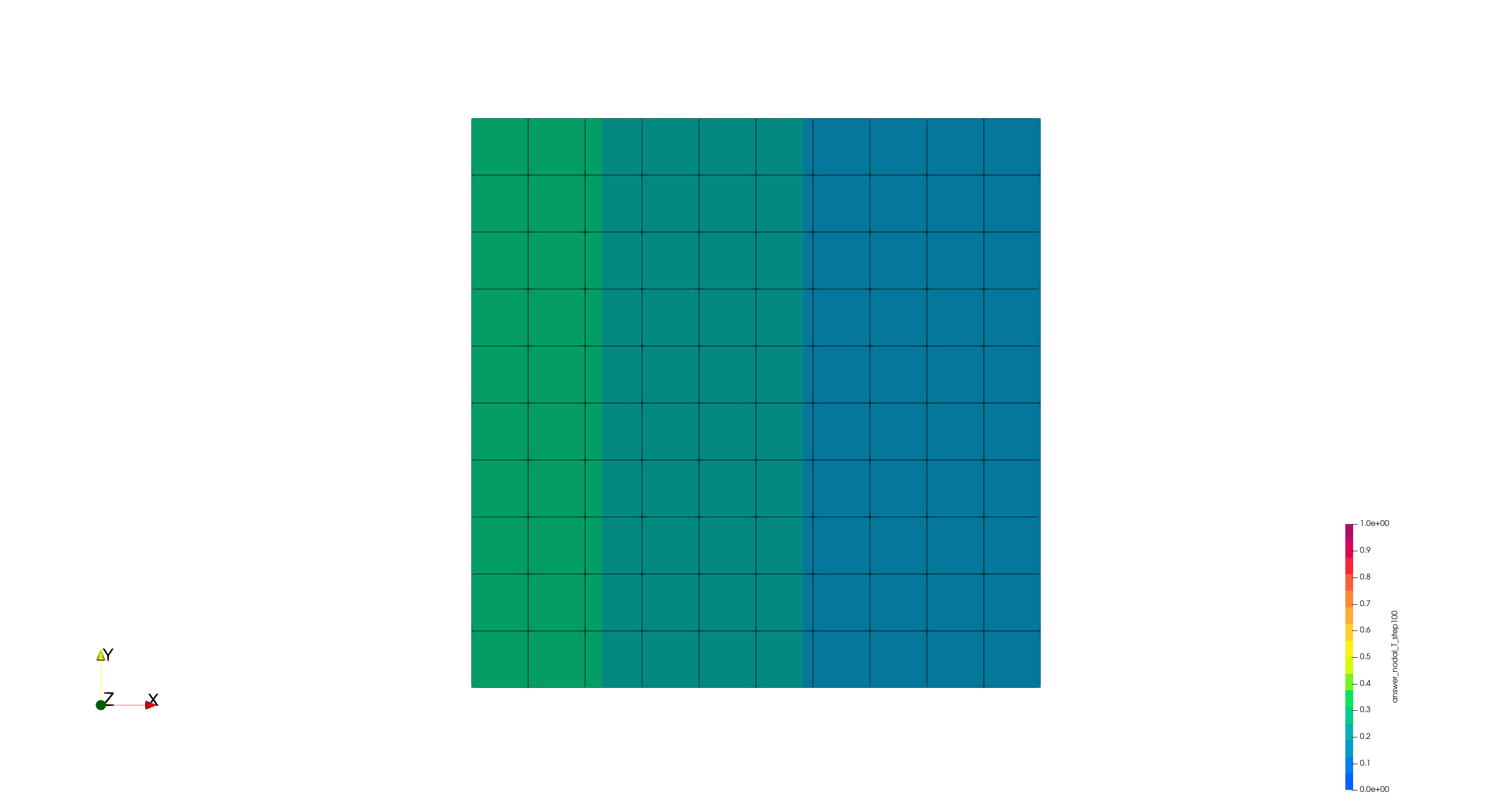}}
  {$t = 1.00$}
  \stackunder[3pt]
  {\includegraphics[trim={27cm 10cm 27cm 0cm},clip,width=0.3\textwidth]
  {figs/ad/penn/u0.0_d0.4_t0.3/0/answer_step100.png}}
  {}
  \hspace{5pt}
  \stackunder[3pt]
  {\includegraphics[trim={27cm 10cm 27cm 0cm},clip,width=0.3\textwidth]
  {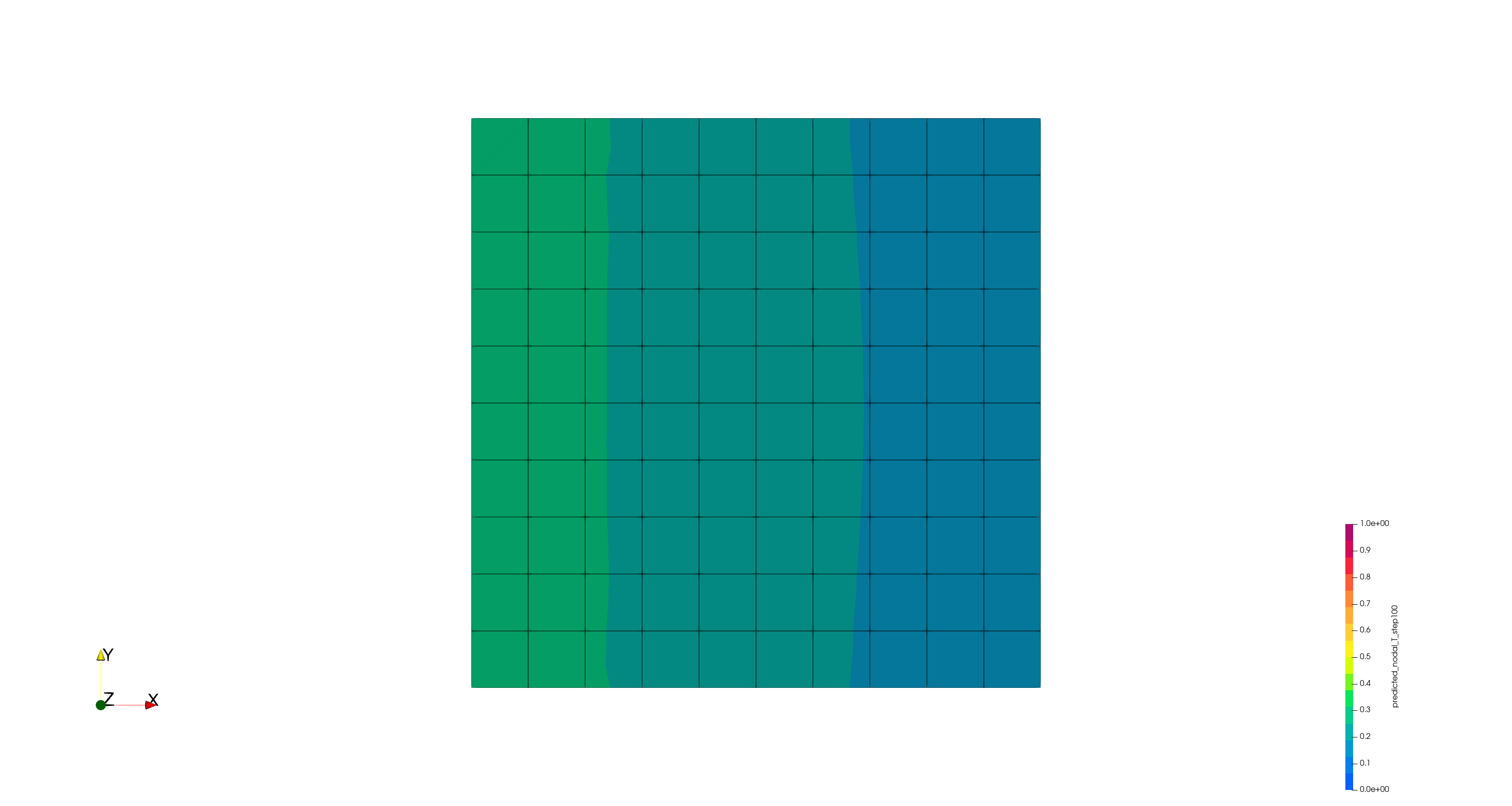}}
  {}
  \\
  {\includegraphics[trim={200cm 10cm 0cm 0cm},clip,width=0.05\textwidth]
  {figs/ad/penn/u0.0_d0.4_t0.3/0/answer_step25.png}}
  {}
  {\includegraphics[trim={0cm 0cm 0cm 0cm},clip,width=0.2\textwidth]
  {figs/ad/ad_colorbar_horizontal.png}}
  \caption{Visual comparison on a test sample between
  (left) ground truth obtained from OpenFOAM computation with fine spatial-temporal resolution and
  (right) prediction by PENN.
  Here, $c =0.0$, $D = 0.4$, and $\hat{T} = 0.3$.
  }
  \label{fig:ad_diffusion}
\end{figure}

\begin{figure}[bt]
  \centering

  \stackunder[-45pt]
  {\includegraphics[trim={200cm 10cm 0cm 0cm},clip,width=0.05\textwidth]
  {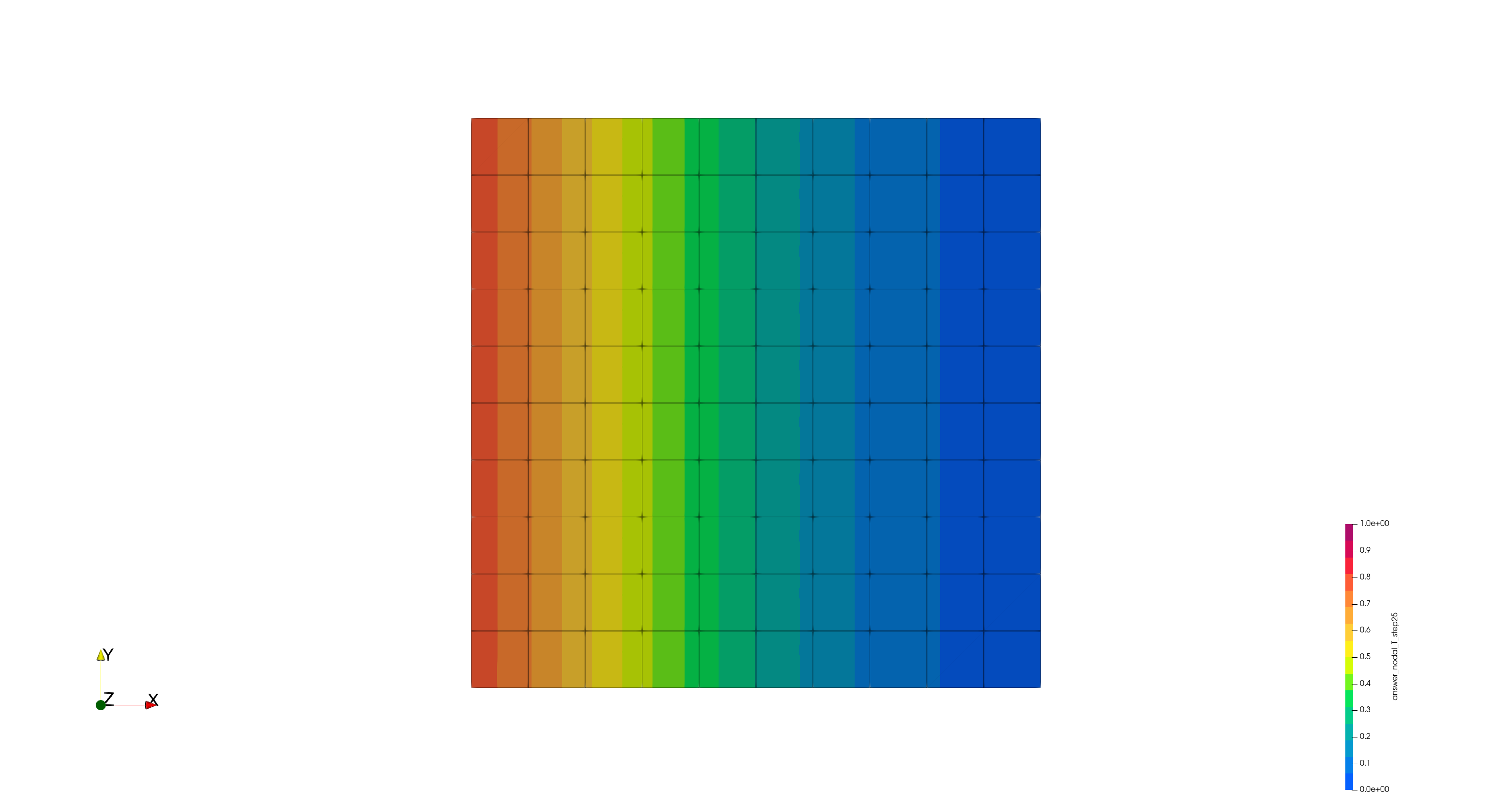}}
  {$t = 0.25$}
  \stackunder[3pt]
  {\includegraphics[trim={27cm 10cm 27cm 0cm},clip,width=0.3\textwidth]
  {figs/ad/penn/u0.6_d0.3_t0.8/0/answer_step25.png}}
  {}
  \hspace{5pt}
  \stackunder[3pt]
  {\includegraphics[trim={27cm 10cm 27cm 0cm},clip,width=0.3\textwidth]
  {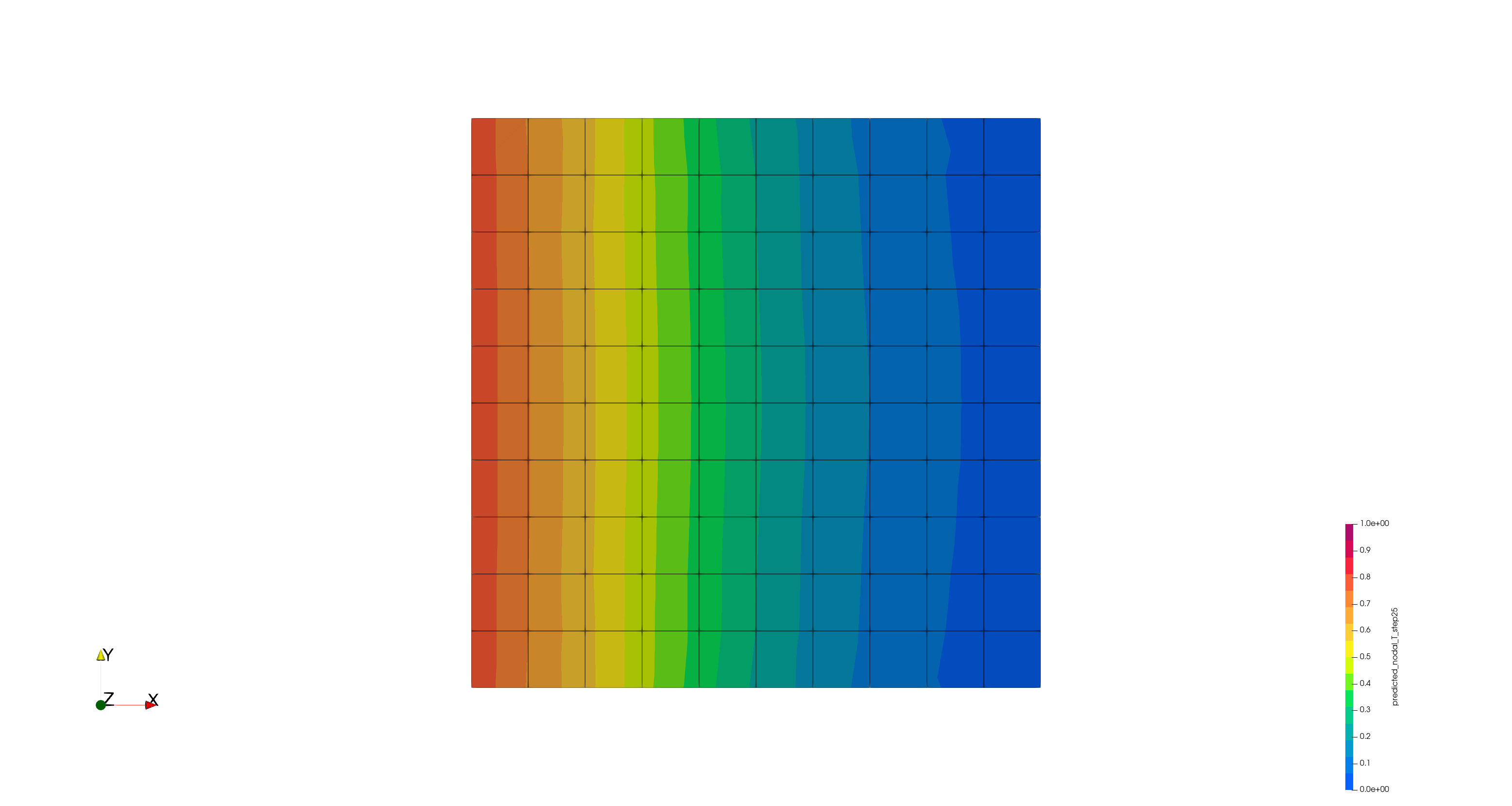}}
  {}
  \\
  \stackunder[-45pt]
  {\includegraphics[trim={200cm 10cm 0cm 0cm},clip,width=0.05\textwidth]
  {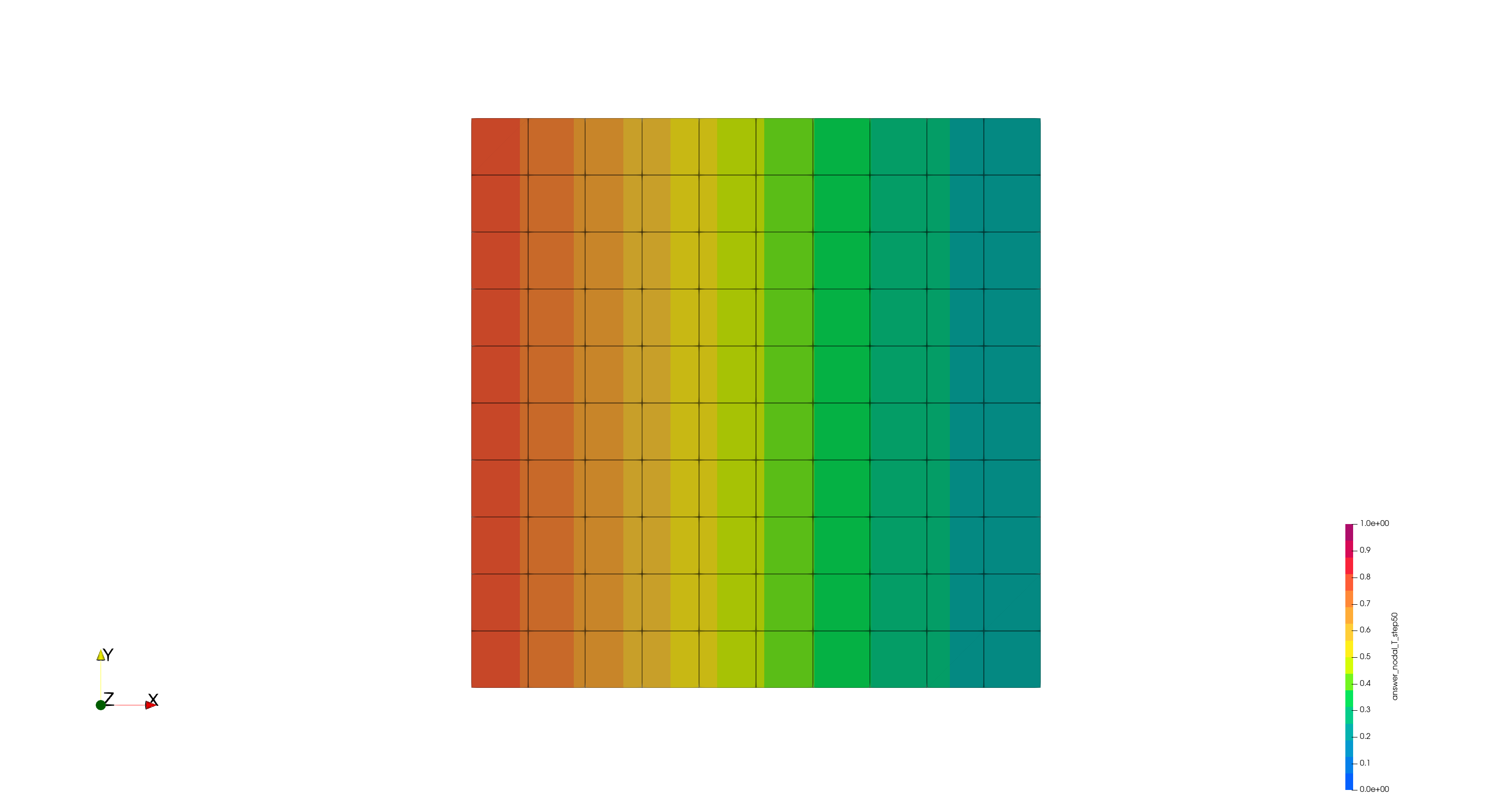}}
  {$t = 0.50$}
  \stackunder[3pt]
  {\includegraphics[trim={27cm 10cm 27cm 0cm},clip,width=0.3\textwidth]
  {figs/ad/penn/u0.6_d0.3_t0.8/0/answer_step50.png}}
  {}
  \hspace{5pt}
  \stackunder[3pt]
  {\includegraphics[trim={27cm 10cm 27cm 0cm},clip,width=0.3\textwidth]
  {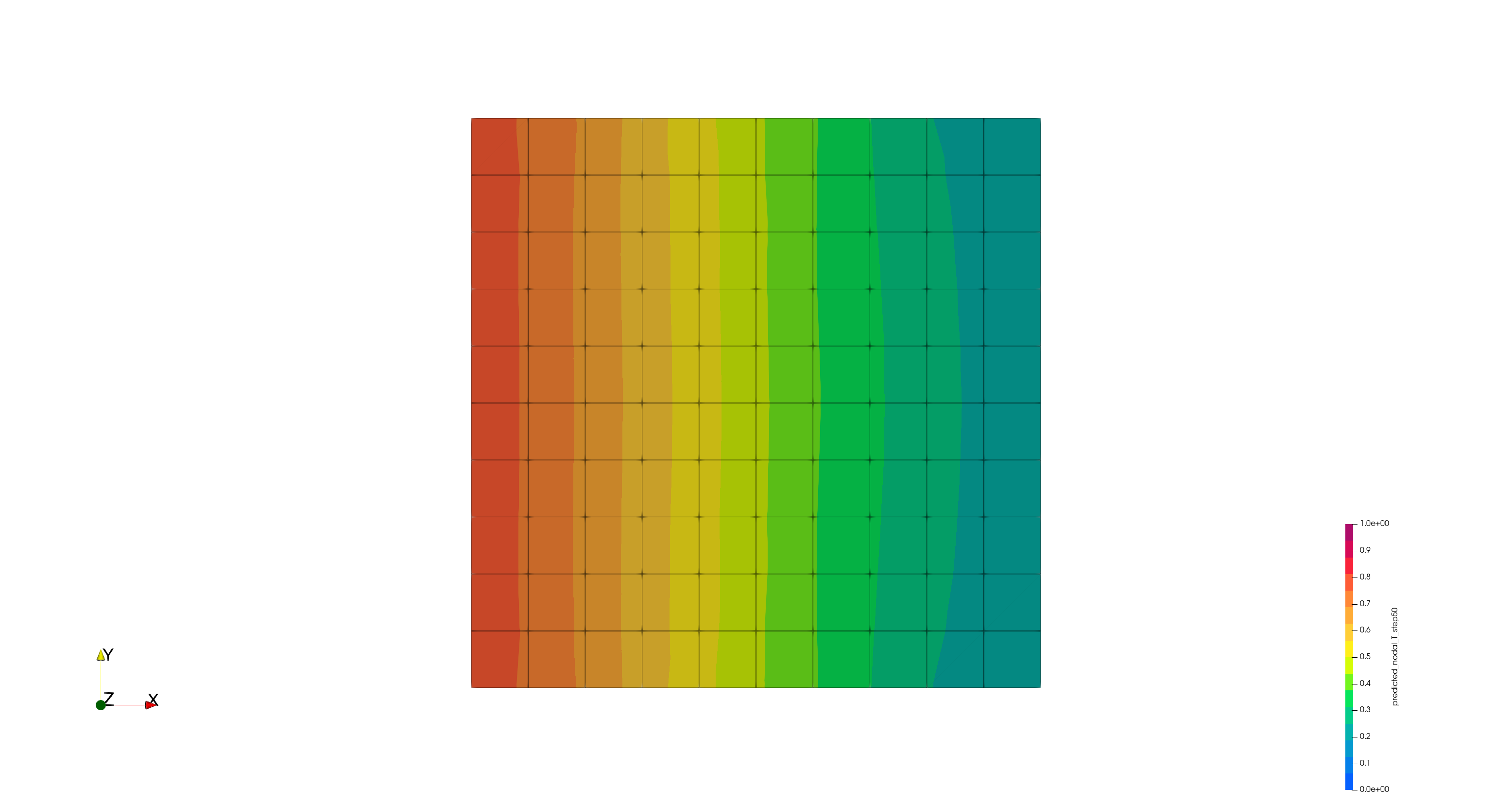}}
  {}
  \\
  \stackunder[-45pt]
  {\includegraphics[trim={200cm 10cm 0cm 0cm},clip,width=0.05\textwidth]
  {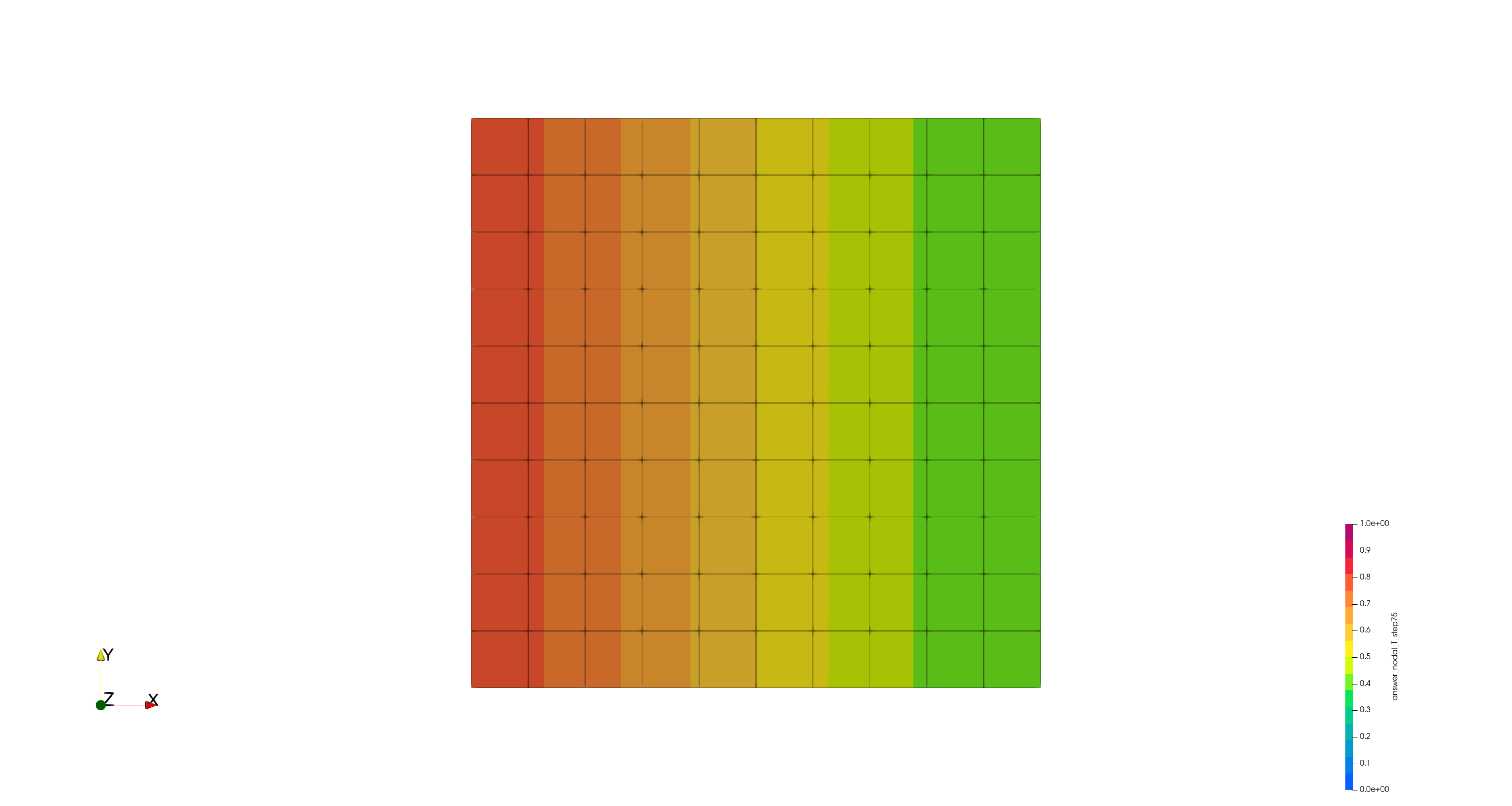}}
  {$t = 0.75$}
  \stackunder[3pt]
  {\includegraphics[trim={27cm 10cm 27cm 0cm},clip,width=0.3\textwidth]
  {figs/ad/penn/u0.6_d0.3_t0.8/0/answer_step75.png}}
  {}
  \hspace{5pt}
  \stackunder[3pt]
  {\includegraphics[trim={27cm 10cm 27cm 0cm},clip,width=0.3\textwidth]
  {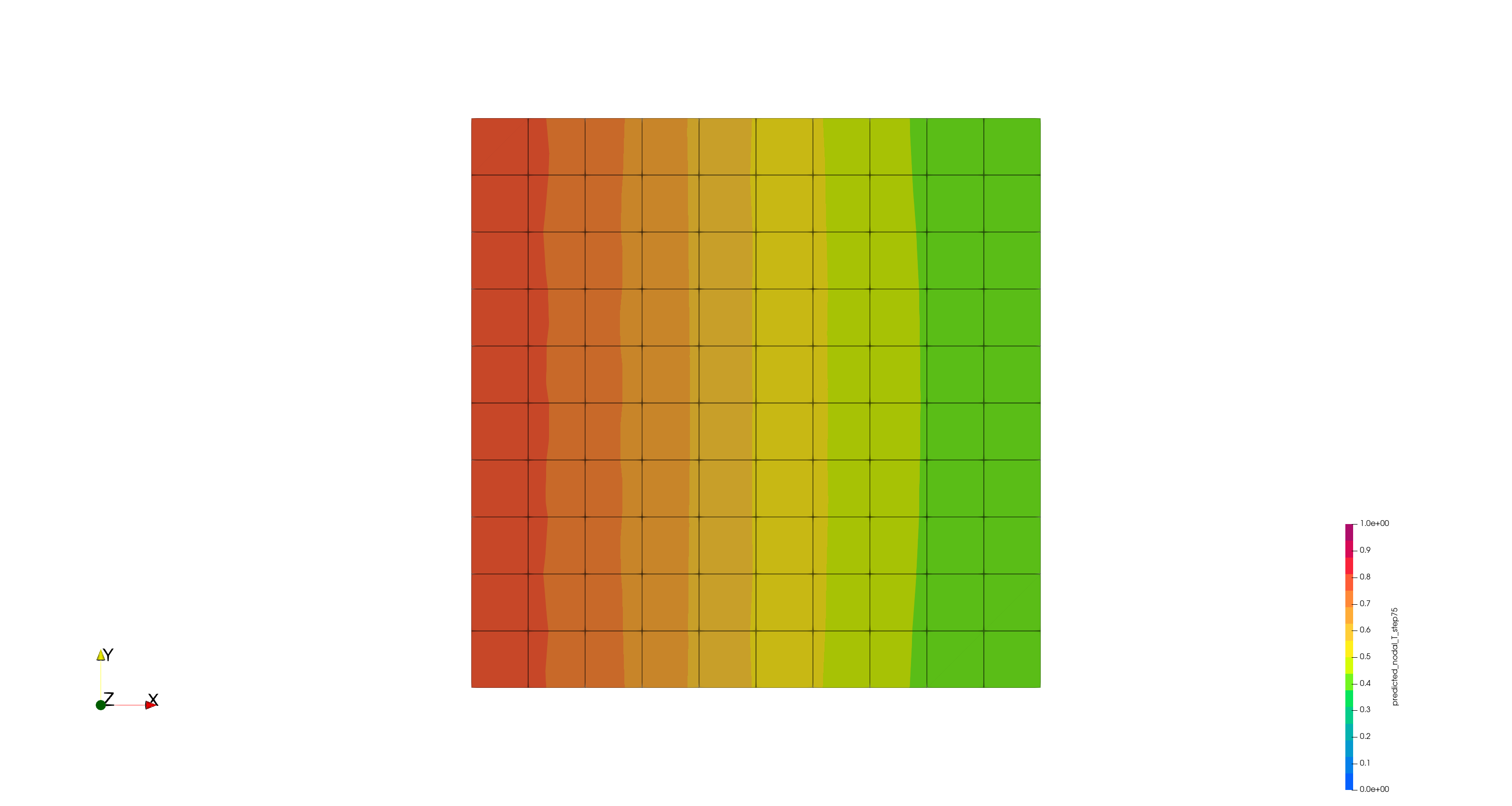}}
  {}
  \\
  \stackunder[-45pt]
  {\includegraphics[trim={200cm 10cm 0cm 0cm},clip,width=0.05\textwidth]
  {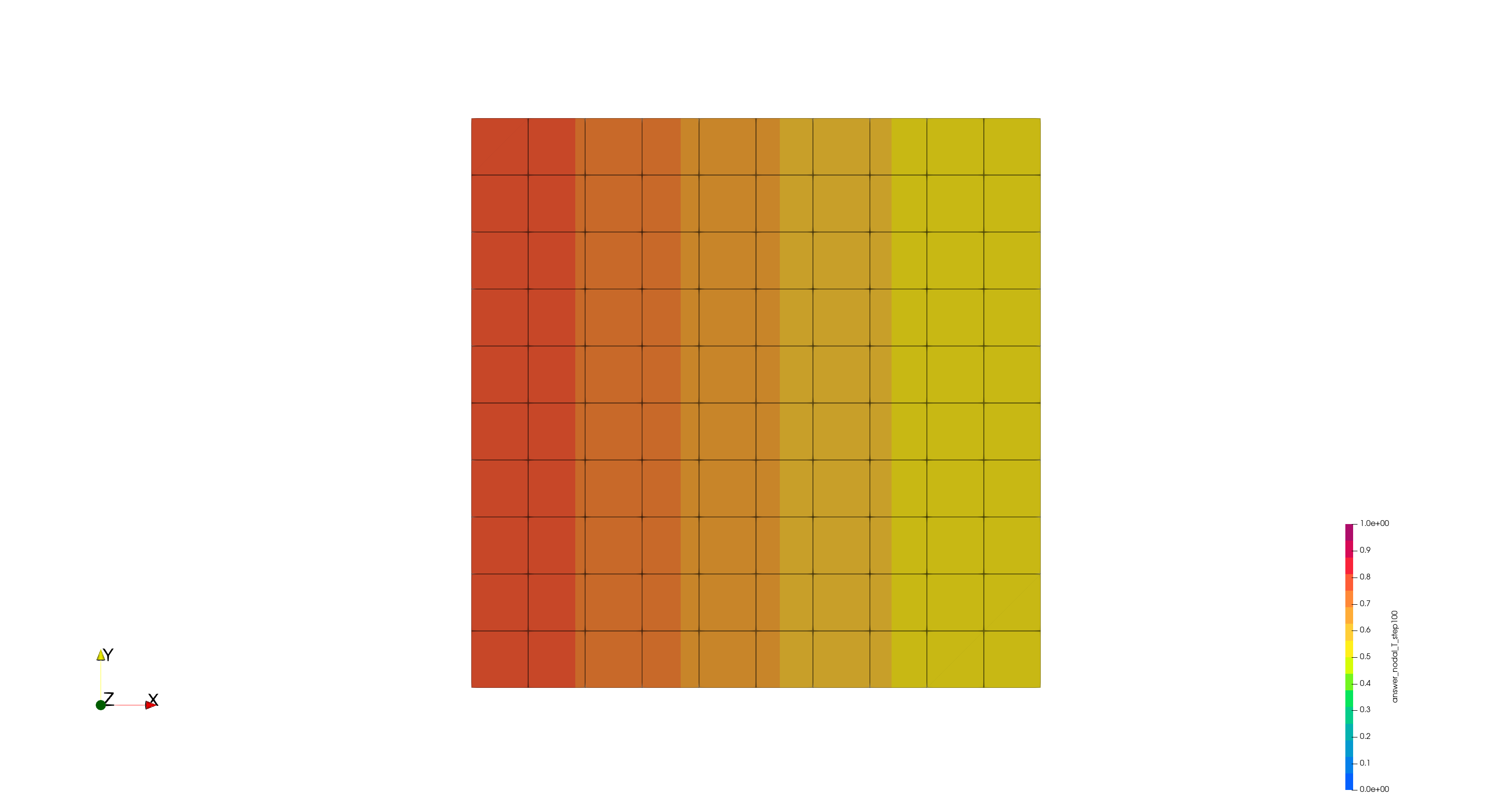}}
  {$t = 1.00$}
  \stackunder[3pt]
  {\includegraphics[trim={27cm 10cm 27cm 0cm},clip,width=0.3\textwidth]
  {figs/ad/penn/u0.6_d0.3_t0.8/0/answer_step100.png}}
  {}
  \hspace{5pt}
  \stackunder[3pt]
  {\includegraphics[trim={27cm 10cm 27cm 0cm},clip,width=0.3\textwidth]
  {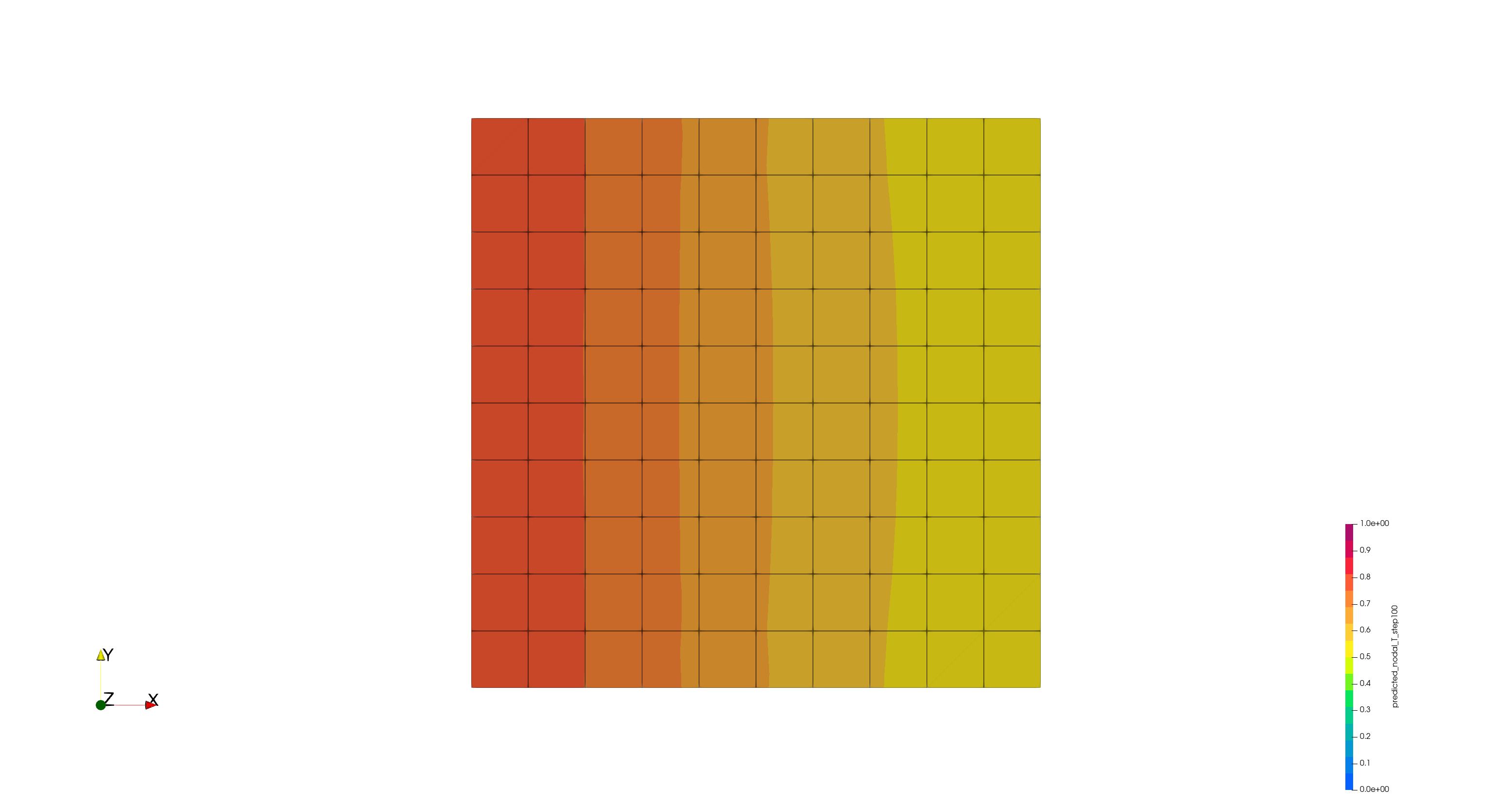}}
  {}
  \\
  {\includegraphics[trim={200cm 10cm 0cm 0cm},clip,width=0.05\textwidth]
  {figs/ad/penn/u0.6_d0.3_t0.8/0/answer_step25.png}}
  {}
  {\includegraphics[trim={0cm 0cm 0cm 0cm},clip,width=0.2\textwidth]
  {figs/ad/ad_colorbar_horizontal.png}}
  \caption{Visual comparison on a test sample between
  (left) ground truth obtained from OpenFOAM computation with fine spatial-temporal resolution and
  (right) prediction by PENN.
  Here, $c =0.6$, $D = 0.3$, and $\hat{T} = 0.8$.
  }
  \label{fig:ad_advection_diffusion}
\end{figure}

Figures
\ref{fig:ad_advection},
\ref{fig:ad_diffusion},
and
\ref{fig:ad_advection_diffusion}
show the visual comparison of the prediction with the PENN model against the ground truth. As seen in the figures, one can see that our model is capable of predicting time series under various boundary conditions and PDE parameters, e.g.,
pure advection (\Figref{fig:ad_advection}),
pure diffusion (\Figref{fig:ad_diffusion}),
and mixed advection and diffusion (\Figref{fig:ad_advection_diffusion}).

\bgroup
\def\arraystretch{1.3}
\begin{table}[bt]
  \caption{MSE loss ($\pm$ the standard error of the mean)
  on test dataset of the advection-diffusion dataset.
  }
  \label{tab:ad}
  \centering
  \scalebox{1.0}{
    \begin{tabular}{lrr}
      \\[-8pt]
      \toprule
      Method
      &
      $T$
      ($\times 10^{-4}$)
      &
      $\hat{T}_\mathrm{Dirichlet}$
      ($\times 10^{-4}$)
      \\
      \hline
(A) Without encoded boundary
&
$54.191 \pm 6.36$ &
$0.0000 \pm 0.0000$
\\
      \makecell[l]{
        (B) Without boundary condition
        \\
        in the neural nonlinear solver
      } &
$390.828 \pm 24.58$ &
$0.0000 \pm 0.0000$
\\
(C) Without neural nonlinear solver
&
$6.630 \pm 1.21$ &
$0.0000 \pm 0.0000$
\\
(D) Without boundary condition input
&
$465.492 \pm 26.47$ &
$868.7009 \pm 15.5447$
\\
(E) Without Dirichlet layer
&
$2.860 \pm 2.46$ &
$1.1703 \pm 0.0328$
\\
(F) Without pseudoinverse decoder
&
$44.947 \pm 6.00$ &
$9.7130 \pm 0.1201$
\\
      \makecell[l]{
        (G) Without pseudoinverse decoder
        \\
        with Dirichlet layer after decoding
      } &
$4.907 \pm 4.87$ &
$0.0000 \pm 0.0000$
\\
      \textbf{PENN}
&
      $\boldsymbol{1.795} \pm 1.33$ &
$0.0000 \pm 0.0000$
\\
    \bottomrule
    \end{tabular}
  }
\end{table}
\egroup

\end{document}

%% file: math_commands.tex
\usepackage{amsmath,amsfonts,bm}

\def\Figref#1{Figure~\ref{#1}}

\def\Secref#1{Section~\ref{#1}}

\def\eqref#1{equation~\ref{#1}}
\def\Eqref#1{Equation~\ref{#1}}

\def\1{\bm{1}}

\def\vb{{\bm{b}}}
\def\vc{{\bm{c}}}

\def\ve{{\bm{e}}}
\def\vf{{\bm{f}}}
\def\vg{{\bm{g}}}
\def\vh{{\bm{h}}}

\def\vn{{\bm{n}}}

\def\vt{{\bm{t}}}
\def\vu{{\bm{u}}}
\def\vv{{\bm{v}}}

\def\vx{{\bm{x}}}

\def\mA{{\bm{A}}}
\def\mB{{\bm{B}}}

\def\mM{{\bm{M}}}

\def\mR{{\bm{R}}}

\def\mW{{\bm{W}}}

\DeclareMathAlphabet{\mathsfit}{\encodingdefault}{\sfdefault}{m}{sl}
\SetMathAlphabet{\mathsfit}{bold}{\encodingdefault}{\sfdefault}{bx}{n}

\DeclareMathOperator*{\argmin}{arg\,min}